\RequirePackage{fix-cm}
\documentclass[twocolumn]{svjour3}          
\smartqed  
\usepackage{graphicx}
\usepackage{epsfig}
\usepackage{amsmath}
\usepackage{amssymb}

\usepackage{xcolor}
\usepackage{soul}
\usepackage[utf8]{inputenc}
\usepackage[breaklinks=true,letterpaper=true,colorlinks,bookmarks=false,urlcolor=red,citecolor=cyan]{hyperref}

\usepackage{gensymb,graphics,subfigure,inputenc}
\usepackage[linesnumbered,algo2e,boxed]{algorithm2e}
\graphicspath{{Figures/}}
\usepackage[figuresright]{rotating}
\usepackage{amsmath,amssymb,textcomp,subfigure,multirow,upgreek}
\usepackage{booktabs}
\usepackage{color,mdwlist}

\newcommand{\tht}[2]{\begin{tabular}{@{}#1@{}}#2\end{tabular}}

\usepackage{longtable}

\usepackage{ragged2e}

\usepackage{array}
\usepackage{tikz}

\usepackage{graphicx,fontawesome}
\graphicspath{{Figures/}}

\usepackage{amssymb} 
\usepackage{xcolor} 
\usepackage{caption}
\usepackage{wrapfig}
\graphicspath{ {image/} }  
\usepackage{enumitem}
\setitemize{noitemsep,topsep=0pt,parsep=0pt,partopsep=0pt}

\begin{document}

\title{Text-to-Image Synthesis: A Decade Survey}


\author{       
Nonghai Zhang \and Hao Tang$^*$  
}


\institute{
Nonghai Zhang \at School of Software \& Microelectronics, Peking University,  Beijing 100871, China. \email{znh@stu.pku.edu.cn} \\
Hao Tang ($^*$corresponding author) \at School of Computer Science, Peking University, Beijing 100871, China. \email{bjdxtanghao@gmail.com} 
}

\date{Received: date / Accepted: date}

\maketitle

\justify
\begin{abstract}
Text-to-image synthesis (T2I) focuses on generating high-quality images from textual descriptions
In this survey, we review 490 papers on T2I spanning the past decade, with representative advances covered through 2026. We start by briefly introducing how GANs, autoregressive models, and diffusion models have been used for image generation, and we additionally examine the shared components whose evolution has driven recent progress. Building on this foundation, we discuss the development of these models for T2I and organize cutting-edge research by problem rather than by chronology alone, covering controllability, personalized and identity-preserving generation, instruction-based image editing, preference alignment and post-training with human and AI feedback, efficiency, safety, and consistency in content and spatial relationships. We further review unified multimodal understanding-and-generation models, summarize the datasets and evaluation metrics commonly used in T2I research, and provide a comparative discussion of the three paradigms together with their respective trade-offs. Finally, we discuss the potential applications of T2I within AIGC, along with the challenges and forward-looking research opportunities in this field.

\end{abstract}
\definecolor{tgan}{HTML}{C65D07}
\definecolor{tganfill}{HTML}{FFF2E7}
\definecolor{tar}{HTML}{B33A3A}
\definecolor{tarfill}{HTML}{FCEBEC}
\definecolor{tdiff}{HTML}{2463A6}
\definecolor{tdifffill}{HTML}{EAF2FB}
\definecolor{tuni}{HTML}{237A57}
\definecolor{tunifill}{HTML}{E9F6F0}
\definecolor{tgrid}{HTML}{DDE3E8}
\definecolor{tline}{HTML}{AEB8C0}

\section{Introduction}
\label{section1}
Text and images are fundamental means of recording the objective world. As the saying goes, ``A picture is worth a thousand words''. When we read pure text, our minds often create corresponding images to help us better understand the content. In computing, text-to-image generation (T2I) \cite{mansimov2015generating,reed2016generative,zhang2017stackgan,xu2018attngan,zhu2019dm,cheng2020rifegan,ramesh2021zero,nichol2021glide,ding2021cogview,ramesh2022hierarchical,ding2022cogview2,saharia2022photorealistic,zhang2023adding,podell2023sdxl,zheng2024cogview3} combines the fields of natural language processing (NLP) \cite{bengio2000neural,vaswani2017attention,brown2020language} and computer vision (CV) \cite{marr2010vision,krizhevsky2012imagenet,dosovitskiy2020image}, referring to the process of generating realistic images from textual descriptions using specific models.
Due to its potential to revolutionize content creation in various domains, T2I has received significant attention. By converting textual descriptions into visual content, this technology bridges the gap between language and imagery, opening new possibilities for art, design, and multimedia applications. Moreover, T2I plays a key role in AI-generated content (AIGC) \cite{wu2023ai} and represents an important milestone on the path towards general artificial intelligence \cite{goertzel2007artificial,goertzel2014artificial,pei2019towards}. Figure \ref{model_timeline} shows a series of representative works in the T2I field.

\begin{figure*}[ht]
    \centering
    \resizebox{\linewidth}{!}{
\definecolor{tgan}{HTML}{C65D07}
\definecolor{tganfill}{HTML}{FFF2E7}
\definecolor{tar}{HTML}{B33A3A}
\definecolor{tarfill}{HTML}{FCEBEC}
\definecolor{tdiff}{HTML}{2463A6}
\definecolor{tdifffill}{HTML}{EAF2FB}
\definecolor{tuni}{HTML}{237A57}
\definecolor{tunifill}{HTML}{E9F6F0}
\definecolor{tgrid}{HTML}{DDE3E8}
\definecolor{tline}{HTML}{AEB8C0}

\begin{tikzpicture}[
    x=1.70cm,
    y=0.95cm,
    every node/.style={font=\sffamily},
    event/.style={
        rounded corners=1.2pt,
        line width=0.55pt,
        align=center,
        text width=1.48cm,
        minimum height=0.45cm,
        inner xsep=1.1pt,
        inner ysep=1.2pt,
        font=\fontsize{4.70}{5.05}\selectfont
    },
    green event/.style={
        event,
        text width=1.30cm,
        minimum height=0.40cm,
        inner ysep=1.0pt,
        font=\fontsize{4.45}{4.80}\selectfont
    },
    lane/.style={font=\fontsize{7.4}{8.0}\selectfont\bfseries,align=right},
    year/.style={font=\fontsize{7.0}{7.5}\selectfont\bfseries,text=black!68}
]

\newcommand{\tlname}[2]{\textbf{#1}\hspace{0.05em}{\fontsize{3.45}{3.70}\selectfont\cite{#2}}}
\newcommand{\tlevent}[6]{%
    \fill[#4] (#1,#2) circle (1.35pt);%
    \draw[#4,line width=0.55pt] (#1,#2) -- (#1,{#2+0.56*(#3-#2)});%
    \node[event,draw=#4,fill=#5] at (#1,#3) {#6};%
}
\newcommand{\tlgreenevent}[3]{%
    \fill[tuni] (#1,0.00) circle (1.35pt);%
    \draw[tuni,line width=0.55pt] (#1,0.00) -- (#1,{0.58*(#2)});%
    \node[green event,draw=tuni,fill=tunifill] at (#1,#2) {#3};%
}

\foreach \x [evaluate=\x as \yr using int(2015+\x)] in {0,...,11}{
    \draw[tgrid,line width=0.4pt] (\x,-1.28) -- (\x,6.92);
    \node[year,anchor=north] at (\x,-1.55) {\yr};
}

\foreach \y in {6.30,4.20,2.10,0.00}{
    \draw[tline,line width=0.85pt] (0,\y) -- (11,\y);
}
\node[lane,text=tgan,anchor=east] at (-0.18,6.30) {GAN-based T2I};
\node[lane,text=tar,anchor=east] at (-0.18,4.20) {AR \& token models};
\node[lane,text=tdiff,anchor=east] at (-0.18,2.10) {Diffusion \& flow};
\node[lane,text=tuni,anchor=east] at (-0.18,0.00)
    {Unified /\\editing /\\alignment /\\evaluation};

\tlevent{0}{6.30}{6.98}{tgan}{tganfill}{\tlname{AlignDRAW}{mansimov2015generating}}
\tlevent{1}{6.30}{5.62}{tgan}{tganfill}{\tlname{Text-GAN}{reed2016generative}}
\tlevent{2}{6.30}{6.98}{tgan}{tganfill}{\tlname{StackGAN}{zhang2017stackgan}}
\tlevent{3}{6.30}{5.62}{tgan}{tganfill}{\tlname{AttnGAN}{xu2018attngan}}
\tlevent{4}{6.30}{6.98}{tgan}{tganfill}{\tlname{DM-GAN}{zhu2019dm}}
\tlevent{7}{6.30}{5.62}{tgan}{tganfill}{\tlname{DF-GAN}{tao2022df}}
\tlevent{8}{6.30}{6.98}{tgan}{tganfill}{\tlname{GigaGAN}{kang2023scaling}}

\tlevent{6}{4.20}{4.88}{tar}{tarfill}{\tlname{DALL-E}{ramesh2021zero}\\\tlname{CogView}{ding2021cogview}}
\tlevent{7}{4.20}{3.52}{tar}{tarfill}{\tlname{Parti}{yu2022scaling}\\\tlname{CogView2}{ding2022cogview2}}
\tlevent{7.45}{4.20}{4.88}{tar}{tarfill}{\tlname{MaskGIT}{chang2022maskgit}}
\tlevent{8}{4.20}{3.52}{tar}{tarfill}{\tlname{Muse}{chang2023muse}}
\tlevent{9}{4.20}{4.88}{tar}{tarfill}{\tlname{HART}{tang2024hart}}
\tlevent{10}{4.20}{3.52}{tar}{tarfill}{\tlname{NextStep-1}{team2025nextstep}}

\tlevent{5}{2.10}{2.78}{tdiff}{tdifffill}{\tlname{DDPM}{ho2020denoising}}
\tlevent{6}{2.10}{1.42}{tdiff}{tdifffill}{\tlname{GLIDE}{nichol2021glide}}
\tlevent{7}{2.10}{2.78}{tdiff}{tdifffill}{\tlname{Imagen}{saharia2022photorealistic}\\\tlname{LDM}{rombach2022high}}
\tlevent{8}{2.10}{1.42}{tdiff}{tdifffill}{\tlname{ControlNet}{zhang2023adding}\\\tlname{SDXL}{podell2023sdxl}}
\tlevent{9}{2.10}{2.78}{tdiff}{tdifffill}{\tlname{SD3}{esser2024scaling}\\\tlname{PixArt}{chen2023pixart}}
\tlevent{10}{2.10}{1.42}{tdiff}{tdifffill}{\tlname{Flow-GRPO}{liu2025flowgrpo}}
\tlevent{11}{2.10}{2.78}{tdiff}{tdifffill}{\tlname{DenseGRPO}{deng2026densegrpo}\\\tlname{PE-OPSD}{lin2026peopsd}}

\tlgreenevent{7}{0.68}{\tlname{Unified-IO}{lu2022unifiedio}}
\tlgreenevent{8}{-0.68}{\tlname{CM3Leon}{yu2023scaling}}
\tlgreenevent{9}{0.68}{\tlname{Show-o}{xie2024showo}}
\tlgreenevent{10}{-0.68}{\tlname{Janus}{wu2024janus}\\\tlname{BAGEL}{deng2025emerging}}
\tlgreenevent{10.45}{0.68}{\tlname{Harmon}{wu2025harmonizing}\\\tlname{Kontext}{labs2025flux}}
\tlgreenevent{11}{-0.76}{\tlname{MME-Unify}{xie2025mmeunify}\\\tlname{Qwen-Image-2.0-RL}{xu2026qwenimagerl}}

\end{tikzpicture}}
    \caption{Supplementary survey timeline of representative T2I milestones from 2015 to 2026. The lanes indicate historical emphasis rather than mutually exclusive model families; hybrid systems can combine different representations, objectives, backbones, and tasks. This figure is a qualitative synthesis and is not a performance ranking.}
    \label{model_timeline}
    \vspace{-0.4cm}
\end{figure*}

With the advancement of deep learning \cite{lecun2015deep}, T2I models have become capable of generating more refined images that closely match textual descriptions. Looking back at the evolution of T2I models, AlignDRAW \cite{mansimov2015generating}, proposed by Elman Mansimov's team, is considered a pioneering work in this field. This model effectively demonstrated the relationship between generated images and text descriptions by introducing attention mechanisms \cite{niu2021review}, although the quality of the generated images still needed improvement in certain cases.

The introduction of Generative Adversarial Networks (GANs) \cite{goodfellow2014generative} brought significant progress to T2I. Inspired by conditional GAN (cGAN) \cite{mirza2014conditional}, Reed et al. designed the GAN-CLS and GNA-INT models \cite{reed2016generative}, which were the first to apply GAN to T2I tasks, demonstrating the advantages of GAN in generating high-quality, detail-rich images. This phase marks an important milestone for the use of GANs in T2I \cite{zhang2017stackgan,xu2018attngan,tao2022df}.

Using the Transformer architecture \cite{vaswani2017attention} from NLP, OpenAI introduced the DALL-E model \cite{ramesh2021zero}. This model was the first to employ an autoregressive (AR) for image generation, utilizing large-scale datasets to produce diverse images. Although the AR approach has demonstrated high-quality generation capabilities \cite{ding2021cogview,yu2022scaling,he2024mars}, its substantial computational cost limits its practicality in certain application scenarios.

In recent years, diffusion models (DM) \cite{sohl2015deep,ho2020denoising}, inspired by nonequilibrium thermodynamics, have gradually become the most advanced approach in the T2I field. GLIDE \cite{nichol2021glide} was the first work to apply diffusion models to T2I tasks, demonstrating exceptional generative capabilities by operating in pixel space. The introduction of Latent Diffusion Model (LDM) \cite{rombach2022high} highlighted the importance of latent space in diffusion models, significantly improving the quality of the generated images. As research has progressed, diffusion models have demonstrated unparalleled effectiveness \cite{saharia2022photorealistic,zhang2023adding,zheng2024cogview3}, making them one of the most popular research directions in T2I.

Since 2024, progress in T2I has been driven less by the emergence of a new model family than by changes across the components shared among existing paradigms. On the architecture side, the convolutional U-Net denoiser has been largely replaced by diffusion Transformers (DiT) \cite{peebles2022scalable}, and flow-matching and rectified-flow objectives \cite{lipman2022flow,liu2022flow} underlie systems such as Stable Diffusion 3 \cite{esser2024scaling}. On the conditioning side, text encoders have evolved from CLIP \cite{radford2021learning} toward T5 and vision-language encoders, as in Qwen-Image \cite{wu2025qwenimage}, while the interaction between the encoder and the generator has itself become an object of study \cite{wang2025a,spingarn2026texttoimage}. In parallel, three directions have come to define the 2025--2026 landscape: (i) preference alignment and post-training with human or AI feedback, ranging from reward models and direct preference optimization \cite{xu2024imagereward,wallace2023diffusion} to reinforcement learning for flow models such as Flow-GRPO \cite{liu2025flowgrpo} and system-level pipelines such as Qwen-Image-2.0-RL \cite{xu2026qwenimagerl}; (ii) instruction-based image editing, exemplified by InstructPix2Pix \cite{brooks2022instructpixpix}, Step1X-Edit \cite{liu2025stepxedit}, and FLUX.1 Kontext \cite{labs2025flux}; and (iii) unified models that couple image generation with multimodal understanding, such as Show-o \cite{xie2024showo}, BAGEL \cite{deng2025emerging}, and Harmon \cite{wu2025harmonizing}. These developments motivate a survey that is organized not only by model family and chronology but also by shared components and research problems, and that remains current through 2026.

The continuous advancement of T2I technology has sparked enthusiastic discussions within the community \cite{zhou2023vision+}. Various research teams are consistently publishing new papers, and the rapid pace of technological updates poses significant challenges for newcomers trying to get started and stay up to date. Current studies have explored the development of GANs in T2I \cite{frolov2021adversarial,zhou2021survey}, the progress of DM models \cite{zhang2023text}, and relevant work on controllable generation modules \cite{cao2024controllable}. However, comprehensive reviews of the latest directions in T2I remain limited.
To help researchers understand the latest advances in T2I, this paper will introduce the basic principles of the GAN, AR, and DM models and provide a detailed review of their progress. Furthermore, this paper will conduct a comprehensive survey of cutting-edge research directions in T2I, with the aim of providing researchers with a clear roadmap and valuable references for further exploration.

The structure of this paper is as follows. Section \ref{sectiron2} introduces the mathematical principles and model structures of GANs, AR models, and diffusion models, and additionally analyzes the shared components whose evolution underlies recent progress---the shift from U-Net to DiT backbones, flow-based training objectives, and text and multimodal conditioning---together with a comparison of the three paradigms and their trade-offs. Section \ref{section3} discusses the cutting-edge works in T2I, covering the historical development of the three model families and unified multimodal understanding-and-generation. Section \ref{section4} then surveys the main research directions in a problem-oriented way, including controllable generation, personalized and identity-preserving generation, instruction-based image editing, preference alignment and post-training, efficiency, safety, and consistency. Section \ref{section5} describes the datasets and evaluation metrics currently employed in T2I research, including recent benchmarks and a discussion of evaluation reliability. Section \ref{section6} presents the applications of T2I and their connection to these technical advances, highlighting its significance and potential impact on AIGC. Finally, Section \ref{section7} summarizes the research achievements in T2I, reviews past challenges, and discusses forward-looking development directions and recommendations for further research. Extended chronological details of model development are provided in Appendix~\ref{section3back}.
\section{Basic models of T2I}
\label{sectiron2}
The currently recognized mainstream foundational models for T2I include GANs, autoregressive models, and diffusion models. In this section, we will explore the mathematical principles and underlying mechanisms of these models to understand why they are effective and why they serve as the foundation for T2I research.

\subsection{Generative Adversarial Networks (GAN)}
GAN was introduced in 2014 \cite{goodfellow2014generative} and has since been widely applied in various fields of CV \cite{tang2019multi,tang2020local,wu2022cross_tmm,tang2024graph,tang2023edge,tang2023graph} and NLP \cite{croce2020gan} fields, achieving remarkable results. A GAN consists of two main components: a Generator and a Discriminator. The Generator is responsible for creating samples, while the Discriminator's role is to distinguish between real samples and those generated by the Generator.
Specifically, the Generator creates samples from noise (i.e., random values), and the optimization problem of a GAN is formulated as a minimax problem, with its objective function shown in Eq. \eqref{eq1}. Given the Generator $G$, the optimal solution for the Discriminator $D$ is to maximize $V(D)$.
\begin{equation}
\begin{aligned}
\min_G \max_D V(D, G) &= \mathbb{E}_{x \sim p_{\text{data}}(x)} \left[ \log D(x) \right] \\
& \quad + \mathbb{E}_{z \sim p_z(z)} \left[ \log(1 - D(G(z))) \right].
\label{eq1}
\end{aligned}
\end{equation}

\begin{figure}[!t]    
	\centering
    \includegraphics[width=1\linewidth]{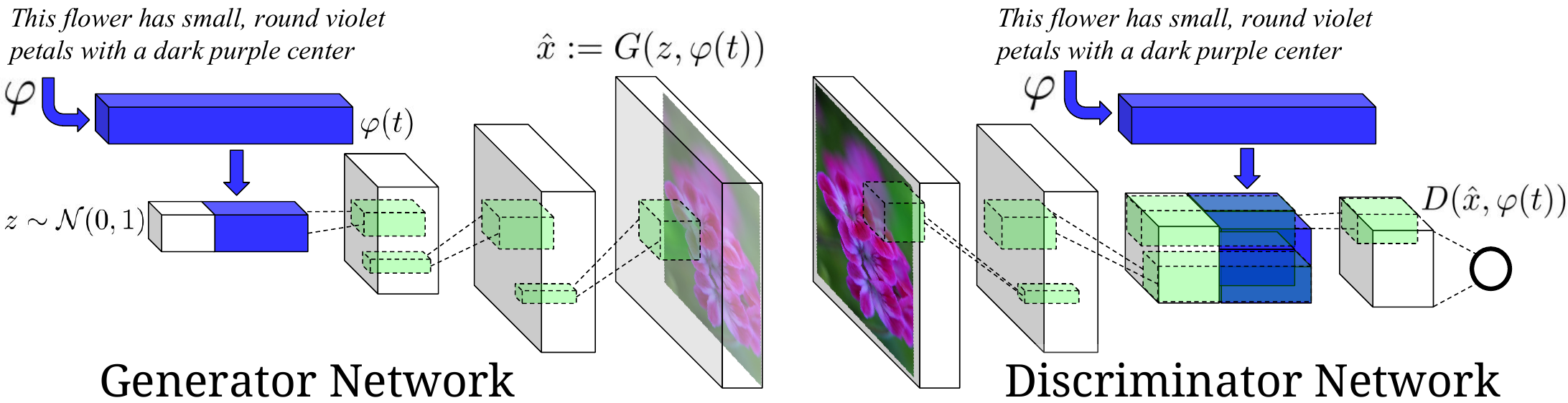} 
	\caption{The text-conditional convolutional GAN architecture \cite{reed2016generative}. Text encoding $p(t)$ is used by both generator and discriminator. It is projected to lower dimensions and depth concatenated with image feature maps for further stages of convolutional processing.}  
	\label{FIG:2}
        \vspace{-0.4cm}
\end{figure}

In T2I tasks, the GAN working mechanism is more complex and sophisticated \cite{reed2016generative,zhang2021cross,dinh2022tise,tao2023galip}. As illustrated in Figure \ref{FIG:2}, the Generator produces corresponding images by transforming text descriptions into latent vectors \cite{kingma2013auto}. This process often involves Conditional GAN (cGAN) techniques \cite{mirza2014conditional}, where the Discriminator not only evaluates the quality of the generated images, but also takes into account the related textual information. The feedback from the Discriminator helps the Generator continuously refine its generation strategy, resulting in images that become increasingly realistic and consistent.

\subsection{Autoregressive Models (AR)}
Autoregressive models were initially used for NLP tasks \cite{radford2018improving}, and their capabilities were further expanded with the introduction of the Transformer architecture \cite{vaswani2017attention}. The Image Generative Pre-trained Transformer (iGPT) \cite{chen2020generative} is a significant work that brought Transformers into the realm of image generation, showcasing their potential in this area. Eq. \eqref{AR-eq} illustrates how the model relies on previously generated parts and the input text prompt when generating each pixel or feature.
\begin{equation}
\begin{aligned}
P(X | T) = \prod_{t=1}^{T} P(X_t | X_{<t}, T).
\label{AR-eq}
\end{aligned}
\end{equation}

\begin{figure}[!t]    
	\centering
    \includegraphics[width=1\linewidth]{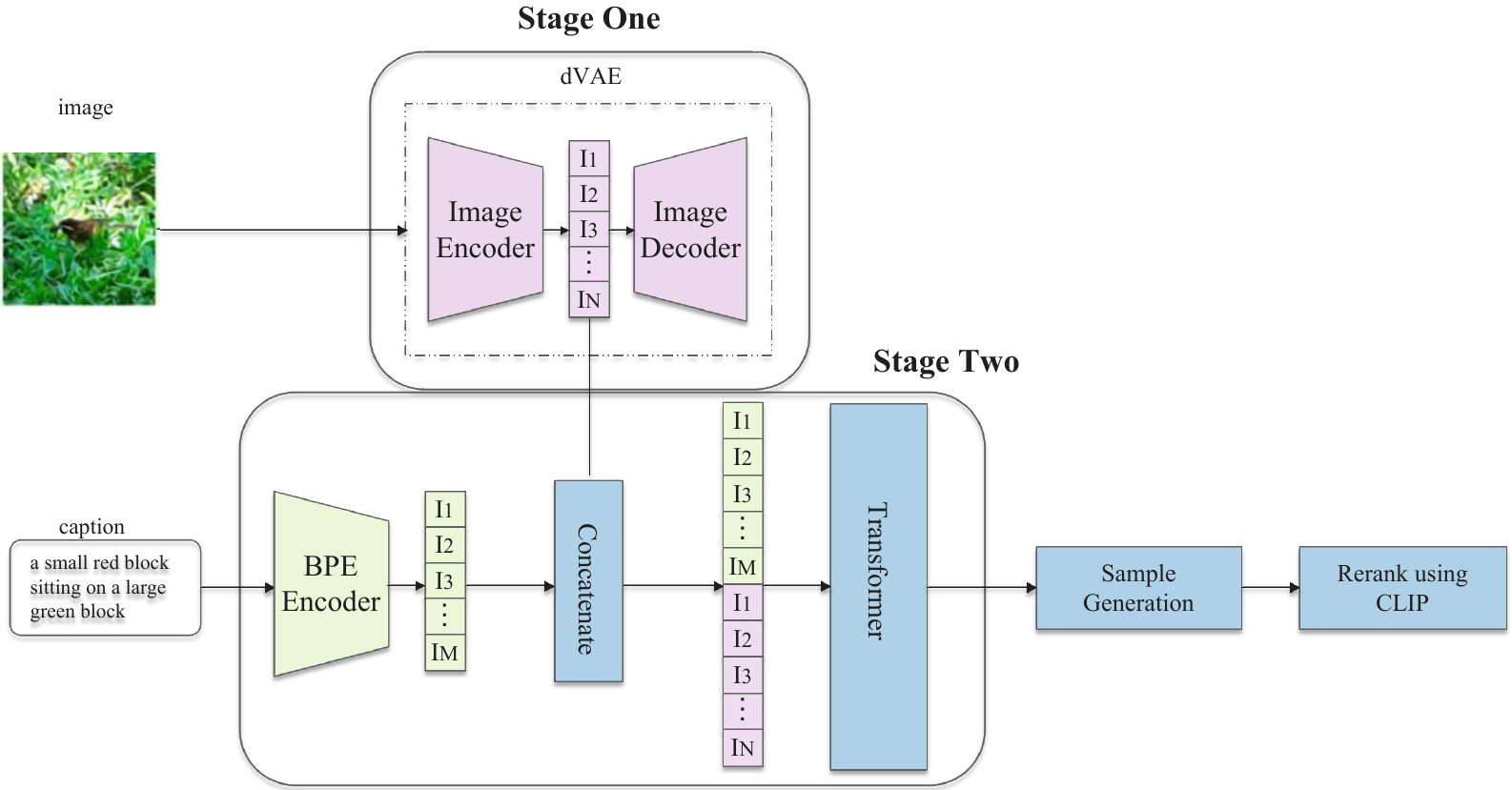} 
	\caption{The training of DALL-E \cite{ramesh2021zero} is divided into two stages. The first stage trains the codebook of VQ-VAE, while the second stage trains the Transformer, corresponding to Stage One and Stage Two indicated in the figure.}  
	\label{FIG:3}
        \vspace{-0.4cm}
\end{figure}

Inspired by iGPT, DALL-E \cite{ramesh2021zero} and CogView \cite{ding2021cogview} were among the first models to apply Transformers to T2I. By generating sequences step by step, these models effectively capture the complex semantic relationships between text and images, achieving results comparable to the state-of-the-art GAN models \cite{xu2018attngan,zhu2019dm,tao2022df} of their time.
As illustrated in Figure \ref{FIG:3}, DALL-E uses a two-stage Transformer architecture that maps text and image features into a shared latent space and progressively generates each pixel of the image. In the first stage, DALL-E employs a discrete variational autoencoder (dVAE) \cite{rolfe2016discrete} to compress images. In the second stage, it concatenates 256 text tokens encoded by byte pair encoding (BPE) with 1024 image tokens (32 $\times$ 32) to train an autoregressive Transformer that models the joint distribution of text and image tokens. For the generated series of images, DALL-E utilizes a pretrained CLIP model \cite{radford2021learning} to evaluate the correspondence between the generated images and the text captions, scoring and ranking them accordingly.

\begin{figure}[!t]     
	\centering
    \includegraphics[width=1\linewidth]{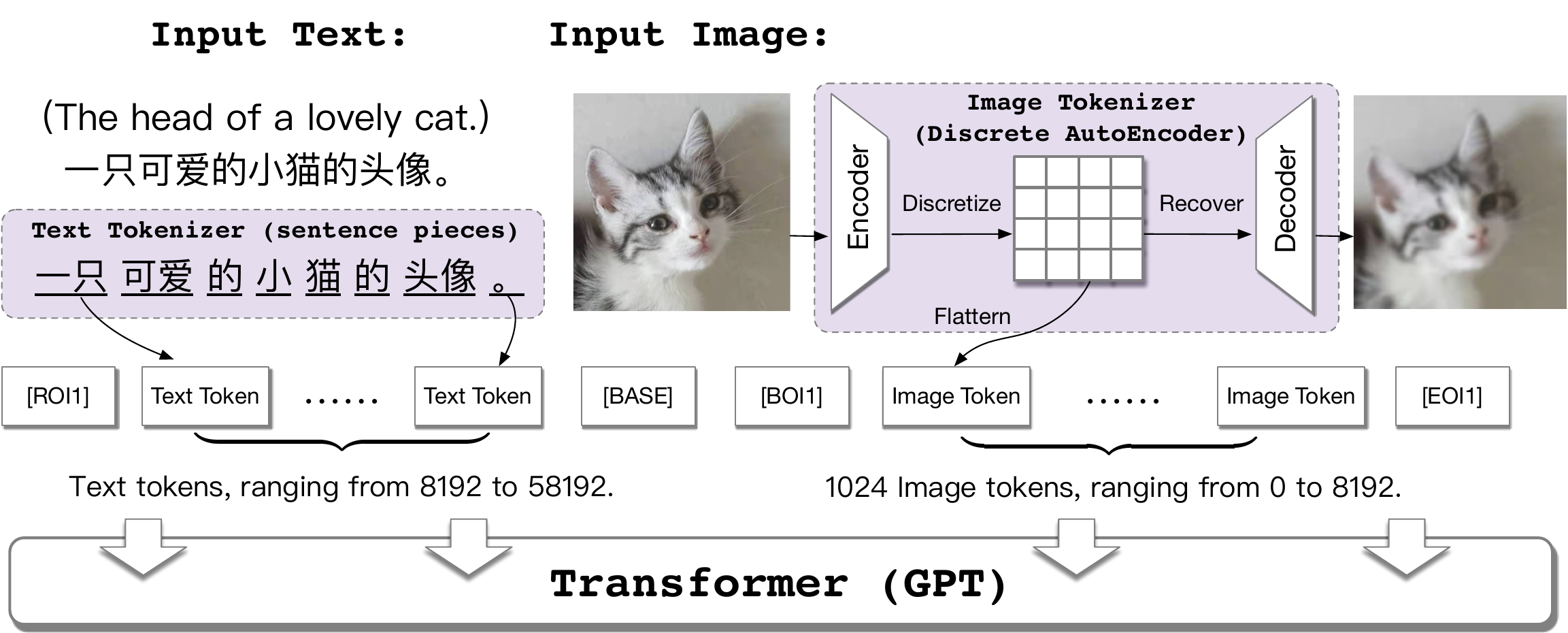} 
	\caption{The framework of CogView \cite{ding2021cogview}. [ROI1], [BASE1], etc., are seperator tokens.}   
	\label{FIG:4}
        \vspace{-0.4cm}
\end{figure}

As illustrated in Figure \ref{FIG:4}, CogView combines text descriptions with image generation employing an autoregressive approach to progressively generate images. First, text and images are converted to tokens using SentencePiece \cite{kudo2018sentencepiece} for text and a discrete autoencoder, similar to stage 1 of VQ-VAE \cite{van2017neural}, for images. The text and image tokens are then concatenated and entered into a unidirectional GPT \cite{radford2018improving} model to learn image generation.
During the T2I generation process, the trained model uses CLIP to rank the generated images by calculating a Caption Score \cite{vinyals2015show}, ultimately selecting the image that best matches the input text.

\begin{figure}[!t]     
	\centering
    \includegraphics[width=1\linewidth]{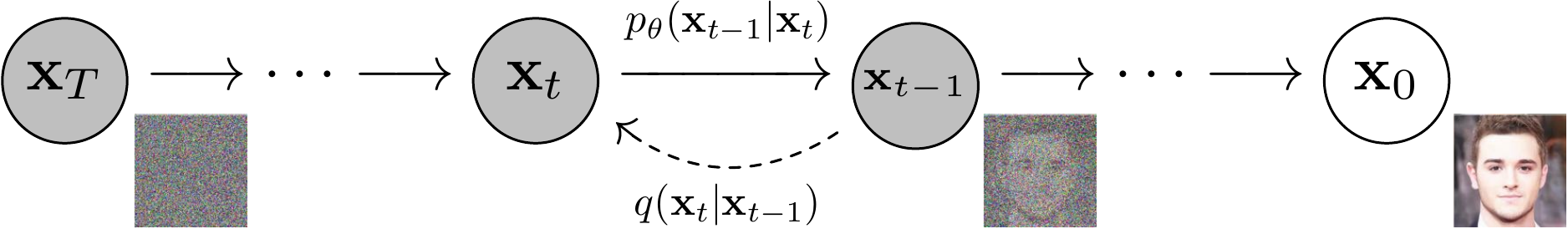}  
	\caption{DDPM \cite{ho2020denoising} generates new images by adding noise and then denoising.}  
	\label{FIG:5}
        \vspace{-0.4cm}
\end{figure}

\subsection{Diffusion Models (DM)}
Building on the fundamental concept of diffusion models \cite{sohl2015deep}, Denoising Diffusion Probabilistic Model (DDPM) \cite{ho2020denoising} was introduced in 2020, representing a major milestone in the field.
In the forward diffusion process, DDPM gradually adds noise to the data, ultimately resulting in a Gaussian noise distribution. The transformation of data $x_0$ into $x_t$ at each time step $t$ can be described by Eq. \eqref{forward}:
\begin{equation}
\begin{aligned}
x_t &= \sqrt{\alpha_t} x_0 + \sqrt{1 - \alpha_t} \epsilon_t.
\label{forward}
\end{aligned}
\end{equation}

In the denoising diffusion process, the model learns to recover the original data from noise. The generation process is described by Eq. \eqref{denoise}:
\begin{equation}
\begin{aligned}
p_\theta(x_{t-1} | x_t) &= \mathcal{N}(x_{t-1}; \mu_\theta(x_t, t), \Sigma_\theta(x_t, t)),
\label{denoise}
\end{aligned}
\end{equation}
where \( \mu_\theta \) and \( \Sigma_\theta \) are the mean and variance parameters learned by the model.

In T2I, the generation process is also conditioned on the text prompt, which can be represented by Eq. \eqref{condition}, where $T$ is the input text prompt, and the model generates images related to the text through conditional diffusion.
\begin{equation}
\begin{aligned}
p_\theta(x_0 | T) = \int p_\theta(x_0 | x_1) p_\theta(x_1 | T) \, dx_1.
\label{condition}
\end{aligned}
\end{equation}

As illustrated in Figure \ref{FIG:5}, the diffusion process gradually adds Gaussian noise to the original image in a fixed Markov chain, ultimately transforming the image into Gaussian noise. The reverse process then restores the original image step by step through denoising, achieving image generation. This approach is known for its excellent generative performance \cite{dhariwal2021diffusion}, which offers new insights for T2I tasks by progressively transforming text descriptions into high-quality images using the diffusion process.

\begin{figure}[!t]    
	\centering
    \includegraphics[width=1\linewidth]{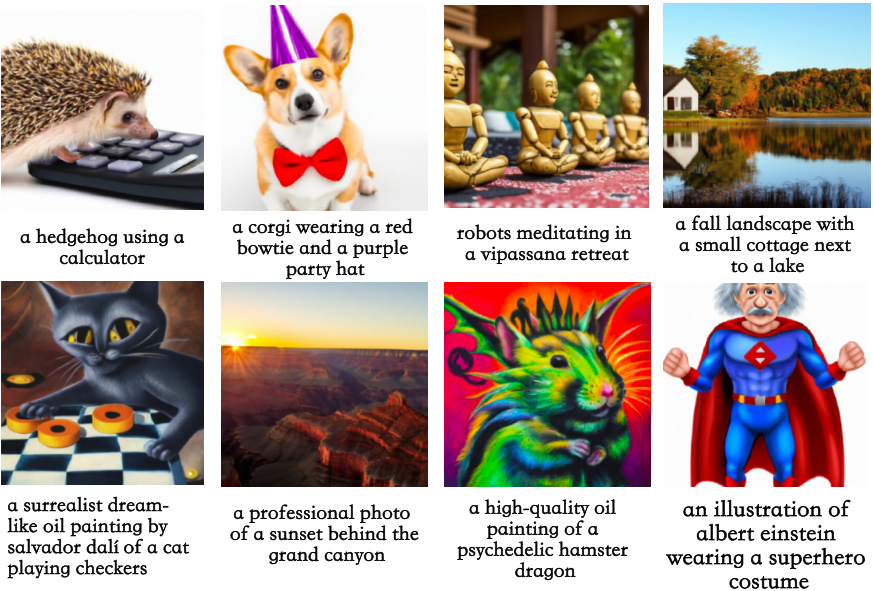} 
	\caption{Selected samples from GLIDE \cite{nichol2021glide} using classifier-free guidance. We observe that the model can produce photorealistic images with shadows and reflections, can compose multiple concepts in the correct way, and can produce artistic renderings of novel concepts.}
	\label{FIG:6}
        \vspace{-0.4cm}
\end{figure}

GLIDE \cite{nichol2021glide} is considered the first work to explore T2I using DDPM. This study compared two different guidance strategies: CLIP guidance \cite{radford2021learning} and classifier-free guidance \cite{ho2022classifier}. The latter achieved more realistic and detailed images by replacing original category labels with text. The results of GLIDE are shown in Figure \ref{FIG:6}.

\begin{figure}[!t]
	\centering
    \includegraphics[width=1\linewidth]{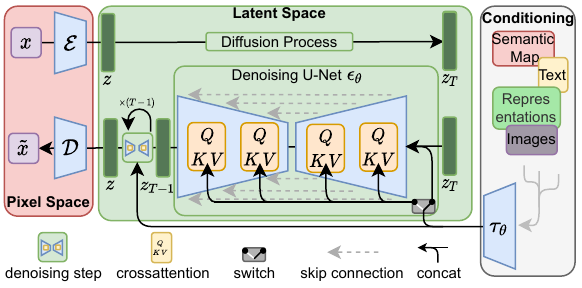}
	\caption{The condition LDM \cite{rombach2022high} either via concatenation or by a more general cross-attention mechanism.}
	\label{FIG:7}
        \vspace{-0.4cm}
\end{figure}

Stable Diffusion was the first to apply diffusion models in the latent space of a powerful pretrained autoencoder, serving as a demonstration of the Latent Diffusion Model (LDM) \cite{rombach2022high}. As illustrated in Figure \ref{FIG:7}, it follows a two-stage approach: in the first stage, the image is compressed into a latent representation \cite{kingma2013auto} to reduce computational complexity; in the second stage, a DDPM structure is used. In addition, a cross-attention mechanism \cite{vaswani2017attention} was introduced, which allows text or image sketches to influence the diffusion model and generate the desired images.

\subsubsection{Flow Matching and Rectified Flow}
The diffusion formulation above describes generation as the reversal of a fixed noising process. A closely related but distinct line of work replaces this stochastic view with a \emph{continuous-time transport} view. Flow matching \cite{lipman2022flow} learns a time-dependent velocity field $v_\theta(x,t)$ that transports a simple prior toward the data distribution along a prescribed probability path. With a linear interpolation path $x_t = (1-t)\,x_0 + t\,x_1$ between a noise sample $x_0$ and a data sample $x_1$, the conditional flow-matching objective takes the simple regression form in Eq.~\eqref{eq:fm}:
\begin{equation}
\begin{aligned}
\mathcal{L}_{\text{FM}} = \mathbb{E}_{t,\,x_0,\,x_1} \left\| v_\theta(x_t, t) - (x_1 - x_0) \right\|^2 .
\label{eq:fm}
\end{aligned}
\end{equation}

Rectified flow \cite{liu2022flow} studies the same transport problem but emphasizes making the trajectories as \emph{straight} as possible, so that the learned path can be followed accurately with few integration steps. The two ideas are therefore related but not synonymous: flow matching specifies a training objective over a probability path, whereas rectified flow is concerned with the geometry of the path and its reflow refinement. In practice, these objectives are often coupled with latent representations and large Transformer backbones rather than convolutional U-Nets; Stable Diffusion~3 \cite{esser2024scaling} and Qwen-Image \cite{wu2025qwenimage} are representative systems that adopt a (rectified-) flow objective. Because the objective, the backbone, and the sampler are independent design axes, a system-level comparison should report all three rather than attributing a difference to ``flow'' alone.

\subsection{Shared Components and Their Evolution}
\label{sec:shared-components}
The three paradigms above are usually presented as separate families, but modern T2I systems increasingly share two components whose evolution---rather than the choice of family---accounts for much of the recent progress: the denoising/generation \emph{backbone} and the \emph{conditioning interface} that injects language into the model.

\subsubsection{From U-Net to DiT and MMDiT}
\begin{figure}[!t]
	\centering
    \includegraphics[width=1\linewidth]{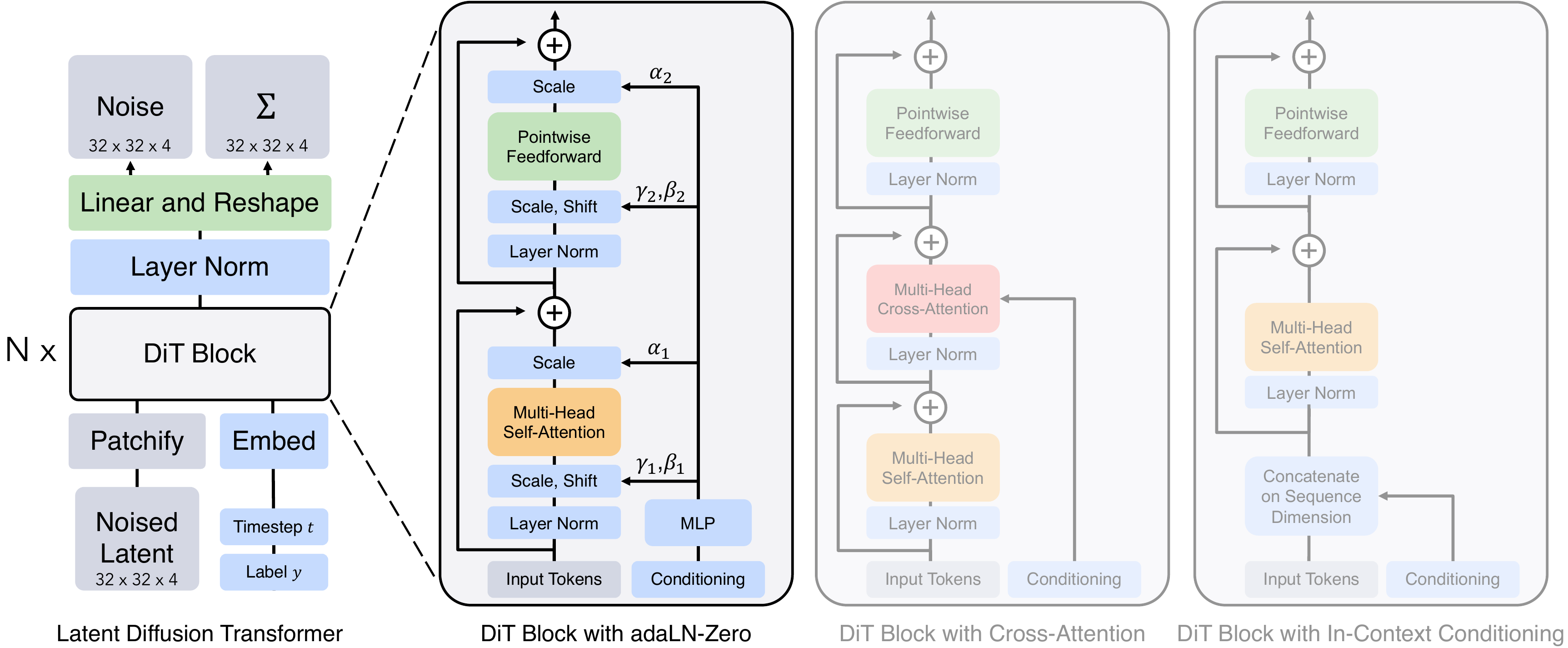}
	\caption{The Diffusion Transformer (DiT) architecture \cite{peebles2022scalable}. The convolutional U-Net denoiser is replaced by a Transformer that operates on a sequence of noised latent patches, with conditioning such as timestep and class or text embeddings injected through adaptive layer normalization.}
	\label{fig:dit}
        \vspace{-0.4cm}
\end{figure}

Early latent-diffusion systems such as Stable Diffusion \cite{rombach2022high} and SDXL \cite{podell2023sdxl} use a convolutional U-Net as the denoiser, with text injected through cross-attention. The Diffusion Transformer (DiT) \cite{peebles2022scalable} replaces this U-Net with a Transformer that operates on sequences of latent patches (Figure~\ref{fig:dit}), and shows that generation quality scales smoothly with model size and compute in the same way observed for Transformers elsewhere. PixArt-$\alpha$ \cite{chen2023pixart} demonstrates that a Transformer denoiser can reach competitive quality at markedly lower training cost, and MMDiT-style designs used in Stable Diffusion~3 \cite{esser2024scaling} extend the backbone with \emph{separate} weights for the text and image streams together with joint attention, so that the two modalities interact while retaining modality-specific parameters. Two cautions are worth stating. First, the backbone (U-Net vs.\ Transformer) is a different axis from the generative objective (score/diffusion vs.\ flow): a system can combine either backbone with either objective. Second, the gains reported for large DiT/MMDiT systems also reflect larger datasets, better autoencoders, and improved captions, so ``switching to a Transformer'' should not be credited with the entire improvement.

\subsubsection{Text and Multimodal Conditioning}
The conditioning interface determines how linguistic structure reaches the generator. Early systems relied on CLIP text embeddings \cite{radford2021learning}, which are aligned to images but relatively shallow as language models. Imagen \cite{saharia2022photorealistic} showed that a large frozen language encoder (T5 \cite{raffel2019exploring}) can substantially improve prompt fidelity, and recent systems such as Qwen-Image \cite{wu2025qwenimage} condition on vision-language encoders that process text and visual context jointly. Crucially, encoder capacity is not the only factor: controlled studies find that how representations are extracted from a decoder-only language model materially affects results \cite{wang2025a}, that text encoders can be scaled \emph{down} through visually grounded distillation with little loss \cite{wang2025scaling}, and that generators often fail to exploit the full contextual information already present in their text embeddings \cite{spingarn2026texttoimage}. The encoder, the adapter that maps its output into the generator, and the generator's own capacity should therefore be analyzed separately rather than summarized as a single ``bigger-encoder-is-better'' trend.

\subsection{Comparison of the Three Paradigms}
\label{sec:paradigm-comparison}
Having introduced GANs, AR models, and diffusion/flow models, we summarize their qualitative trade-offs in Table~\ref{tab:paradigm}. The entries describe \emph{typical} designs and common tendencies rather than absolute properties, and deliberately avoid cross-paper performance rankings, since reported scores depend on data, resolution, sampling budget, and evaluation protocol. In broad terms, GANs offer single-step, low-latency sampling but can be unstable to train and prone to limited mode coverage; AR models provide a unified token interface that composes naturally with language models and unified multimodal systems, at the cost of sequential decoding; and diffusion/flow models currently dominate open-ended T2I quality and controllability, at the cost of iterative sampling that recent distillation and few-step flow methods aim to reduce. The appropriate choice is therefore scenario-dependent: an interactive application with a strict latency budget may favor a single-step or few-step model, whereas a task requiring fine-grained control and editing (Section~\ref{section3}) currently favors diffusion/flow backbones with explicit conditioning interfaces.

\begin{table*}[t]
\centering\small
\caption{Qualitative comparison of the three T2I paradigms. Entries describe typical designs and tendencies, not absolute guarantees, and are not a performance ranking.}
\label{tab:paradigm}
\begin{tabular}{p{0.17\textwidth}p{0.24\textwidth}p{0.24\textwidth}p{0.25\textwidth}}
\toprule
Aspect & GAN & Autoregressive (AR) & Diffusion / Flow \\
\midrule
Training objective & Adversarial min--max game \cite{goodfellow2014generative} & Likelihood over image tokens \cite{ramesh2021zero} & Denoising score matching \cite{ho2020denoising} or flow matching \cite{lipman2022flow} \\
Image representation & Continuous pixels/latents & Discrete (or continuous) tokens & Continuous latents (or pixels) \\
Sampling & Single forward pass & Sequential token decoding & Iterative denoising / ODE integration \\
Main inference cost & Low (one step) & Grows with sequence length & Grows with number of steps \\
Typical risk & Training instability, mode collapse & Exposure bias, slow decoding & High sampling cost (mitigated by distillation / few-step flow) \\
Control \& editing & Latent manipulation & Token in-filling / prompting & Cross-attention control, inversion, instruction editing \\
Best-suited scenarios & Low-latency or single-step generation \cite{kang2023scaling} & Unification with language and multimodal models \cite{xie2024showo} & Open-ended high-fidelity generation, fine-grained control and editing \cite{esser2024scaling} \\
\bottomrule
\end{tabular}
\end{table*}
\section{Cutting-edge works in T2I}
\label{section3}

\begin{table*}[ht]
\centering
\caption{Overview of GAN models in T2I by year.}
\label{developgan}
			\resizebox{1\linewidth}{!}{%
\begin{tabular}{@{}l|l|l@{}}
\toprule
\textbf{Year} & \textbf{Source} & \textbf{Model and Description} \\ \midrule
2016 & ICML & \textbf{Text-conditional GAN} \cite{reed2016generative}: The first GAN used for T2I tasks. \\ \midrule
2017 & arXiv & \textbf{TAC-GAN} \cite{dash2017tac}: Incorporated auxiliary classification tasks into the discriminator. \\
     & ICCV & \textbf{StackGAN} \cite{zhang2017stackgan}: 1. Generate Stage-I low-resolution images; 2. Generate high-resolution images using text descriptions. \\
     & TPAMI & \textbf{StackGAN++} \cite{zhang2018stackgan++}: Composed of multiple generators and discriminators in a tree-like structure. \\ \midrule
2018 & ECCV & \textbf{Fused GAN} \cite{bodla2018semi}: Two generators for conditional and unconditional image synthesis. \\
     & CVPR & \textbf{HDGAN} \cite{zhang2018photographic}: Accompanying hierarchical-nested adversarial objectives inside the network hierarchies. \\
     & CVPR & \textbf{AttnGAN} \cite{xu2018attngan}: Integrates attention mechanisms into a multi-stage optimization pipeline. \\ \midrule
2019 & AAAI & \textbf{Text-SeGAN} \cite{cha2019adversarial}: Modify the discriminator to regress semantic relevance between text and image. \\
     & CVPR & \textbf{Obj-GANs} \cite{li2019object}: Using a novel object-wise discriminator based on the Fast R-CNN. \\
     & CVPR & \textbf{MirrorGAN} \cite{qiao2019mirrorgan}: Focuses on global and local attention and semantic preservation. \\
     & SIGGRAPH & \textbf{GauGAN} \cite{park2019gaugan}: Using spatially-adaptive normalization. \\
     & CVPR & \textbf{DM-GAN} \cite{zhu2019dm}: Refines blurry images using a dynamic storage module. \\
     & Access & \textbf{ControlGAN} \cite{lee2019controllable}: Manipulates visual attributes in T2I generation. \\ \midrule
2020 & CVPR & \textbf{RiFeGAN} \cite{cheng2020rifegan}:Enriching the given caption from prior knowledge, formed by the training dataset. \\
     & ECCV & \textbf{CPGAN} \cite{liang2020cpgan}: Design a memory mechanism in text and encode the generated image in an object-aware manner. \\
     & CVPR & \textbf{CookGAN} \cite{zhu2020cookgan}: Simulates visual effects in the causal chain. \\
     & arXiv & \textbf{SegAttnGAN} \cite{gou2020segattngan}: Utilizes segmentation information for synthesis. \\ \midrule
2021 & CVPR & \textbf{XMC-GAN} \cite{zhang2021cross}: Addresses cross-modal contrastive loss. \\
     & CVPR & \textbf{VQ-GAN} \cite{esser2021taming}: Discrete vector representations in VQ space. \\
     & ICCV & \textbf{DAE-GAN} \cite{ruan2021dae}: Comprehensively represents information from sentence-level, word-level, and aspect-level. \\
     & ACM MM & \textbf{CI-GAN} \cite{wang2021cycle}: Learns similarity between text representations and latent encodings. \\
     & ACM MM & \textbf{R-GAN} \cite{qiao2021r}: Constructing a scene graph to capture the monolithic structural representation of the text. \\ \midrule
2022 & ECCV & \textbf{VQGAN-CLIP} \cite{crowson2022vqgan}: Guides VQGAN using CLIP. \\
     & TMM & \textbf{VLMGAN} \cite{cheng2022vision}: Enhances image quality and consistency through a dual vision-language matching mechanism. \\
     & CVPR & \textbf{DF-GAN} \cite{tao2022df}: Novel one-stage text-to-image backbone, Target-Aware Discriminator and deep text-image fusion block. \\
     & CVPR & \textbf{SSA-GAN} \cite{liao2022text}: a Semantic-Spatial Aware block learns semantic-adaptive transformations and generates semantic masks. \\
     & CVPR & \textbf{LAFITE} \cite{zhou2022towards}: The first work
 enables the language-free training using pretrained CLIP model. \\
     & ECCV & \textbf{AttnGAN++} \cite{dinh2022tise}: Adding the spectral normalization layers to the discriminator based on AttnGAN. \\ \midrule
2023 & CVPR & \textbf{GigaGAN} \cite{kang2023scaling}: The first GAN-based method trains a billion-scale model on billions of real-world complex Internet images. \\
     & ICML & \textbf{StyleGAN-T} \cite{sauer2023stylegan}: Using Fourier features, a second-order style mechanism, and improved text conditioning mechanisms. \\
     & CVPR & \textbf{GALIP} \cite{tao2023galip}: Leverages the powerful pretrained CLIP model both in the discriminator and generator. \\
     & TMM & \textbf{RAT-GAN} \cite{ye2023recurrent}: Propose a recurrent affine transformation (RAT) connects all the CAT blocks. \\
     & CVPR & \textbf{Layout-VQGAN} \cite{wu2022text}: Propose an object-guided joint-decoding module. \\
     & SIGGRAPH & \textbf{DragGAN} \cite{pan2023drag}: An interactive point-based image editing framework. \\ \midrule
2024 & CVPR & \textbf{UFOGen} \cite{xu2024ufogen}: Hybrid approach integrating diffusion with GAN objectives. \\
     & CVPR & \textbf{HyperCGAN} \cite{haydarov2024adversarial}: Represents 2D images as implicit neural representations.  \\ \bottomrule
\end{tabular}}
    \vspace{-0.4cm}
\end{table*}

In Section \ref{sectiron2}, we introduce the foundational work on GAN \cite{reed2016generative}, AR \cite{ramesh2021zero}, and DM \cite{nichol2021glide} in the field of T2I. In this section, we will discuss the development of these three models in T2I, providing a comprehensive overview of their evolution from inception to the present, in order to illustrate the ongoing advancement of foundational models in T2I. 

Building on this, we will conduct a detailed investigation of the current research directions in T2I, including various recent studies. Finally, we will discuss the datasets and evaluation metrics, emphasizing their importance in T2I. Overall, this chapter encompasses a range of cutting-edge work in T2I.

\subsection{Development of GAN, AR, and DM in T2I}
As shown in Table \ref{developgan}, the application of GANs in the field of T2I has undergone significant advances. Starting from the pioneering work of Text-conditional GAN \cite{reed2016generative}, a series of studies, including StackGAN \cite{zhang2017stackgan}, StackGAN++ \cite{zhang2018stackgan++}, AttnGAN \cite{xu2018attngan}, DM-GAN \cite{zhu2019dm}, MirrorGAN \cite{qiao2019mirrorgan}, CPGAN \cite{liang2020cpgan}, VQ-GAN \cite{esser2021taming}, DAE-GAN \cite{ruan2021dae}, DF-GAN \cite{tao2022df}, VQGAN-CLIP \cite{crowson2022vqgan}, and the recent GigaGAN \cite{kang2023scaling}, have established GANs as a landmark model in T2I research. These models will be introduced in chronological order in the appendix. These continuously evolving GAN models have not only improved the quality of text-to-image generation but have also expanded their range of applications, driving rapid advancements in the T2I field.

As shown in Table \ref{developAR}, The latest advances in autoregressive models in the field of T2I demonstrate their strong generative capabilities and remarkable flexibility. DALL-E \cite{ramesh2021zero} is the first work applying Autoregressive methods to T2I. With the development of models such as CogView \cite{ding2021cogview}, NÜWA \cite{wu2022nuwa}, Make-A-Scene \cite{gafni2022make}, CogView2 \cite{ding2022cogview2}, Parti \cite{yu2022scaling}, Emu \cite{dai2023emu}, CM3Leon \cite{yu2023scaling}, MARS \cite{he2024mars}, and GILL \cite{koh2024generating}, these continuously evolving autoregressive models have not only improved the quality of text-to-image generation but also driven rapid advancements in the field. More recent work further loosens the assumption that AR generation requires discrete tokens: NextStep-1 \cite{team2025nextstep} pairs next-token prediction over \emph{continuous} image tokens with a lightweight flow-matching head, while the unified autoregressive systems discussed in Section~\ref{sec:unified} reuse this token interface to couple generation with multimodal understanding.

\begin{table*}[ht]
\centering
\caption{Overview of AR models in T2I by year.}
\label{developAR}
			\resizebox{1\linewidth}{!}{%
\begin{tabular}{@{}l|l|l@{}}
\toprule
\textbf{Year} & \textbf{Source} & \textbf{Model and Description} \\ \midrule
2021 & ICML & \textbf{DALL-E} \cite{ramesh2021zero}: The first to use AR for generation in T2I. \\ \midrule
     & NeurIPS & \textbf{CogView} \cite{ding2021cogview}: A 4-billion-parameter Transformer with a VQ-VAE tokenizer. \\ 
     & arXiv & \textbf{M6} \cite{lin2021m6}: The largest dataset for Chinese multimodal pre-training. \\ \midrule
2022 & TMLR & \textbf{Parti} \cite{yu2022scaling}: Uses image token sequences as the target output instead of text tokens. \\ 
     & NeurIPS & \textbf{CogView2} \cite{ding2022cogview2}: A solution based on hierarchical transformers and local parallel autoregressive generation. \\ 
     & ECCV & \textbf{Make-A-Scene} \cite{gafni2022make}: Control mechanism complementary to text in the form of a scene. \\ 
     & ECCV & \textbf{NÜWA} \cite{wu2022nuwa}: Using a 3D transformer encoder-decoder framework. \\ \midrule
2023 & arXiv & \textbf{Emu} \cite{dai2023emu}: Quality-tuning to effectively guide a pretrained model. \\ 
     & NeurIPS & \textbf{CM3Leon} \cite{yu2023scaling}: The first multimodal model trained using a recipe adapted from text-specific language models. \\ \midrule
2024 & arXiv & \textbf{MARS} \cite{he2024mars}: Retains NLP capabilities of LLMs while endowing specialized visual understanding. \\ 
     & CVPR & \textbf{Unified-IO 2} \cite{lu2024unified}: The first model trained on a diverse and extensive skill set, unifying three generative abilities. \\ 
     & arXiv & \textbf{STAR} \cite{ma2024star}: Employs a scale-wise autoregressive paradigm for text-to-image modeling. \\ 
     & arXiv & \textbf{HART} \cite{tang2024hart}: Introduces a hybrid tokenizer for efficient high-resolution visual synthesis. \\ 
     & arXiv & \textbf{SJD+AR} \cite{teng2024accelerating}: Integrating Jacobi to accelerate the autoregressive model. \\ 
     & NeurIPS & \textbf{GILL} \cite{koh2024generating}: Combining frozen large language models (LLMs) with pretrained image encoder and decoder models. \\ 
     & arXiv & \textbf{DART} \cite{gu2024dart}: Utilizes a denoising autoregressive encoder. \\ \midrule
2025 & arXiv & \textbf{NextStep-1} \cite{team2025nextstep}: Pairs autoregressive modeling of continuous image tokens with a lightweight flow-matching head. \\
     & \tht{l}{CVPR/ICLR} & \textbf{Unified AR systems} \cite{wu2024janus,xie2024showo,zhou2024transfusion}: (see Section~\ref{sec:unified}). \\ \bottomrule
\end{tabular}}
    \vspace{-0.4cm}
\end{table*}

As noted by OpenAI, diffusion models have significant advantages in image generation \cite{dhariwal2021diffusion}. In recent years, diffusion models have rapidly gained prominence in the field of T2I, sparking widespread interest and inspiring new research that reflects the enthusiasm of the scientific community. The development of Denoising Diffusion Probabilistic Models (DDPM) in T2I is summarized in Table \ref{developddpm}. With the emergence of various works \cite{gu2022vector,ramesh2022hierarchical,saharia2022photorealistic,rombach2022high,betker2023improving,ruiz2023dreambooth,zhang2023adding,zhang2023adding,podell2023sdxl}, DDPM has demonstrated its fresh vitality in T2I. The most recent systems make the component-level shift discussed in Section~\ref{sec:shared-components} concrete: Qwen-Image \cite{wu2025qwenimage} combines a vision-language condition encoder, an MMDiT-style backbone, and a flow-matching objective; FLUX.1 Kontext \cite{labs2025flux} unifies in-context generation and editing; and Qwen-Image-2.0-RL \cite{xu2026qwenimagerl} adds reward-model-based post-training on top of such a backbone, illustrating how post-training (Section~\ref{section4}) has become part of the diffusion/flow development line itself.

More details of the development of GAN, AR and DDPM in T2I are provided in Appendix~\ref{section3back}.

\begin{table*}[ht]
\centering
\caption{Overview of DM models in T2I by year.}
\label{developddpm}
			\resizebox{1\linewidth}{!}{%
\begin{tabular}{@{}l|l|l@{}}
\toprule
\textbf{Year} & \textbf{Source} & \textbf{Model and Description} \\ \midrule
2021 & arXiv & \textbf{GLIDE} \cite{nichol2021glide}: The first diffusion work in T2I. \\ \midrule
2022 & CVPR & \textbf{VQ-Diffusion} \cite{gu2022vector}: Based on a vector quantization diffusion model. \\ 
     & NeurIPS & \textbf{Imagen} \cite{saharia2022photorealistic}: Combines Transformer language models and diffusion processes. \\ 
     & CVPR & \textbf{Stable Diffusion} \cite{rombach2022high}: Applies the diffusion model to the latent space of a powerful pretrained autoencoder. \\ 
     & arXiv & \textbf{DALLE-2} \cite{ramesh2022hierarchical}: A two-stage model generating CLIP image embeddings based on text. \\ 
     & CVPR & \textbf{Blended Diffusion} \cite{avrahami2022blended}: Using natural language guidance to produce realistic and diverse images. \\ \midrule
2023 & arXiv & \textbf{DALLE-3} \cite{betker2023improving}: Using diffusion based on DALLLE-2 and using GPT-3. \\ 
     & CVPR & \textbf{Dreambooth} \cite{ruiz2023dreambooth}: Fine-tunes a pretrained T2I model using a few images of an object. \\ 
     & ICCV & \textbf{ControlNet} \cite{zhang2023adding}: Adds spatial conditional control to a large pretrained text-to-image diffusion model. \\ 
     & CVPR & \textbf{Specialist Diffusion} \cite{lu2023specialist}: Handles multiple streams of text-to-image and image-to-text within a unified model. \\ 
     & ICML & \textbf{UniDiffuser} \cite{bao2023one}: Perturbs data in all modalities instead of a single modality. \\  
     & ICCV & \textbf{Versatile Diffusion} \cite{xu2023versatile}: Handles multiple flows of text-to-image, image-to-text, and variations in one unified model. \\ 
     & CVPR & \textbf{Custom Diffusion} \cite{kumari2023multi}:  Optimizing a few parameters in the text-to-image conditioning mechanism. \\ 
     & ICCV & \textbf{SVDiff} \cite{han2023svdiff}: Fine-tunes singular values of weight matrices for a compact parameter space. \\ 
     & CVPR & \textbf{ERNIE-ViLG 2} \cite{feng2023ernie}: The first large T2I model in the Chinese field. \\ 
     & CVPR & \textbf{Corgi} \cite{zhou2023shifted}: Based on a shifted diffusion model. \\ \midrule
2024 & arXiv & \textbf{CogView3} \cite{zheng2024cogview3}: Implementing relay diffusion in text-to-image generation. \\ 
     & ICML & \textbf{Stable Diffusion 3} \cite{esser2024scaling}: Uses different weights for modalities, enabling bidirectional information flow. \\ 
     & ICLR & \textbf{Würstchen} \cite{pernias2023wurstchen}: Develops a latent diffusion technique using a compact semantic image representation. \\ 
     & ICLR & \textbf{Stable Diffusion XL} \cite{podell2023sdxl}: Utilizes a three times larger UNet backbone network. \\ 
     & ICLR & \textbf{PixArt-$\alpha$} \cite{chen2023pixart}: By 1) Training strategy decomposition. 2) Efficient T2I Transformer. 3) High-informative data. \\
     & CVPR & \textbf{Ranni} \cite{feng2024ranni}: Using a semantic panel as the middleware and incorporating visual concepts parsed by LLM. \\ 
     & ICML & \textbf{RPG} \cite{yang2024mastering}: Integrates text-guided image generation and editing through MLLMoRPG. \\ 
     & NeurIPS & \textbf{ConPreDiff} \cite{yang2024improving}: Improves diffusion-based image synthesis through context prediction. \\ 
     & CVPR & \textbf{DistriFusion} \cite{li2024distrifusion}: Uses parallelism across multiple GPUs for integration into computational pipelines. \\ 
     & CVPR & \textbf{InstanceDiffusion} \cite{wang2024instancediffusion}: Increases precise instance-level control by UniFusion, ScaleU and Multi-instance Sampler blocks. \\ 
     & CVPR & \textbf{Instruct-Imagen} \cite{hu2024instruct}: Handling heterogeneous image generation tasks by multi-modal instruction. \\ 
     & CVPR & \textbf{ElasticDiffusion} \cite{haji2024elasticdiffusion}: Decouple the generation trajectory of a pretrained model into local and global signals. \\ 
     & arXiv & \textbf{PIXART-$\delta$} \cite{chen2024pixart1}: Integration of LCM significantly accelerates inference speed. \\ 
     & arXiv & \textbf{PixArt-$\Sigma$} \cite{chen2024pixart2}: Evolves from the weaker' baseline to a stronger' model via incorporating higher quality data. \\ 
     & arXiv & \textbf{SD3-Turbo} \cite{sauer2024fast}: Introduces latent adversarial diffusion distillation (LADD). \\ 
     & arXiv & \textbf{StreamMultiDiffusion} \cite{lee2024streammultidiffusion}: The first real-time region-based text-to-image generation framework. \\ 
     & arXiv & \textbf{OmniGen} \cite{xiao2024omnigen}: Unifying image generation architecture, simplifying user workflows, enabling effective knowledge transfer. \\ 
     & arXiv & \textbf{Hunyuan-DiT} \cite{li2024hunyuan}: Using carefully designed transformer architecture and a MLLM.  \\ 
     & CVPR & \textbf{DragDiffusion} \cite{shi2024dragdiffusion}: Enhances interactive point-based editing for real and diffusion-generated images. \\ 
     & ICLR & \textbf{ContextDiff} \cite{yang2024cross}: Incorporating cross-modal context into both the forward and reverse diffusion processes. \\ \midrule
2025 & arXiv & \textbf{Qwen-Image} \cite{wu2025qwenimage}: Uses a vision-language condition encoder, an MMDiT-style backbone, and a flow-matching objective. \\
     & arXiv & \textbf{FLUX.1 Kontext} \cite{labs2025flux}: Unifies in-context image generation and editing with combined text--image context. \\
2026 & arXiv & \textbf{Qwen-Image-2.0-RL} \cite{xu2026qwenimagerl}: GRPO-style optimization, and on-policy distillation to generation and editing. \\ \bottomrule
\end{tabular}}
    \vspace{-0.4cm}
\end{table*}

\subsection{Unified Multimodal Understanding and Generation}
\label{sec:unified}
A growing body of work builds \emph{unified} models that place image generation and visual understanding (and often text generation) inside a single framework. It is useful to distinguish three settings that are easily conflated: (i) unifying generation and editing within one image model (e.g., FLUX.1 Kontext, Section~\ref{section4}); (ii) using an external vision-language model only to plan or rewrite prompts; and (iii) genuinely unifying understanding and generation in one model, which is the subject of this subsection. This direction is a natural extension of the autoregressive line in Table~\ref{developAR} rather than a separate field, and we therefore keep it inside the T2I narrative.

Early systems such as Unified-IO \cite{lu2022unifiedio} and CM3Leon \cite{yu2023scaling} connect image generation with a broad set of vision-language tasks through a shared token interface. More recent designs differ mainly in \emph{how} the image side is represented and trained. Janus \cite{wu2024janus} decouples the visual encoders used for understanding and for generation within one autoregressive model; Show-o \cite{xie2024showo} combines autoregressive text modeling with discrete diffusion on the image side, whereas Transfusion \cite{zhou2024transfusion} combines next-token prediction with continuous image diffusion in a single Transformer. BAGEL \cite{deng2025emerging} adopts a mixture-of-transformers design for unified multimodal pretraining, and Harmon \cite{wu2025harmonizing} shares a masked-autoregressive visual representation across understanding and generation. Table~\ref{tab:unified} summarizes these design axes; crucially, a shared Transformer does \emph{not} imply that the visual encoder, representation, and objective are shared, so a single ``AR-plus-diffusion'' label is misleading unless it states which modality uses which objective.

Two cautions follow. First, joint task support is not evidence of cross-task \emph{benefit}: separate competence in understanding and in generation must be distinguished from measurable mutual improvement. Benchmarks such as MME-Unify \cite{xie2025mmeunify} are designed to probe mixed-modality interaction rather than reporting understanding and generation scores in isolation, and weakness-targeted post-training of unified models \cite{chen2026unified} studies whether textual reasoning can be converted into better image generation. Second, a unified model is not automatically an autonomous multimodal agent; the reasoning-and-feedback outlook is discussed in Section~\ref{section7}.

\begin{table*}[t]
\centering\small
\caption{Representative unified understanding-and-generation models. The table records architecture choices, not a cross-paper performance ranking obtained under different protocols.}
\label{tab:unified}
\begin{tabular}{p{0.15\textwidth}p{0.24\textwidth}p{0.20\textwidth}p{0.29\textwidth}}
\toprule
Model & Visual representation / sharing & Image-generation objective & Main survey interpretation \\
\midrule
Unified-IO \cite{lu2022unifiedio} & Unified task/token interface & Token generation & Connects image generation with a broad set of vision-language tasks. \\
CM3Leon \cite{yu2023scaling} & Token interface, LM recipe & Autoregressive tokens & Shows a text-LM training recipe transfers to multimodal generation. \\
Janus \cite{wu2024janus} & Decoupled visual encoders & Autoregressive generation & A unified framework need not share one visual encoder. \\
Show-o \cite{xie2024showo} & Unified Transformer & Discrete diffusion & Image-side discrete diffusion differs from continuous latent diffusion. \\
Transfusion \cite{zhou2024transfusion} & Mixed-modal shared model & Continuous diffusion & Language and image prediction use different objectives in one model. \\
BAGEL \cite{deng2025emerging} & Mixture of Transformers & Flow-based generation & Joint task support must be separated from evidence of cross-task benefit. \\
Harmon \cite{wu2025harmonizing} & Shared MAR representation & Masked image generation & Representation sharing targets both semantics and synthesis. \\
\bottomrule
\end{tabular}
    \vspace{-0.4cm}
\end{table*}

\subsection{Recent Research Directions, Datasets and Evaluation Metrics}
Research on T2I generation not only focuses on improving the performance metrics of various models \cite{hu2021lora,mokady2023null} but also requires deeper exploration of more specific topics. This section discusses key research directions in T2I based on the latest findings, including but not limited to controllable T2I generation \cite{zhang2023adding,mou2024t2i}, personalized image generation \cite{gal2022image,ruiz2023dreambooth}, consistency issues in generation \cite{chatterjee2025getting,jiang2024comat,wang2023tokencompose,phung2024grounded}, safety concerns \cite{li2024self,lyu2024one}, copyright protection \cite{liang2023adversarial,wang2024simac}, incorporating human feedback \cite{liang2024rich,zhao2024diffagent}, style transfer \cite{wu2021styleformer,wu2023uncovering}, fine-grained generation control \cite{cheng2024learning,patashnik2023localizing}, and the supporting role of large language models (LLMs) in T2I \cite{qu2023layoutllm,huang2024smartedit}. As shown in Table \ref{research}, these studies not only enhance the performance of generative models, but also provide greater flexibility and safety for practical applications. 

These research directions are discussed in detail in Section~\ref{section4}.

\begin{table*}[!ht]
\centering
\caption{Overview of T2I research directions.}
\label{research}
\resizebox{1\linewidth}{!}{
\begin{tabular}{@{}ll@{}}
\toprule
\textbf{Research Area} & \textbf{Model} \\ \midrule

Sec.\ref{Alignment}: Alignment and Human Feedback & RAHF\cite{liang2024rich}, DifffAgent\cite{zhao2024diffagent}, HPS\cite{wu2023human}, ImageReward\cite{xu2024imagereward}, Multi-dimensional Human \\ & Preference (MHP)\cite{zhang2024learning}, MPS\cite{zhang2024learning} \\ \midrule

Sec.\ref{Personalized}: Personalized Generation & Textual Inversion\cite{gal2022image}, PIA\cite{zhang2024pia}, Prompt-Free Diffusion\cite{xu2024prompt}, ADI\cite{huang2024learning}, HyperDreamBooth\cite{ruiz2024hyperdreambooth}, \\ 
& DreamBooth\cite{ruiz2023dreambooth}, Personalized Residuals\cite{ham2024personalized}, PALP\cite{arar2024palp}, PhotoMaker\cite{li2024photomaker}, DreamTurner\cite{hua2023dreamtuner}, \\ 
& ELITE\cite{wei2023elite}, Tailored Visions\cite{chen2024tailored}, RealCustom\cite{huang2024realcustom}, SAG\cite{chan2024improving}, CoRe\cite{wu2024core}, Imagine Yourself\cite{he2024imagine}, \\ 
& FastComposer\cite{xiao2024fastcomposer}, InstantID\cite{wang2024instantid}, Specialist Diffusion\cite{lu2023specialist}, ControlStyle\cite{chen2023controlstyle}, UniPortrait\cite{he2024uniportrait}, \\ 
& TIGC\cite{li2025tuning}, DETEX\cite{cai2024decoupled}, FlashFace\cite{zhang2024flashface}, IDAdapter\cite{cui2024idadapter}, SSR-Encoder \cite{zhang2024ssr} \\ \midrule

Sec.\ref{Controllable}: Controllable T2I Generation & ControlNet\cite{zhang2023adding}, T2I-Adapter\cite{mou2024t2i}, Custom Diffusion\cite{kumari2023multi}, SID\cite{kim2024selectively}, JeDi \cite{zeng2024jedi}, ViCo\cite{hao2023vico}, \\ & DreamMatcher\cite{nam2024dreammatcher}, RealCustom\cite{huang2024realcustom}, StyleGAN Meets Stable Diffusion\cite{li2024stylegan}, Uni-ControlNet\cite{zhao2024uni}, \\ 
& BLIP-Diffusion\cite{li2024blip}, LayerDiffusion\cite{li2023layerdiffusion}, ReCo\cite{yang2023reco}, SpaText\cite{avrahami2023spatext}, Codi\cite{mei2024codi}, StyleTokenizer\cite{li2024styletokenizer}, \\ 
& P+\cite{voynov2023p+}, IP-Adapter\cite{ye2023ip}, ResAdapter\cite{cheng2024resadapter}, DetDiffusion\cite{wang2024detdiffusion}, CAN\cite{cai2024condition}, SceneDiffusion\cite{ren2024move}, \\ & Zero-Painter\cite{ohanyan2024zero}, FreeControl\cite{mo2024freecontrol}, PCDMs\cite{shen2023advancing},  ControlNet++\cite{li2025controlnet}, ControlNeXt\cite{peng2024controlnext}, \\ & Composer\cite{xiao2024fastcomposer}, MultiDiffusion\cite{bar2023multidiffusion} \\ \midrule

Sec.\ref{Transfer}: T2I Style Transfer  & Styleformer\cite{wu2021styleformer}, InST\cite{zhang2023inversion}, ArtAdapt\cite{chen2024artadapter}, InstantBooth\cite{shi2024instantbooth}, OSASIS\cite{cho2024one}, DEADiff\cite{qi2024deadiff}, \\ & Uncovering the Disentanglement Capability\cite{wu2023uncovering}, InstantStyle\cite{wang2024instantstyle}, StyleAligned\cite{hertz2024style}, FreeCustom\cite{ding2024freecustom} \\ \midrule

Sec.\ref{Text-Guided}: Text-guided Image Generation  & DisenDiff\cite{zhang2024attention}, Predicated Diffusion\cite{sueyoshi2024predicated}, LEDITS++\cite{brack2024ledits++}, PH2P\cite{mahajan2024prompting}, GSN\cite{chefer2023attend},  NMG\cite{cho2024noise}, \\ & Imagic\cite{kawar2023imagic}, SINE\cite{zhang2023sine}, AdapEdit\cite{ma2024adapedit}, FISEdit\cite{yu2024accelerating}, BARET\cite{qiao2024baret}, D-Edit\cite{feng2024item}, PromptCharm\cite{wang2024promptcharm}, \\&  RAPHAEL\cite{xue2024raphael}, SUR-Adapter\cite{zhong2023adapter}, Prompt Diffusion\cite{wang2023context}, SDG\cite{liu2023more}, InfEdit\cite{xu2023inversion}, FPE\cite{liu2024towards}, \\ 
& FoI\cite{guo2024focus}, DiffEditor\cite{mou2024diffeditor}, Prompt Augmentation\cite{bodur2024prompt}, TurboEdit\cite{wu2024turboedit}, PPE \cite{xu2022predict}, DE-Net \cite{tao2023net} \\ \midrule

Sec.\ref{Performance}: Performance and  Effectiveness & LowRank Adaptation (LoRA)\cite{hu2021lora}, ECLIPSE\cite{patel2024eclipse}, InstaFlow\cite{liu2023instaflow}, SVDiff\cite{han2023svdiff}, LinFusion\cite{liu2024linfusion}, \\ 
& Speculative Jacobi Decoding\cite{teng2024accelerating}, Token-Critic\cite{lezama2022improved}, Null-text Inversion\cite{mokady2023null}, DiffFit\cite{xie2023difffit},  \\ 
& Progressive Distillation\cite{salimans2022progressive}, YOSO\cite{luo2024you}, SwiftBrush\cite{nguyen2024swiftbrush}, StreamDiffusion\cite{kodaira2023streamdiffusion}, TiNO-Edit\cite{chen2024tino}, \\ & On the Scalability of Diffusion-based T2I Generation\cite{li2024scalability},Muse\cite{chang2023muse}, UFC-BERT\cite{zhang2021ufc}, MarkovGen\cite{jayasumana2024markovgen} \\ \midrule

Sec.\ref{Reward}: Reward Mechanism  & Prompt AutoEditing (PAE)\cite{mo2024dynamic}, AIG\cite{jena2024elucidating}, IterComp\cite{zhang2024itercomp}, Powerful and Flexible\cite{wei2024powerful}, \\ 
& Optimizing Prompts for T2I Generation\cite{hao2024optimizing}, ImageReward\cite{xu2024imagereward}, SPIN-Diffusion\cite{yuan2024self}, \\  \midrule

Sec.\ref{Safety}: Safety Issues & Self-Discovering\cite{li2024self}, POSI\cite{wu2024universal}, Detection-based Method \cite{rando2022red}, SteerDiff\cite{zhang2024steerdiff}, MetaCloak\cite{liu2024metacloak}, \\ & OpenBias\cite{d2024openbias}, SPM\cite{lyu2024one}, Guidance-based Methods \cite{wu2024universal,rombach2022high,schramowski2023safe}, Fine-tuning-based Method \cite{gandikota2023erasing}, \\ 
& HFI\cite{kim2024safeguard}, RECE\cite{gong2024reliable}, RIATIG\cite{liu2023riatig} \\ \midrule

Sec.\ref{Copyright}: Copyright Protection Measures & AdvDM\cite{liang2023adversarial}, Mist\cite{liang2023mist}, Anti-DreamBooth\cite{van2023anti}, InMark\cite{liu2024countering}, VA3\cite{li2024va3}, SimAC\cite{wang2024simac} \\ \midrule

Sec.\ref{Consistency}: Consistency Between  & TokenCompose\cite{wang2023tokencompose}, Attention Refocusing\cite{phung2024grounded}, DPT\cite{qu2024discriminative}, MIGC\cite{zhou2024migc}, INITNO\cite{guo2024initno},  \\ 
Text and Image Content  &  DiffusionCLIP\cite{kim2022diffusionclip}, SDEdit\cite{meng2021sdedit}, MasaCtrl\cite{cao2023masactrl}, RIE\cite{liu2024referring}, Scene TextGen\cite{zhangli2024layout}, \\ & AnyText\cite{tuo2023anytext}, SceneGenie\cite{farshad2023scenegenie}, ZestGuide\cite{couairon2023zero}, Adma-GAN\cite{wu2022adma}, AtHom\cite{shi2022athom}, TL-Norm\cite{chen2022background}, \\ & GSN\cite{chefer2023attend}, DiffPNG\cite{yang2024exploring}, Compose and Conquer\cite{lee2024compose}, StoryDiffusion\cite{zhou2024storydiffusion}, SynGen\cite{rassin2024linguistic}, \\ & ParaDiffusion\cite{wu2023paragraph}, Token Merging (Tome)\cite{hu2024token}, Training-Free Structured Diffusion Guidance\cite{feng2022training}, \\ & StyleT2I\cite{li2022stylet2i}, Bounded Attention\cite{dahary2024yourself}, DAC\cite{song2024doubly}, Attention Map Control\cite{wang2024compositional}, DDS\cite{hertz2023delta}, \\ & CDS\cite{nam2024contrastive}, StableDrag\cite{cui2024stabledrag}, FreeDrag\cite{ling2024freedrag}, RegionDrag\cite{lu2024regiondrag} \\ \midrule

Sec.\ref{Spatial}: Spatial Consistency & SPRIGHT\cite{chatterjee2025getting}, CoMat\cite{jiang2024comat}, CLIP\cite{radford2021learning}, MULAN\cite{tudosiu2024mulan}, TCTIG\cite{yan2022trace}, PLACE\cite{lv2024place},  \\ 
Between Text and Image  & Backward Guidance\cite{chen2024training}, SSMG\cite{jia2024ssmg} \\ \midrule

Sec.\ref{Specific}: Specific Content Generation & CosmicMan\cite{li2024cosmicman}, Text2Human\cite{jiang2022text2human}, CAFE\cite{zhou2024customization}, ZS-SBIR\cite{lin2023zero}, HanDiffuser\cite{narasimhaswamy2024handiffuser}, EmoGen\cite{yang2024emogen}, \\ 
& Cross Initialization for Face Personalization\cite{pang2023cross}, HcP\cite{wang2024towards}, PanFusion\cite{zhang2024taming}, StoryGen\cite{liu2024intelligent}, \\ & Text2Street\cite{su2024text2street}, SVGDreamer\cite{xing2024svgdreamer}, Face-Adapter\cite{han2024face}, StoryDALL-E\cite{maharana2022storydall}, TexFit\cite{wang2024texfit}, EditWorld\cite{yang2024editworld} \\ \midrule

Sec.\ref{Fine-Grained}: Fine-grained  & Continuous 3D Words\cite{cheng2024learning}, FiVA\cite{wufiva},  GLIGEN\cite{li2023gligen}, InteractDiffusion\cite{hoe2024interactdiffusion}, PreciseControl \cite{parihar2025precisecontrol},  \\ 
Control in Generation & Localizing Object-level Shape Variations\cite{patashnik2023localizing}, Motion Guidance\cite{geng2024motion}, SingDiffusion\cite{zhang2024tackling}, CFLD\cite{lu2024coarse},   \\ 
& NoiseCollage\cite{shirakawa2024noisecollage}, Concept Weaver\cite{kwon2024concept}, Playground v2.5\cite{li2024playground}, LayerDiffuse\cite{zhang2024transparent}, ConForm\cite{meral2024conform} \\ \midrule

Sec.\ref{LLM-Assisted}: LLM-assisted T2I & LayoutGPT\cite{feng2024layoutgpt}, DiagrammerGPT\cite{zala2023diagrammergpt}, AutoStory\cite{wang2023autostory}, CAFE\cite{zhou2024customization}, LayoutLLM-T2I\cite{qu2023layoutllm}, MoMA\cite{song2024moma},  \\ 
& SmartEdit\cite{huang2024smartedit}, RFNet\cite{yao2025fabrication}, TIAC\cite{liao2024text}, RPG\cite{yang2024mastering}, ELLA\cite{hu2024ella}, DialogGen\cite{huang2024dialoggen}, MGIE\cite{fu2023guiding}\\ \midrule

Recent advances (2025--2026) & \textit{Alignment/post-training:} Flow-GRPO\cite{liu2025flowgrpo}, DenseGRPO\cite{deng2026densegrpo}, Qwen-Image-2.0-RL\cite{xu2026qwenimagerl}, PE-OPSD\cite{lin2026peopsd},\\ &
CHATS\cite{fu2025chats}; \textit{Editing:} Step1X-Edit\cite{liu2025stepxedit}, In-Context Edit\cite{zhang2025incontext}, ReFlex\cite{kim2025reflex}, FLUX.1 Kontext\cite{labs2025flux}; \\ &
\textit{Personalization:} AnyStory\cite{he2025anystory}, XVerse\cite{chen2025xverse}; \textit{Safety:} OVERT\cite{cheng2025overt}, Fine-Grained Erasure\cite{thakral2025finegrained}; \\ &\textit{Evaluation:} MME-Unify\cite{xie2025mmeunify}, FineGRAIN\cite{hayes2025finegrain}, Qwen-Image-Bench\cite{li2026qwenimagebench}, LongT2IBench\cite{yang2025longtibench} \\

\bottomrule
\end{tabular}}
    \vspace{-0.4cm}
\end{table*}

Datasets and evaluation are central to T2I, since enhancing generation capabilities requires large amounts of data \cite{brock2018large,karras2020analyzing} and reliable ways to measure progress \cite{hartwig2024evaluating}. The T2I literature draws on image-caption corpora, web-scale image-text pairs, preference/feedback data, and task-specific prompt or benchmark sets. We give the detailed treatment---including dataset descriptions, data augmentation, the full list of evaluation metrics, and recent 2025--2026 benchmarks---in Section~\ref{section5}; Table~\ref{dataset} and Table~\ref{table:evaluation} there summarize the datasets and metrics, respectively.

Evaluation must jointly consider image fidelity, text-image alignment, diversity, and, more recently, compositionality, editing, and safety, yet no single metric captures all of these dimensions.
\section{Recent Research Directions in T2I}
\label{section4}

\subsection{Alignment and Human Feedback}
\label{Alignment}
In the field of T2I research, scholars face a range of challenges, including artifacts in generated images, inconsistencies with text descriptions, and deficiencies in aesthetic quality. Ensuring alignment between generated images and text descriptions, as well as with human cognition, has consistently been a key area of focus in T2I research \cite{grimal2024tiam}.

The ImageReward model \cite{xu2024imagereward}, as the first general-purpose human preference reward model for T2I, effectively encodes human preferences and supports Reinforcement Learning from Human Feedback (ReFL). ReFL is a direct tuning algorithm that has shown advantages over other methods through both automatic and human evaluations.

Inspired by the success of human feedback learning \cite{bai2022training} in large language models, Liang et al. \cite{liang2024rich} collected extensive human feedback and designed RAHF, a multimodal Transformer model. As illustrated in Figure \ref{rahf}, RAHF uses self-attention mechanisms to understand and predict human evaluations of generated images, significantly improving its ability to identify inconsistencies between text and images.

DiffAgent \cite{zhao2024diffagent}, an LLM-based model agent, adopts a novel two-stage training framework (SFTA), which enables it to accurately align T2I API responses with user input based on human preferences.

In terms of the impact of datasets on alignment research, HPS (Human Preference Score) \cite{wu2023human} introduced the HPD dataset, which focuses on images generated by stable diffusion models. This dataset uses human preference annotations to train the HPS model to better align with human preferences. HPS v2 \cite{wu2023human2} further expanded the dataset to include eight generative models and incorporated titles from the COCO \cite{lin2014microsoft} dataset. Although HPS v2 primarily addresses overall human preferences, it does not yet fully account for the diversity of individual tastes.

Multidimensional Human Preference (MHP) dataset \cite{zhang2024learning} considers a wider range of human preference dimensions, with human annotators providing multidimensional labels for image pairs based on aesthetics, detail quality, semantic alignment, and overall scores. Furthermore, its MPS model aims to predict scores under different preference conditions and establishes a benchmark to evaluate existing T2I generation models.

Beyond reward modeling, a rapidly growing line of work turns human (and AI) feedback into explicit post-training, which can be grouped into three families. (i) \emph{Reward fine-tuning} optimizes a differentiable preference reward such as ImageReward directly, as in ReFL. (ii) \emph{Reinforcement learning} treats denoising as a sequential decision process: DDPO \cite{black2023training} and DPOK \cite{fan2023dpok} fine-tune diffusion models with policy gradients, whereas Diffusion-DPO \cite{wallace2023diffusion} adapts direct preference optimization to avoid training an explicit reward model. (iii) \emph{AI feedback} replaces part of the human labeling: the RLAIF study \cite{lee2023rlaif} provides conceptual evidence from language tasks, while AGFSync \cite{an2024agfsync} supplies T2I-specific AI-generated preference data. More recent work extends these ideas to flow-matching models, whose deterministic sampling does not directly admit policy-gradient exploration: Flow-GRPO \cite{liu2025flowgrpo} converts the sampler into an equivalent stochastic form, and DenseGRPO \cite{deng2026densegrpo} studies denser reward signals for flow-model alignment. At the system level, Qwen-Image-2.0-RL \cite{xu2026qwenimagerl} combines task-specific reward models, GRPO-style optimization, and on-policy distillation for both generation and editing, while PE-OPSD \cite{lin2026peopsd} internalizes prompt-enhancer behavior through on-policy self-distillation so that a model can be deployed directly with raw prompts. Because these methods optimize different feedback signals, a higher reward alone is not a sufficient basis for comparison: reward hacking \cite{hong2026understanding} and preference mode collapse \cite{chen2025taming} motivate reporting image quality, diversity, and prompt generalization under independent, non-optimized judges, and test-time strategies such as CHATS \cite{fu2025chats} should be distinguished from training-time alignment because their sampling and compute budgets differ.

\subsection{Personalized Generation}
\label{Personalized}

\begin{figure*}[!t]    
	\centering
    \includegraphics[width=0.9\linewidth]{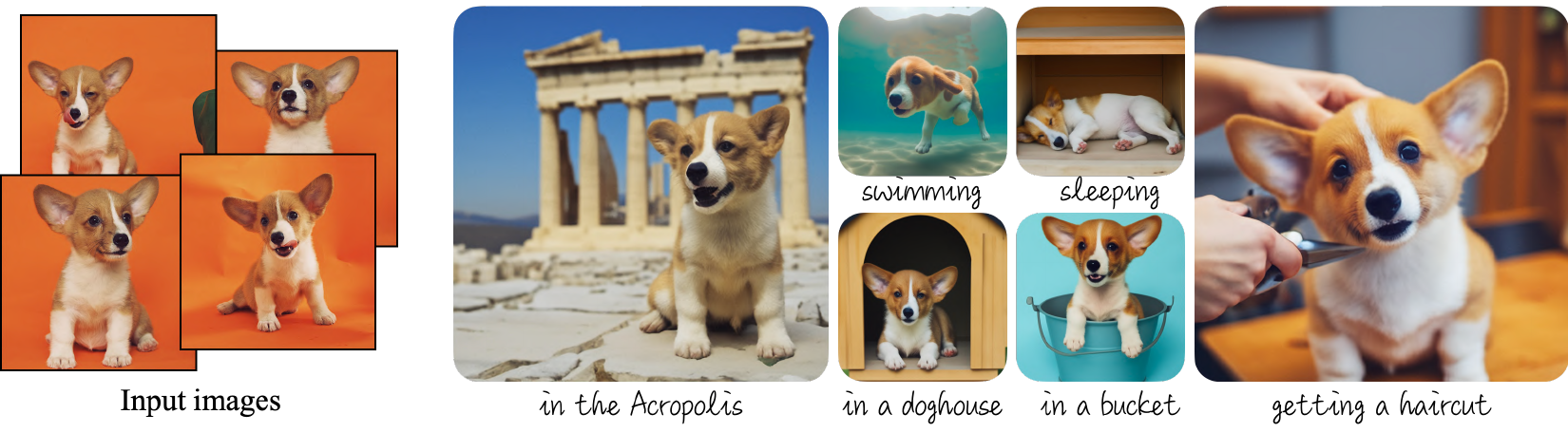} 
	\caption{With just a few images (typically 3-5) of a subject (left), DreamBooth\cite{ruiz2023dreambooth}, the AI-powered photo booth, can generate a myriad of images of the subject in different contexts (right), using the guidance of a text prompt. The results exhibit natural interactions with the environment, as well as novel articulations and variation in lighting conditions, all while maintaining high fidelity to the key visual features of the subject.}  
	\label{Dreambooth}
        \vspace{-0.4cm}
\end{figure*}

Personalized T2I generation refers to the process of creating images based on specific user needs, preferences, or personal characteristics. In 2022, the DreamBooth model \cite{ruiz2023dreambooth}, developed by the Google research team, was widely recognized as a landmark achievement in this field, with its personalized generation capabilities illustrated in Figure \ref{Dreambooth}.

DreamBooth fine-tunes diffusion models by incorporating custom subjects, focusing on the ability to generate images of specific entities while maintaining consistency across different scenes, poses, and perspectives. This approach not only expands the model's language-vision dictionary, enhancing the association between new terms and subjects, but also effectively addresses common issues of overfitting and mode collapse typically encountered in traditional fine-tuning methods.
\begin{equation}
\begin{aligned}
\mathbb{E}_{x,c,\epsilon,\epsilon',t} [ & w_t \| \hat{x}_{\theta}(\alpha_t x + \sigma_t \epsilon, c) - x \|_2^2  \\
& \quad + \lambda w_{t'} \| \hat{x}_{\theta}(\alpha_{t'} x_{pr} + \sigma_{t'} \epsilon', c_{pr}) - x_{pr} \|_2^2 ].
\end{aligned}
\label{eq2}
\end{equation}

\begin{figure}[!t] 
	\centering
    \includegraphics[width=1\linewidth]{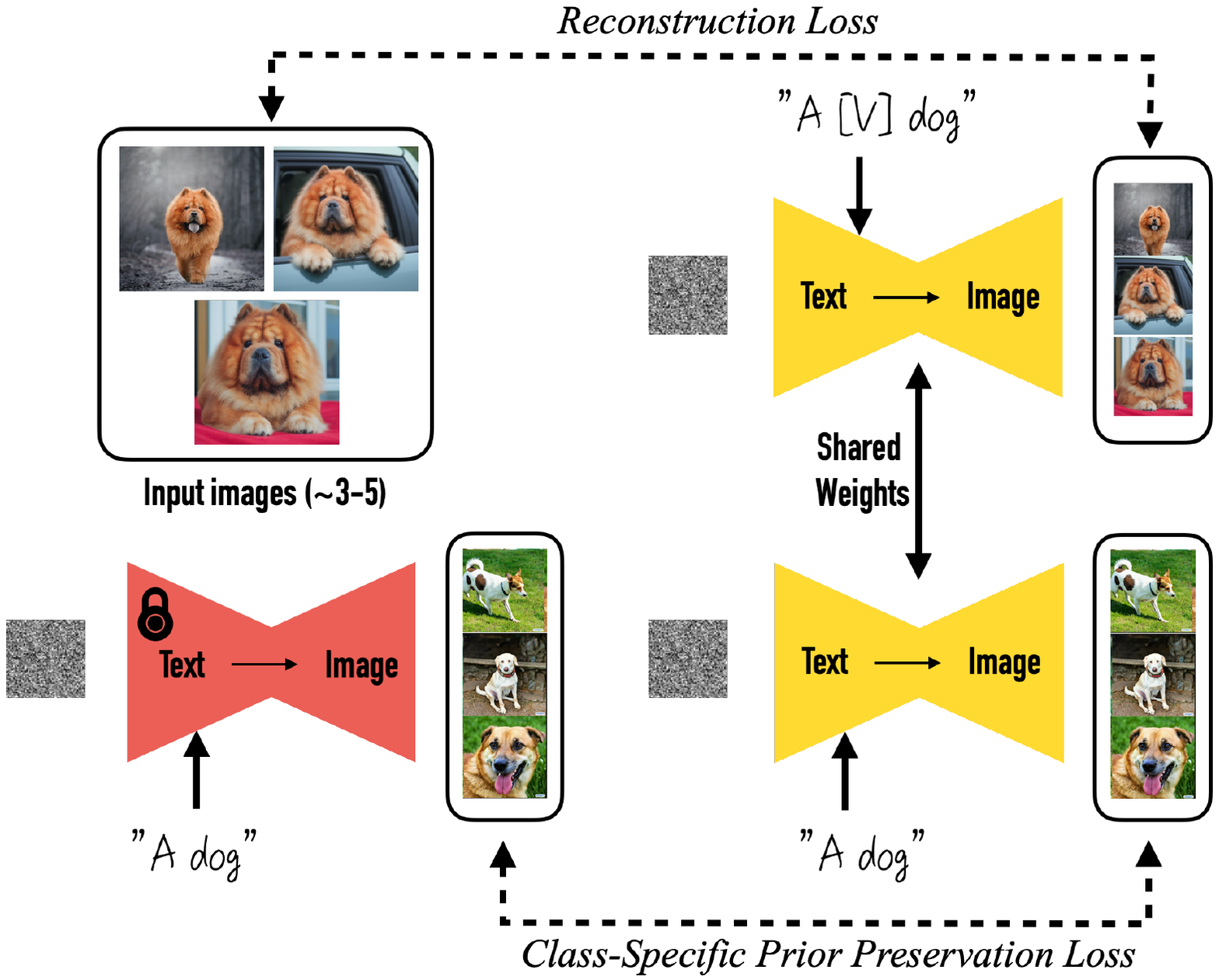} 
	\caption{Fine-tuning of DreamBooth\cite{ruiz2023dreambooth}. Given approximately 3-5 images of a subject, we fine-tune a text-to-image diffusion model with the input images paired with a text prompt containing a unique identifier and the name of the class the subject belongs to (e.g., ``A [V] dog''). In parallel, we apply a class-specific prior preservation loss, which leverages the semantic prior that the model has on the class and encourages it to generate diverse instances belonging to the subject's class using the class name in a text prompt (e.g., ``A dog''). }  
	\label{Dreamboothmethod}
        \vspace{-0.4cm}
\end{figure}

As illustrated in Figure \ref{Dreamboothmethod}, the model uses rare identifier tokens to represent given subjects and gradually optimizes the generation results through a staged fine-tuning process. Additionally, the proposed ``autogenous class-specific prior preservation loss'' function plays a crucial role in balancing the discrepancies between real images and the generated images of the specified subjects, as shown in Eq. \eqref{eq2}.

Complementing DreamBooth is Textual Inversion \cite{gal2022image}, which addresses the challenge of introducing new concepts in large models by learning new ``words'' to represent user-provided objects or styles, enabling personalized creation. Furthermore, PIA \cite{zhang2024pia} provides a flexible solution for personalized image animation, ensuring controllable movement between images and text while integrating various personalized models.

Prompt-Free Diffusion \cite{xu2024prompt} improves image generation quality by eliminating the need for prompts, thereby simplifying the personalized generation process. PALP \cite{arar2024palp} maintains consistency between text and generated images, even under complex prompting conditions, demonstrating excellent performance. PhotoMaker \cite{li2024photomaker} efficiently encodes the input ID images using stacked ID embeddings, ensuring the retention of identity information.

Furthermore, ADI \cite{huang2024learning} introduces disentangled identifiers, which enable the generation of new images with shared actions through specific symbols, enhancing the diversity of output. DreamTurner \cite{hua2023dreamtuner} achieves theme-driven image generation through a theme encoder, allowing for more detailed preservation of subject identity.

\begin{figure*}[t]  
	\centering
    \includegraphics[width=0.9\linewidth]{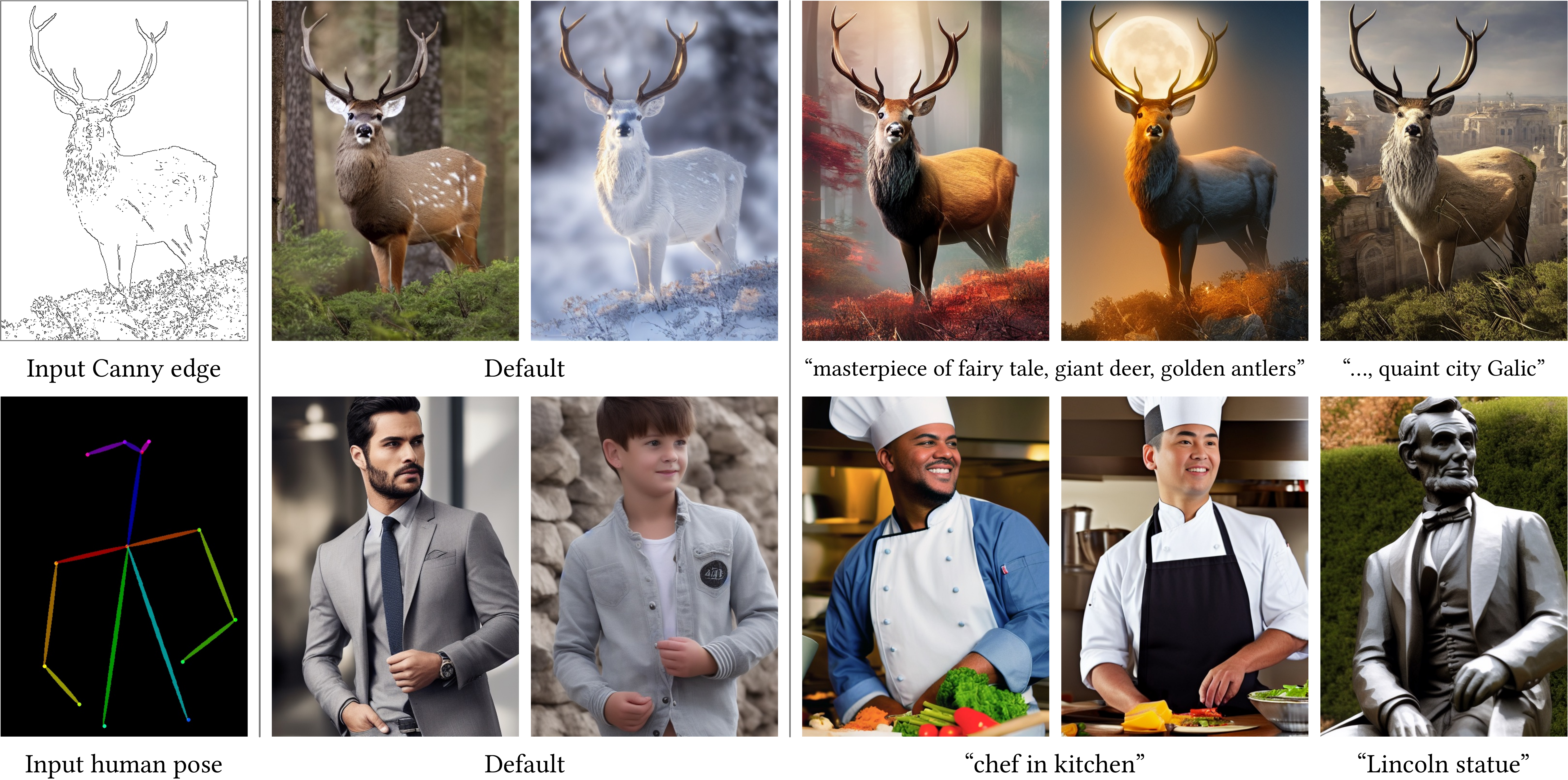} 
	\caption{Controlling Stable Diffusion\cite{rombach2022high} with Learned Conditions. ControlNet\cite{zhang2023adding} allows users to add conditions like Canny edges (top), human pose (bottom), etc., to control the image generation of large pretrained diffusion models. The default results use the prompt ``a high-quality, detailed, and professional image''. Users can optionally give prompts like ``chef in kitchen''.} 
	\label{controlnet}
        \vspace{-0.4cm}
\end{figure*}

\begin{figure}[t]  
	\centering
    \includegraphics[width=1\linewidth]{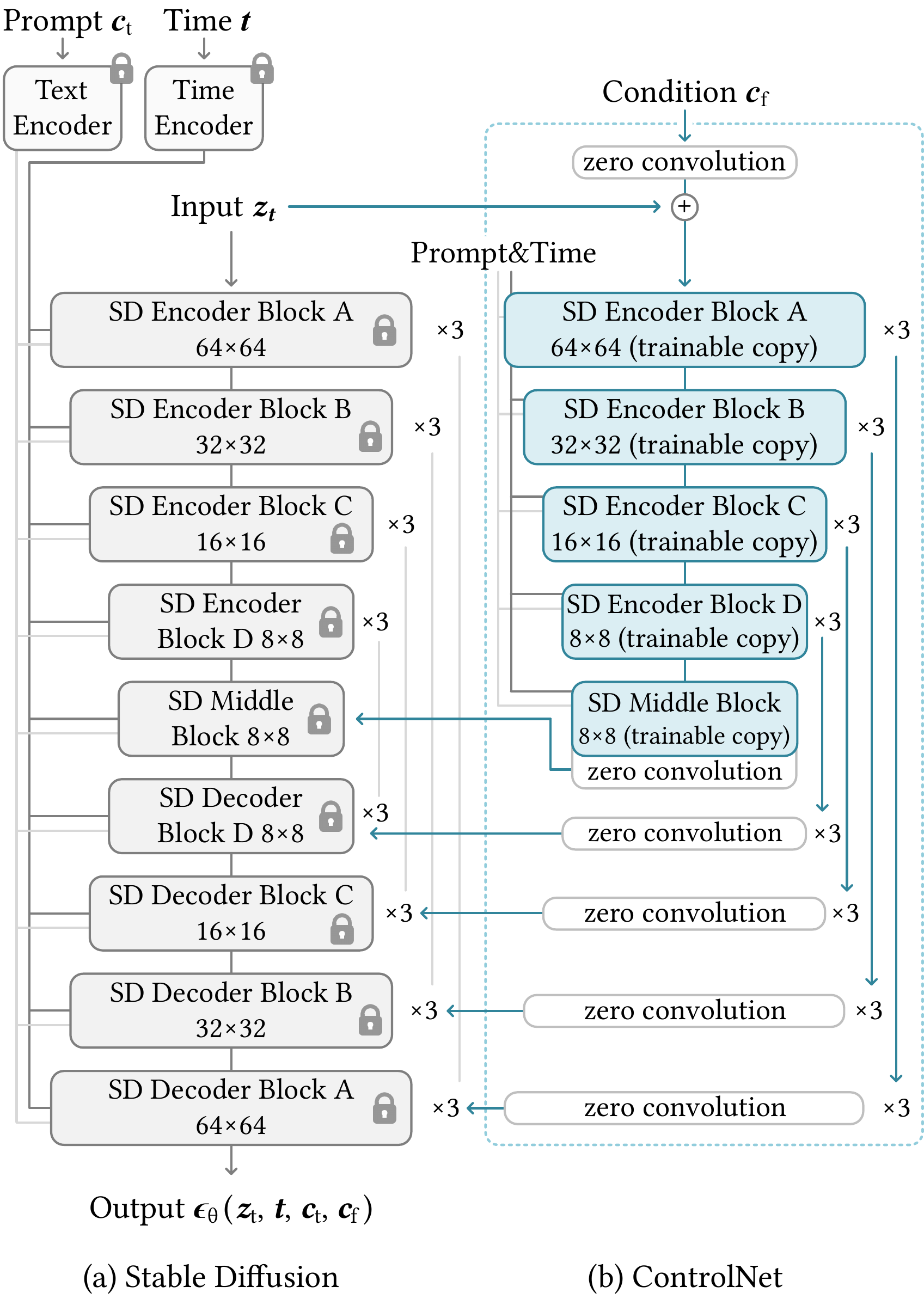} 
	\caption{Stable Diffusion\cite{rombach2022high}’s U-net architecture is connected with a ControlNet\cite{zhang2023adding} on the encoder blocks and middle block. The locked, gray blocks show the structure of Stable Diffusion V1.5 (or V2.1, as they use the same U-net architecture). The trainable blue blocks and the white zero convolution layers are added to build a ControlNet.} 
	\label{controlnetmethod}
        \vspace{-0.4cm}
\end{figure}

In terms of encoding, ELITE \cite{wei2023elite} introduces a combination of global and local mapping networks to enable fast and accurate customized generation, while Tailored Visions \cite{chen2024tailored} enhances personalization by rewriting user prompts based on historical interactions. LayoutDiffusion \cite{zheng2023layoutdiffusion} focuses on layout-controlled image generation, overcoming the challenges of multimodal fusion between images and layouts, thus offering improved generation quality and controllability.

HyperDreamBooth \cite{ruiz2024hyperdreambooth} uses a hypernetwork to generate personalized weights from a single image, allowing for efficient style and context switching. SAG \cite{chan2024improving} employs dual classifier-free guidance to ensure that generated outputs align with both themes and input text prompts, enhancing generation accuracy. Personalized Residuals \cite{ham2024personalized} introduces a low-rank personalization method that improves effectiveness while maintaining generation speed.

FastComposer \cite{xiao2024fastcomposer} is a multi-theme generation method that does not require fine-tuning, using an image encoder to extract theme embeddings for efficient personalized generation. InstantID \cite{wang2024instantid} retains identity in a zero-shot manner, guiding generation through facial images and text prompts to ensure high fidelity. Specialist Diffusion \cite{lu2023specialist} can be seamlessly integrated into existing models, learning complex styles and demonstrating efficient tuning capabilities with high-quality samples.

ControlStyle \cite{chen2023controlstyle} focuses on text-driven stylized generation, aiming to enhance the editability of content creation, while UniPortrait \cite{he2024uniportrait} offers a unified framework for portrait image personalization, supporting customization for both single and multiple identities. Face-Diffuser \cite{wang2024high} generates high-fidelity images by jointly learning scenes and characters, using a saliency-adaptive noise fusion mechanism to improve image quality.

TIGC \cite{li2025tuning} introduces a non-tuning image customization framework that modifies detailed attributes based on text descriptions while preserving key subject features. DETEX \cite{cai2024decoupled} enhances the flexibility of personalized generation using decoupled concept embeddings, and FlashFace \cite{zhang2024flashface} improves identity retention accuracy by encoding facial identity through feature maps. IDAdapter \cite{cui2024idadapter} and CoRe \cite{wu2024core} enhance generation diversity and text alignment accuracy using non-tuning methods and contextual regularization techniques, respectively.

\begin{figure*}[h]   
	\centering
    \includegraphics[width=0.9\linewidth]{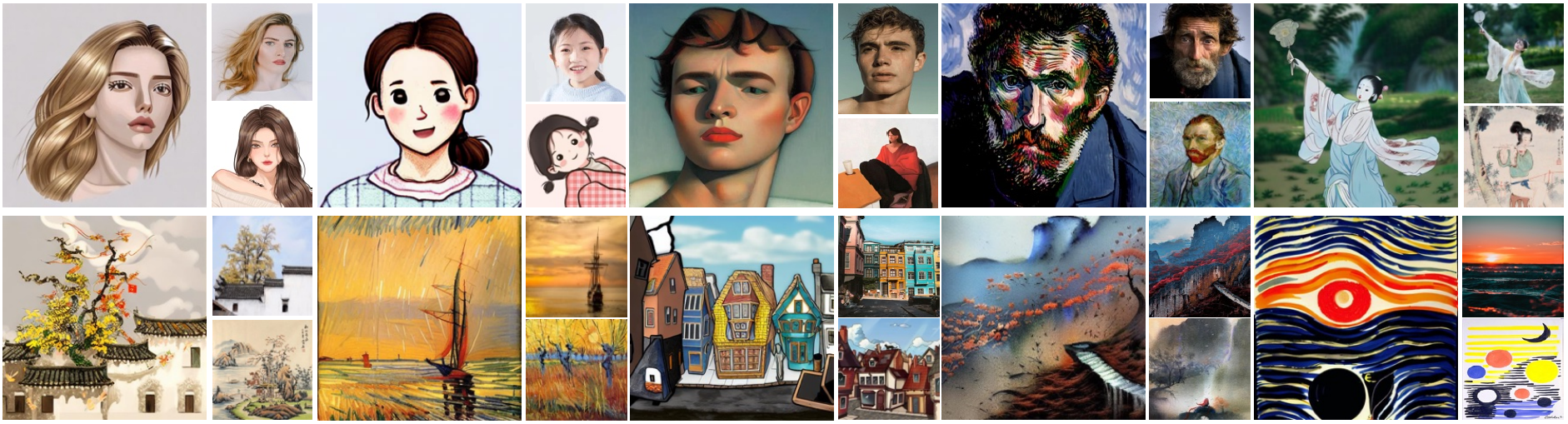} 
	\caption{Style transfer results using the proposed method. Given only a single input painting, InST\cite{zhang2023inversion} can accurately transfer the style
attributes such as semantics, material, object shape, brushstrokes and colors of the references to a natural image with a very simple learned textual description ``[C]''.} 
	\label{inst}
        \vspace{-0.4cm}
\end{figure*}

Finally, Imagine Yourself \cite{he2024imagine} represents an advanced model that requires no fine-tuning, pushing the boundaries of personalized image generation through a novel synthesis pairing mechanism and parallel attention architecture. Collectively, these research advances drive the development of personalized T2I generation technology, providing users with diverse and high-quality generation experiences.

More recent work broadens personalization from a single reference subject to multi-subject and narrative settings. AnyStory \cite{he2025anystory} targets unified single- and multiple-subject personalization, and XVerse \cite{chen2025xverse} controls the identity and semantic attributes of multiple subjects via DiT modulation. These settings sharpen the central tension of personalization---preserving subject identity while retaining editing freedom---and add the further challenge of keeping several distinct identities consistent within a single image.

\subsection{Controllable T2I Generation}
\label{Controllable}

Advancements in controllable T2I generation technology allow users to exercise more precise control over the content and style of generated images.

ControlNet \cite{zhang2023adding} is a key technology in this field, as illustrated in Figure \ref{controlnet}. It works by introducing conditioning images, e.g., sketches, edge maps, and pose keypoints, to guide the generation process of pretrained image diffusion models, thereby enhancing the flexibility and customization of the generated content.

Specifically, ControlNet maps additional conditional information into the parameters of a fixed pretrained large model, such as Stable Diffusion \cite{rombach2022high}. As illustrated in Figure \ref{controlnetmethod}, the gray module on the left represents the pretrained Stable Diffusion model, whose parameters remain unchanged. ControlNet includes encoder blocks and middle blocks consistent with the Stable Diffusion model, with initialized parameters copied from the pretrained model.

Furthermore, Uni-ControlNet \cite{zhao2024uni} offers a unified framework that simultaneously utilizes local controls (such as edge maps, depth maps, and segmentation masks) and global controls (such as CLIP image embeddings), significantly reducing the cost of training from scratch. The T2I-Adapter \cite{mou2024t2i} acts as a pluggable module, enhancing the control capabilities of T2I models by injecting additional conditional information, including text descriptions, image templates, and keypoints, into the generation process.

Additionally, Custom Diffusion \cite{kumari2023multi} achieves rapid model adjustments to represent new concepts by optimizing a minimal number of parameters, with the ability to train multiple concepts concurrently. Jedi \cite{zeng2024jedi} generates personalized images based on an arbitrary number of reference images, constructing a synthetic dataset of related images to learn the shared distribution of multiple text-image pairs. ViCo \cite{hao2023vico} provides a fast and lightweight solution for personalized generation, supporting plug-and-play features that avoid fine-tuning the original diffusion model while preserving the details of newly generated concepts.

DreamMatcher \cite{nam2024dreammatcher} introduces a semantically consistent masking strategy to more accurately replicate target attributes such as color and texture. RealCustom \cite{huang2024realcustom} improves alignment between text descriptions and generated content by gradually reducing the influence of real words using a cross-attention mechanism. Selectively Informative Description \cite{kim2024selectively} serves as an innovative text description strategy that utilizes descriptions generated by multimodal GPT-4, integrating seamlessly into the optimized model to enhance flexibility and accuracy in generation.

Additionally, StyleGAN Meets Stable Diffusion \cite{li2024stylegan} combines the identity-preserving capabilities of StyleGAN with diffusion models, enhancing semantic editing capabilities. LayerDiffusion \cite{li2023layerdiffusion} enables non-rigid editing and attribute modification of specific subjects through hierarchical control. ReCo \cite{yang2023reco} allows users to control arbitrary objects using open text by adding location markers and natural language region descriptions. Meanwhile, SpaText \cite{avrahami2023spatext} employs open-vocabulary scene control, combining global text prompts and segmentation maps to generate high-fidelity images. BLIP-Diffusion \cite{li2024blip} introduces a new multimodal encoder pretrained in BLIP-2 \cite{li2023blip} to achieve improved controllable generation.

Regarding generation efficiency, Codi \cite{mei2024codi} presents a conditional diffusion model capable of producing high-quality results quickly. StyleTokenizer \cite{li2024styletokenizer} enhances alignment between style and text representations through zero-shot style control \cite{romera2015embarrassingly} for image generation. P+ \cite{voynov2023p+} improves reconstruction capacity and expressiveness by introducing extended text inversion, while IP-Adapter \cite{ye2023ip}, as a lightweight adapter, facilitates efficient image prompting capabilities.

ResAdapter \cite{cheng2024resadapter} enables the generation of images with unrestricted resolutions and is compatible with various diffusion models. DetDiffusion \cite{wang2024detdiffusion} improves the effectiveness of the generated data by leveraging the synergy between perceptual models and generative models. CAN \cite{cai2024condition} adds control over image generation by dynamically adjusting neural network weights, making the generation process more flexible. SceneDiffusion \cite{ren2024move} optimizes hierarchical scene representation during the diffusion sampling process, supporting a range of spatial editing operations.

Zero-Painter \cite{ohanyan2024zero} is an innovative training-free framework that generates images based on text layout conditions, while FreeControl \cite{mo2024freecontrol} serves as a general-purpose zero-training solution supporting controllable generation across different conditions and architectures. PCDMs \cite{shen2023advancing} improve generation accuracy by gradually reducing discrepancies in pose-guided image generation.

Lastly, ControlNet++ \cite{li2025controlnet} and ControlNeXt \cite{peng2024controlnext} further enhance controllable generation, improving consistency between conditional controls and the resulting images. MultiDiffusion \cite{bar2023multidiffusion} provides a unified framework that leverages pretrained T2I diffusion models to achieve diverse and controllable image generation without requiring further training or fine-tuning. Composer \cite{xiao2024fastcomposer} introduces a new approach for flexibly controlling spatial layouts and color palettes in generated images while maintaining synthesis quality and model creativity.

\begin{figure*}[h]   
	\centering
    \includegraphics[width=0.9\linewidth]{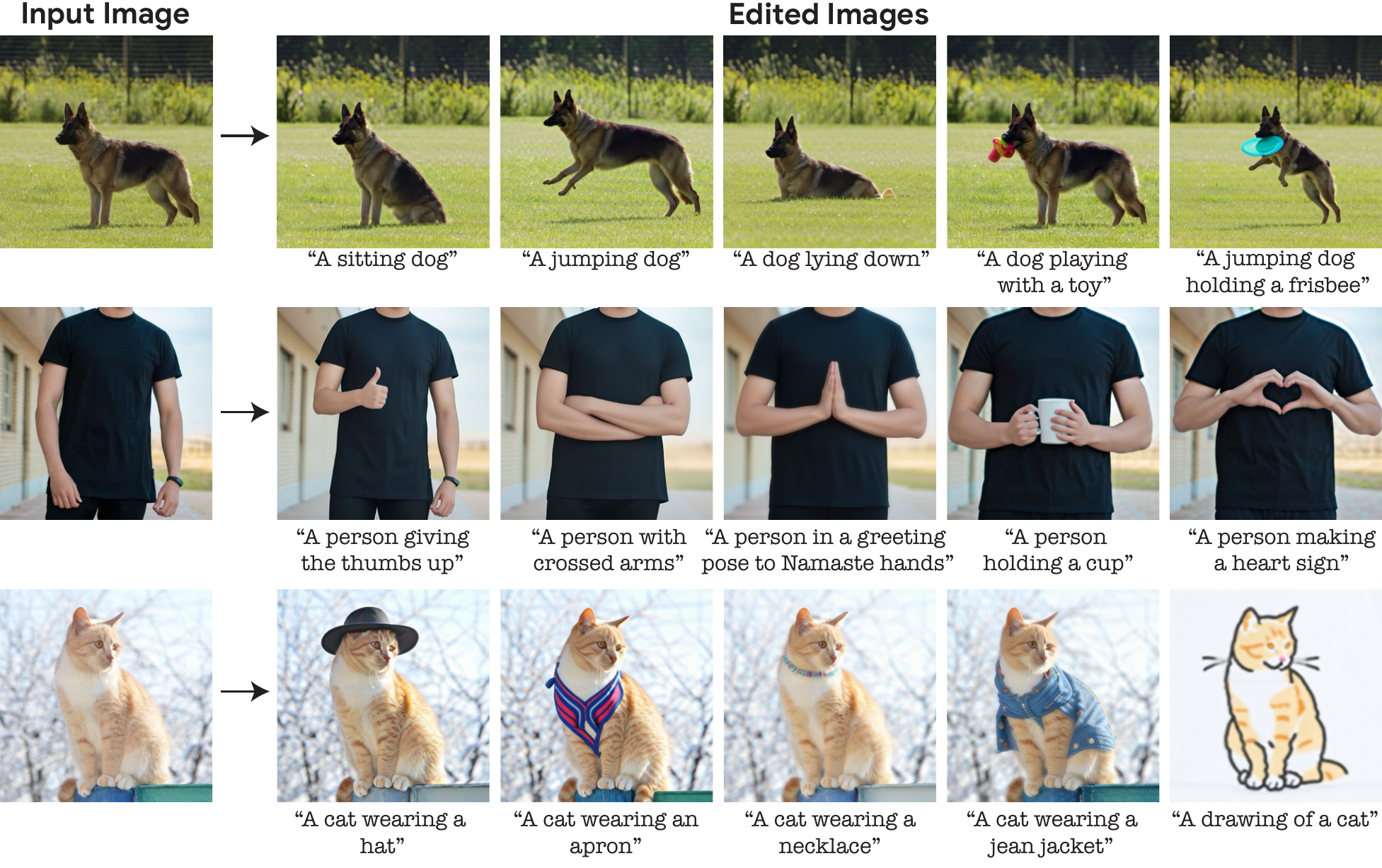} 
	\caption{Different target texts applied to the same image. Imagic\cite{kawar2023imagic} edits the same image differently depending on the input text.}
	\label{imagic}
        \vspace{-0.4cm}
\end{figure*}

\subsection{T2I Style Transfer}
\label{Transfer}

We need artistic images. T2I Style Transfer combines the benefits of T2I generation and style transfer, enabling users to create images in specific artistic styles by inputting text descriptions and style images. This area has a rich body of research, with various innovative methods developed to achieve effective style transfer.

First, InST \cite{zhang2023inversion} is a deep learning-based artistic style transfer framework capable of generating high-fidelity new artistic images while preserving the creative attributes of the original work. It does so by providing a single painting image and controlling the content using natural images or text descriptions, as illustrated in Figure \ref{inst}.

Styleformer \cite{wu2021styleformer} is a deep learning project based on the Transformer architecture that can alter the writing style while preserving the original text content, allowing the style transfer of textual information in the T2I domain. Furthermore, research has shown that by fixing the Gaussian noise introduced during the denoising process and adjusting the text embeddings \cite{wu2023uncovering}, it is possible to generate the target style without changing the semantic content, laying the foundation for future lightweight algorithms.

Similarly, ArtAdapter \cite{chen2024artadapter} uses a multilevel style encoder and an explicit adaptation mechanism to ensure a balance between style and text semantics, enabling rapid style adaptation.

Additionally, InstantBooth \cite{shi2024instantbooth} achieves personalized image generation without requiring test-time fine-tuning, effectively capturing both global and local features of the input image. OSASIS \cite{cho2024one} demonstrates a one-shot stylization method that decouples the semantics and structure of images, allowing for flexible style application.

DEADiff \cite{qi2024deadiff} uses a mechanism to decouple style and semantics, employing Q-Formers to extract feature representations for optimal results in visual stylization. InstantStyle \cite{wang2024instantstyle} introduces a simple mechanism to separate style and content, ensuring a balance between style control and visual quality.

StyleAligned \cite{hertz2024style} maintains style consistency among a series of generated images through a shared attention mechanism, allowing users to create stylistically consistent images with straightforward inversion operations. Finally, FreeCustom \cite{ding2024freecustom} offers a tuning-free approach that uses a multireference self-attention mechanism and weighted masking strategies to generate customized images based on multiconcept compositions from reference concepts.

\subsection{Text-guided Image Generation}
\label{Text-Guided}

Text-guided Image Generation encompasses not only the ability to create entirely new images but also the editing and modification of existing ones. This technology allows users to make detailed adjustments and reconfigurations of existing images based on textual prompts. Imagic \cite{kawar2023imagic}, as the first method aimed at semantic editing of a single image through text, significantly improves the precision of image editing by optimizing text embeddings and fine-tuning pretrained diffusion models, establishing itself as a benchmark in the field, as illustrated in Figure \ref{imagic}.

DisenDiff \cite{zhang2024attention}, on the contrary, uses a cross-attention mechanism to disentangle multiple concepts within a single image, allowing users to describe different elements independently, thus increasing the flexibility of image generation. Noise Map Guidance (NMG) \cite{cho2024noise} provides a context-rich inversion approach for editing real images, enhancing the naturalness of the editing results. SINE \cite{zhang2023sine} introduces a novel classifier-free guiding model that facilitates the extraction of knowledge from a single image for application to pretrained diffusion models, enabling effective content creation from a single image.

\begin{figure*}[h]    
	\centering
    \includegraphics[width=0.9\linewidth]{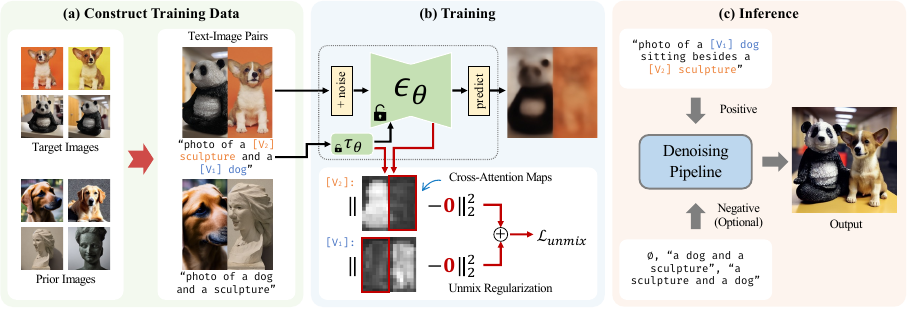} 
	\caption{SVDiff \cite{han2023svdiff} proposes the Cut-Mix-Unmix data augmentation technique to enhance the quality of multi-subject image generation, which is a simple text-based image editing framework.} 
	\label{svdiff}
        \vspace{-0.4cm}
\end{figure*}

AdapEdit \cite{ma2024adapedit} addresses complex, continuity-sensitive image editing tasks by employing variable spatiotemporal guiding scales, which enhance the naturalness and contextual consistency of edits. FISEdit \cite{yu2024accelerating} integrates cache-supported sparse diffusion model inference, allowing users to perform efficient image editing through minor text modifications, thus reducing inference time.

BARET \cite{qiao2024baret} enhances the flexibility of image editing by optimizing the image reconstruction process through target text inversion scheduling, progressive transition strategies, and balanced attention modules. D-Edit \cite{feng2024item} decouples the interaction between images and prompts, supporting various image editing by manipulating learned prompts.

Predicated Diffusion \cite{sueyoshi2024predicated} employs predicate logic to model the relationships between words in prompts, improving the fidelity of generated images. LEDITS++ \cite{brack2024ledits++} addresses the efficiency challenges associated with the processing of multiple text commands, allowing users to input several editing instructions simultaneously and generate realistic images, thus improving the user experience. PH2P \cite{mahajan2024prompting} examines the impact of prompt conditions during the diffusion process and introduces an immediate inversion framework that does not require auxiliary tasks.

PromptCharm \cite{wang2024promptcharm} advances T2I generation through multimodal prompt engineering and refinement, allowing users to explore different image styles within a large database. RAPHAEL \cite{xue2024raphael} uses a mixture of expert layers to create highly artistic images that accurately reflect complex textual prompts.

SUR-Adapter \cite{zhong2023adapter} introduces a semantic understanding and reasoning adapter, enhancing the ability of pretrained diffusion models to handle concise narratives. Prompt Diffusion \cite{wang2023context} establishes a contextual learning framework within diffusion generation models, enabling automatic comprehension of latent tasks and execution of corresponding operations.

SDG \cite{liu2023more} provides a unified framework for guiding image generation through reference images and language, allowing effective generation on datasets without textual annotations. InfEdit \cite{xu2023inversion} standardizes attention control mechanisms in text-guided editing, proposing an inversion-free method to accommodate complex modifications.

FPE \cite{liu2024towards} examines the influence of attention layers on image editing outcomes and introduces free prompt editing to simplify the editing process. The FoI \cite{guo2024focus} method ensures precise and coordinated editing of multiple instructions without requiring additional training. DiffEditor \cite{mou2024diffeditor} incorporates image prompts for fine-grained image editing, further improving the quality of edits.

Text-Driven Image Editing \cite{lin2024text} via Learnable Regions introduces a new region generation model, enabling mask-based T2I models to perform local image editing without the need for user-provided masks. Prompt Augmentation \cite{bodur2024prompt} expands a single input prompt to multiple target prompts, enriching the textual context and facilitating localized image editing.

\begin{figure*}[!t]    
	\centering
    \includegraphics[width=0.9\linewidth]{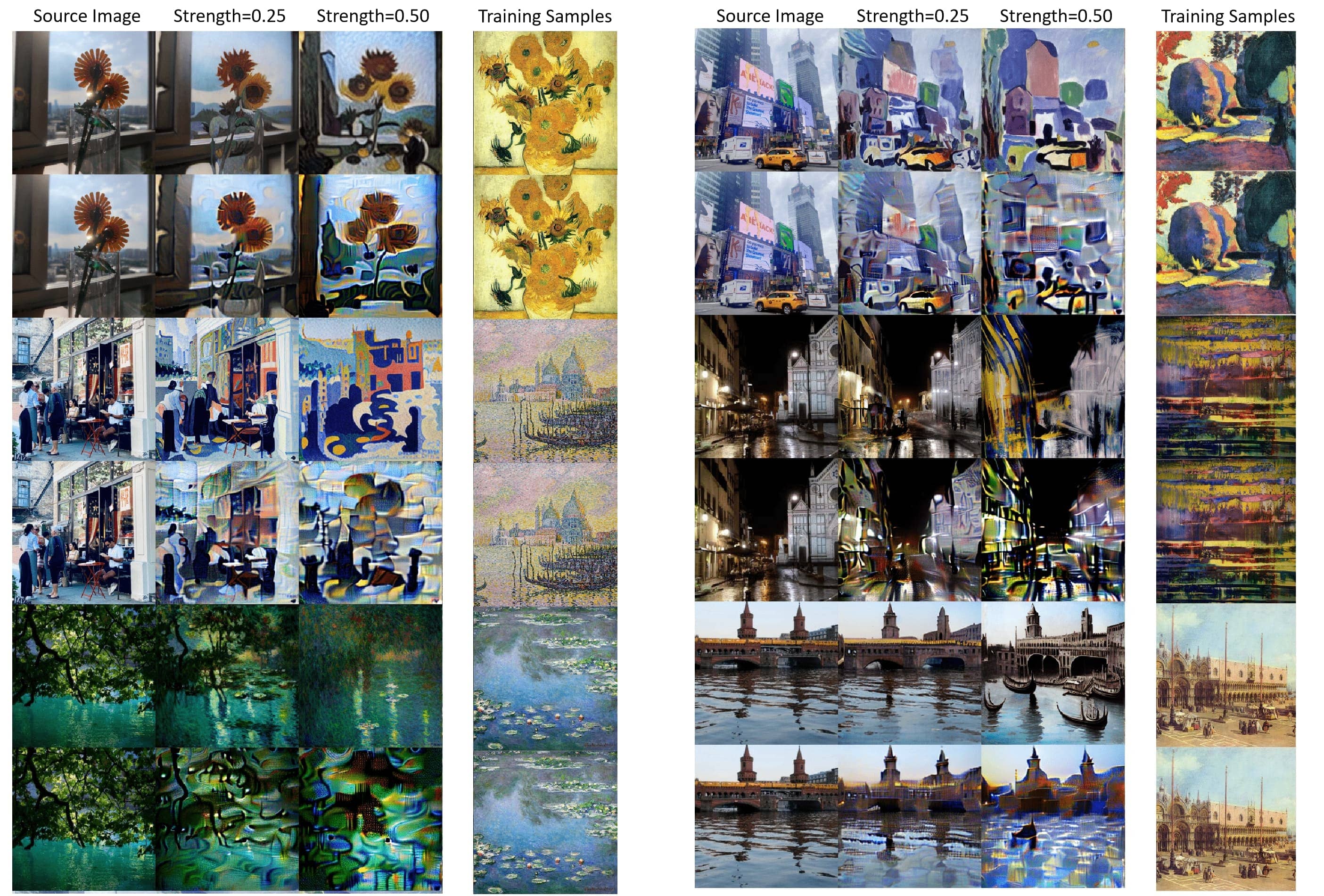}
	\caption{Images shown in each group
share the same source image. AdvDm\cite{liang2023adversarial} use textual inversion\cite{gal2022image} to extract the style of training samples from WikiArt, shown in a separate column. For each group, the top row shows the generated images based on the style extracted from the clean examples. The bottom row shows the generated images based on the style extracted from the adversarial examples. Strength is a hyper-parameter that indicates how much the style of the source image is covered by the target style. LDM fails to capture the style from adversarial examples, compared to clean images.}  
	\label{AdvDM}
        \vspace{-0.4cm}
\end{figure*}

TurboEdit \cite{wu2024turboedit} employs iterative inversion techniques using encoders for fine-grained editing of input images, supporting decoupled control in low-step diffusion models.

The comprehensive application of these technologies is driving rapid progress in the field of text-guided image generation. In particular, the breakthroughs in semantic editing by Imagic \cite{kawar2023imagic} have laid a solid foundation for future advances in this domain.

UFC-BERT \cite{zhang2021ufc} differs from existing two-stage autoregressive methods (such as DALL-E \cite{ramesh2021zero} and VQ-GAN \cite{esser2021taming}) by employing non-autoregressive generation (NAR) in the second stage. This design enhances the overall consistency of the synthesized images, supports the retention of specific image segments, and improves the synthesis speed.

Muse \cite{chang2023muse} employs a masked generative transformer architecture that randomly masks portions of the input image during training while designing pretraining text tasks for reconstruction learning. Compared to pixel-space diffusion models such as Imagen \cite{saharia2022photorealistic} and DALLE-2 \cite{ramesh2022hierarchical}, Muse is more efficient because it utilizes discrete tokens and requires fewer sampling iterations. Furthermore, in contrast to autoregressive models such as Parti \cite{yu2022scaling}, Muse achieves greater efficiency through parallel decoding. Using pretrained LLM, Muse improves fine-grained language understanding, resulting in high-fidelity image generation and a deeper comprehension of visual concepts.

MarkovGen \cite{jayasumana2024markovgen} is the first study to utilize MRF to enhance the efficiency and quality of text-to-image models. This model achieves a 1.5-fold acceleration and quality improvement by replacing the final steps of Muse with the learned MRF layers, as validated by human evaluation and FID metrics. The parameters of the MRF model can be trained in a few hours, facilitating the rapid integration of the MRF model with the pretrained Muse model, further enhancing both efficiency and quality.

The instruction-based editing paradigm has matured substantially since these early single-image methods. InstructPix2Pix \cite{brooks2022instructpixpix} established a representative formulation in which a model is trained to follow a textual edit instruction applied to a source image. Recent large-scale systems extend this interface: Step1X-Edit \cite{liu2025stepxedit} provides a general instruction-based editing framework, In-Context Edit \cite{zhang2025incontext} performs instructional editing through in-context generation within a large diffusion Transformer, and ReFlex \cite{kim2025reflex} edits real images directly in rectified-flow models via mid-step feature manipulation. FLUX.1 Kontext \cite{labs2025flux} further unifies in-context generation and editing and emphasizes multi-turn consistency. A recurring lesson is that editing must be assessed along separate axes---instruction satisfaction, preservation of unedited regions, and multi-turn stability---because an attractive re-rendering is not a successful edit if protected content is lost; the corresponding benchmarks are discussed in Section~\ref{section5}.

\subsection{Performance and Effectiveness}
\label{Performance}

In the field of T2I, balancing performance and efficiency is essential. Low-Rank Adaptation (LoRA) \cite{hu2021lora}, initially used to fine-tune large models in natural language processing, was later introduced to SD models. Specifically, as shown in Eq. \eqref{eq3}, consider a weight matrix $W$ that needs fine-tuning. LoRA represents the update to $W$ as the product of two low-rank matrices:
\begin{equation}
W' = W + BA.
\label{eq3}
\end{equation}

By decomposing the weight updates of a pretrained model into low-dimensional matrices, LoRA significantly enhances the efficiency of fine-tuning. Block-wise LoRA \cite{li2024block} further optimizes the fine-tuning strategy for U-Net, resulting in improved generation quality.

Latent variable models \cite{bartholomew2011latent,rombach2022high} have seen remarkable success in T2I, allowing the production of high-quality images from text prompts or other modalities. SVDiff \cite{han2023svdiff} introduces a compact parameter space approach that mitigates overfitting and language drift issues by fine-tuning the singular values of matrices. It also incorporates a cut-mix-decomposition data augmentation technique. The basic framework of SVDiff is illustrated in Figure \ref{svdiff}.

Token-Critic \cite{lezama2022improved}, an auxiliary model, enhances the sampling capabilities of non-autoregressive generative transformers, thereby improving the quality of generated content. Null-text Inversion \cite{mokady2023null} enhances image editing by employing diffusion key inversion and optimizing empty text prompts.

The study \cite{li2024scalability} conducted a large-scale comparative analysis to explore the scalability of diffusion-based T2I generation methods. ECLIPSE \cite{patel2024eclipse} leverages pretrained visual language models for contrastive learning, significantly reducing data requirements while maintaining generation quality. InstaFlow \cite{liu2023instaflow} is the first one-step T2I generator capable of producing high-quality images.

The Progressive Distillation technique \cite{salimans2022progressive} greatly reduces the sampling time by distilling the behavior of pretrained diffusion models into new models, with minimal impact on sample quality. YOSO \cite{luo2024you} achieves high-fidelity, one-step image synthesis with enhanced training stability. SwiftBrush \cite{nguyen2024swiftbrush} also focuses on one-step generation, producing images that are comparable in quality to those generated by stable diffusion models.

Speculative Jacobi Decoding (SJD) \cite{teng2024accelerating} accelerates the inference process in autoregressive T2I generation by allowing the model to predict multiple tokens simultaneously at each step. LinFusion \cite{liu2024linfusion} introduces a novel linear attention mechanism that significantly reduces both time and memory complexity while maintaining performance. DiffFit \cite{xie2023difffit} adapts quickly to new domains by fine-tuning only specific layer parameters, thus reducing training time.

StreamDiffusion \cite{kodaira2023streamdiffusion} is a real-time interactive generation pipeline that enhances speed through batch denoising. TiNO-Edit \cite{chen2024tino} focuses on optimizing the noise patterns and diffusion time steps during the editing process, proposing new loss functions to accelerate optimization.

The combined application of these technologies effectively balances performance and efficiency in T2I, driving the continued advancement of T2I technology.

\subsection{Reward Mechanism}
\label{Reward}

In T2I generation, designing an effective reward mechanism is crucial to optimize the generation process. By introducing effective reward functions \cite{watkins1989learning}, generative models can better align with human aesthetics and preferences.

ImageReward \cite{xu2024imagereward} is the first general-purpose reward model for T2I generation that effectively encodes human preferences. It can be used as a reward function in RLHF (Reinforcement Learning from Human Feedback), thus further optimizing generative models' performance. ImageReward provides a clear scoring mechanism for the generation process and has had a significant impact in T2I.

\cite{hao2024optimizing} uses reinforcement learning to explore better prompting strategies that generate images more aligned with user expectations.

Prompt AutoEditing (PAE) \cite{mo2024dynamic} aims to transform user-provided raw prompts into refined, dynamic prompts for generating high-quality images. PAE adjusts the weights of each word and modifies the injection time steps through an online reinforcement learning strategy, establishing a multidimensional reward system that considers aesthetic scores, semantic consistency, and user preferences.

Annealed Importance Guidance (AIG) \cite{jena2024elucidating} is an inference-time regularization method that achieves a Pareto optimal trade-off between rewards and image diversity, allowing the model to optimize rewards while preserving diversity. This approach effectively mitigates issues related to indiscriminate regularization and reference mismatches.

IterComp \cite{zhang2024itercomp} improves the generative capabilities of the base diffusion model by collecting composition-aware preferences from a model library and utilizing iterative feedback learning. The associated RL fine-tuning framework \cite{miao2024training} further improves generation diversity by incorporating theoretically supported diversity rewards.

SPIN-Diffusion \cite{yuan2024self} introduces a self-play fine-tuning algorithm that fosters competition between the diffusion model and its earlier versions, encouraging an iterative self-improvement process. This approach serves as an alternative to traditional supervised fine-tuning and reinforcement learning strategies.

Lastly, Powerful and Flexible \cite{wei2024powerful} presents a novel reinforcement learning framework that employs deterministic policy gradient methods for personalized T2I generation. This framework enables the integration of various objectives to enhance the quality of the generated images. Collectively, these works advance the development of reward mechanisms in the T2I field, providing effective strategies for achieving high-quality image generation.

\subsection{Safety Issues}
\label{Safety}

In the field of T2I generation, safety concerns are gaining increasing attention, especially since generated images can contain biases and unsafe elements, such as gender discrimination or depictions of violence that may be harmful to children. To address these issues, researchers have proposed various methods to ensure ethical and safe generation processes.

Recent studies have employed detection-based methods \cite{rando2022red}, guiding techniques \cite{wu2024universal,rombach2022high,schramowski2023safe}, and fine-tuning approaches \cite{gandikota2023erasing} to intervene in both text and image content, aiming to make T2I generation safer.

Structured Permeable Membrane (SPM) \cite{lyu2024one} introduces a conceptual structure into diffusion models to perform targeted erasure, effectively mitigating unwanted variations and preventing harmful content from appearing in generated images.

The Prompt Optimizer (POSI) \cite{wu2024universal} is an automated system that optimizes text prompts, guiding T2I models to generate images that are both safe and semantically aligned with the original prompts. This system preprocesses potentially toxic prompts using GPT \cite{radford2018improving} and applies fine-tuning based on datasets, thus enhancing the safety and reliability of T2I models.

SteerDiff \cite{zhang2024steerdiff} is a lightweight adapter module designed to serve as an intermediary between user input and the diffusion model, ensuring that generated images meet ethical and safety standards. This method identifies and adjusts inappropriate concepts within the text embedding space to guide the model away from harmful outputs.

Recently, the Self-Discovering method proposed by Li et al. \cite{li2024self} identifies ethically-related latent vectors by finding corresponding latent directions in the U-Net bottleneck layer and suppressing their activation, thereby reducing the likelihood of generating inappropriate content. This approach offers a new perspective by discovering and controlling ethical concepts in an interpretable latent space to achieve responsible generation.

Moreover, MetaCloak \cite{liu2024metacloak} prevents unauthorized subject-driven synthesis by creating perturbations using a pool of proxy diffusion models to generate transferable, model-agnostic perturbations that mitigate the harmfulness of personalized generation. OpenBias \cite{d2024openbias} tackles open-set bias detection in T2I generation models by using large language models to extract potential biases and quantify them through visual question-answering models.

HFI (Human Feedback Inversion) \cite{kim2024safeguard} proposes a framework that condenses guidance for generating images based on human feedback, enhancing the model's alignment with human judgment, and significantly reducing the generation of inappropriate content. RECE (Reliable and Efficient Concept Erasure) \cite{gong2024reliable} provides an innovative approach to rapidly modify models without additional fine-tuning, employing closed-form solutions to derive new target embeddings that mitigate inappropriate content generation.

Finally, RIATIG \cite{liu2023riatig} introduces a reliable and stealth adversarial attack method for T2I models using subtle examples. This approach formulates the example creation process as an optimization problem to generate prompts that remain undetectable to the model.

Together, these studies and methods contribute to advancements in addressing safety issues within the T2I field, offering effective solutions to generate ethically sound and high-quality images.

Recent work also refines how safety is measured and enforced. Fine-Grained Erasure \cite{thakral2025finegrained} removes specific concepts from diffusion foundation models at finer granularity while limiting collateral damage to unrelated concepts. Conversely, OVERT \cite{cheng2025overt} benchmarks \emph{over-refusal}---the tendency of safety-tuned models to reject benign prompts---underscoring that safety interventions must be evaluated jointly with usability, not by harmful-content suppression alone.

\subsection{Copyright Protection Measures}
\label{Copyright}

As T2I models continue to advance, concerns about copyright infringement are becoming more prominent. To address this challenge, researchers have proposed various technologies and methods to protect artists' rights.

AdvDM \cite{liang2023adversarial} effectively disrupts the fine-tuning process of models by adding pixel-level micro watermarks to artworks, preventing them from replicating artistic styles. After fine-tuning, the model generates new artworks based on learned style concepts. As illustrated in Figure \ref{AdvDM}, images generated after AdvDM processing generally display lower quality, chaotic textures, and a loss of artistic usability. This shows that AdvDM can effectively prevent artistic style transfer in diffusion models.

Mist \cite{liang2023mist} employs adversarial attacks to inject imperceptible noise into generative models, preserving the uniqueness of the original artistic style. Anti-DreamBooth \cite{van2023anti} applies subtle perturbations to user images to actively protect users from unauthorized personalized generators, thereby safeguarding privacy.

Furthermore, InMark \cite{liu2024countering} introduces a robust watermarking technique to protect images from unauthorized use, ensuring that personal semantics remain intact even if the images are modified. VA3 \cite{li2024va3}, a new online attack framework, exposes vulnerabilities in probabilistic copyright protection mechanisms and proposes effective strategies to enhance attack stability. Lastly, SimAC \cite{wang2024simac} addresses the issue of generating fake news featuring celebrity photos by designing a feature-based optimization framework to protect user privacy.

Together, these studies and technologies provide effective solutions for copyright protection and promote responsible use of T2I technology.

\subsection{Consistency Between Text and Image Content}
\label{Consistency}

\begin{figure*}[h]     
	\centering
    \includegraphics[width=0.9\linewidth]{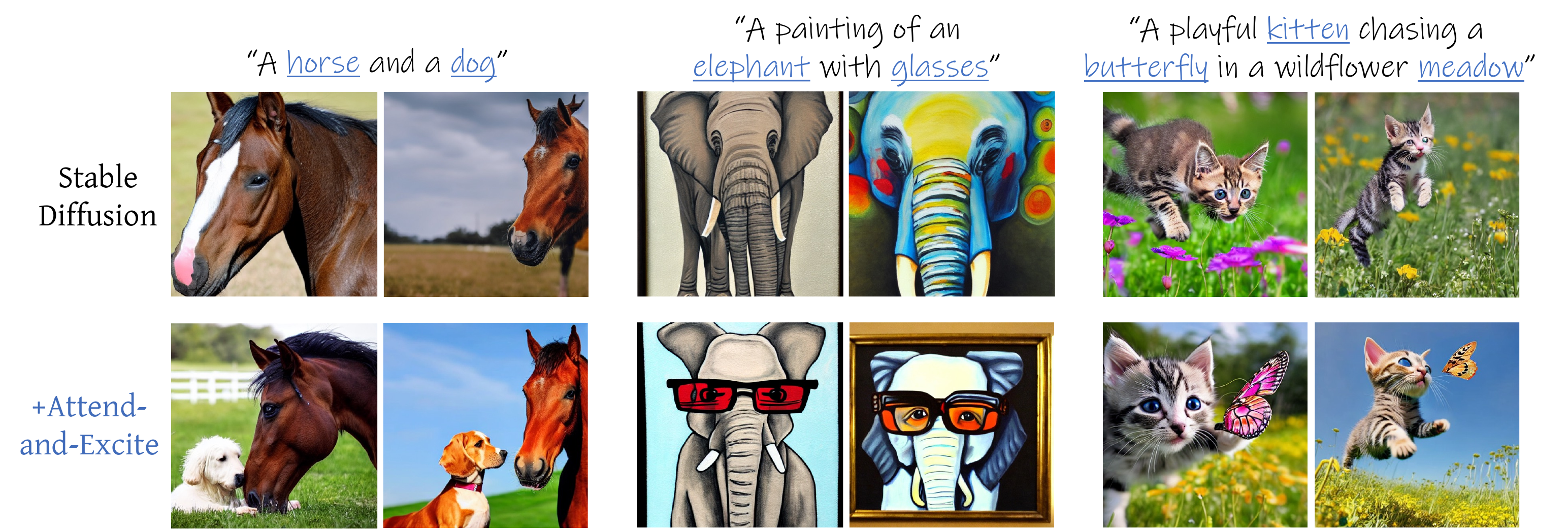} 
	\caption{Given a pretrained text-to-image diffusion model (e.g., Stable Diffusion\cite{rombach2022high} ) method in GSN\cite{chefer2023attend}, Attend-and-Excite, guides the generative model to modify the cross-attention values during the image synthesis process to generate images that more faithfully depict the input text prompt. Stable Diffusion alone (top row) struggles to generate multiple objects (e.g., a horse and a dog). However, by incorporating Attend-and-Excite (bottom row) to strengthen the subject tokens (marked in blue), we achieve images that are more semantically faithful with respect to the input text prompts.}  
	\label{gsn}
        \vspace{-0.4cm}
\end{figure*}

In the field of T2I, ensuring consistency between generated images and input text is a crucial research focus. To address this challenge, several studies have employed various methods and techniques. Recent research on GSN (Generative Semantic Nursing) \cite{chefer2023attend}, as shown in Figure \ref{gsn}, highlights that T2I models can sometimes overlook the main subject of the text during generation or incorrectly assign attributes to the wrong subject. GSN addresses these issues by adjusting latent codes during the denoising process, thereby improving the model's semantic understanding of the input text and reducing semantic errors in the generated output.

TokenCompose \cite{wang2023tokencompose} introduces a token consistency term that enhances the correlation between user-specified text prompts and generated images during fine-tuning, making it particularly effective for images containing multiple objects. Attention Refocusing \cite{phung2024grounded} utilizes large language models to generate layouts, thus improving the controllability of the generated images and reducing the likelihood of errors.

DiffusionCLIP \cite{kim2022diffusionclip} proposes a method for text-driven image manipulation using diffusion models, enabling zero-shot image operations in unseen domains and increasing text-image consistency. SDEdit \cite{meng2021sdedit} applies image denoising using stochastic differential equations, maintaining the realism and fidelity of the generated images while ensuring that the content aligns with the input text.

MasaCtrl \cite{cao2023masactrl} achieves consistent image generation and complex image editing through a mutual self-attention mechanism, ensuring coordination between the foreground and background elements. DPT \cite{qu2024discriminative} proposes a method to enhance T2I alignment by improving the alignment between text and images using discriminative detection and adjustments, effectively addressing misalignment issues.

In the domain of object recognition and editing, Referring Image Editing \cite{liu2024referring} provides a framework that allows users to identify and edit specific objects in images using text prompts, thereby enhancing text-image consistency. Scene TextGen \cite{zhangli2024layout} integrates character and word-level models to ensure accurate and coherent text rendering in images, improving overall consistency. AnyText \cite{tuo2023anytext} focuses on multilingual visual text generation, ensuring that the rendered text is accurate and coherent across different languages.

SceneGenie \cite{farshad2023scenegenie} utilizes bounding boxes and segmentation information to ensure accurate scene representation when generating high-resolution images. ZestGuide \cite{couairon2023zero} introduces a zero-shot segmentation guidance method that effectively aligns the generation process with input masks, further enhancing text-image consistency. DiffPNG \cite{yang2024exploring} employs panoramic narrative positioning to handle multiple noun phrases within text prompts, thereby boosting text-image consistency.

In terms of generating layouts and ensuring detailed fidelity, TL-Norm \cite{chen2022background} redefines the text-to-layout process by combining background and object layouts, which enhances the fidelity and consistency of generated images. INITNO \cite{guo2024initno} improves the consistency between prompts and generated images by employing initial noise optimization methods.

Compose and Conquer \cite{lee2024compose} incorporates three-dimensional object positioning control to ensure strict consistency between generated content and input prompts. StoryDiffusion \cite{zhou2024storydiffusion} introduces a consistency-driven self-attention mechanism, significantly enhancing the consistency among generated images.

\begin{figure*}[h]    
	\centering
    \includegraphics[width=0.9\linewidth]{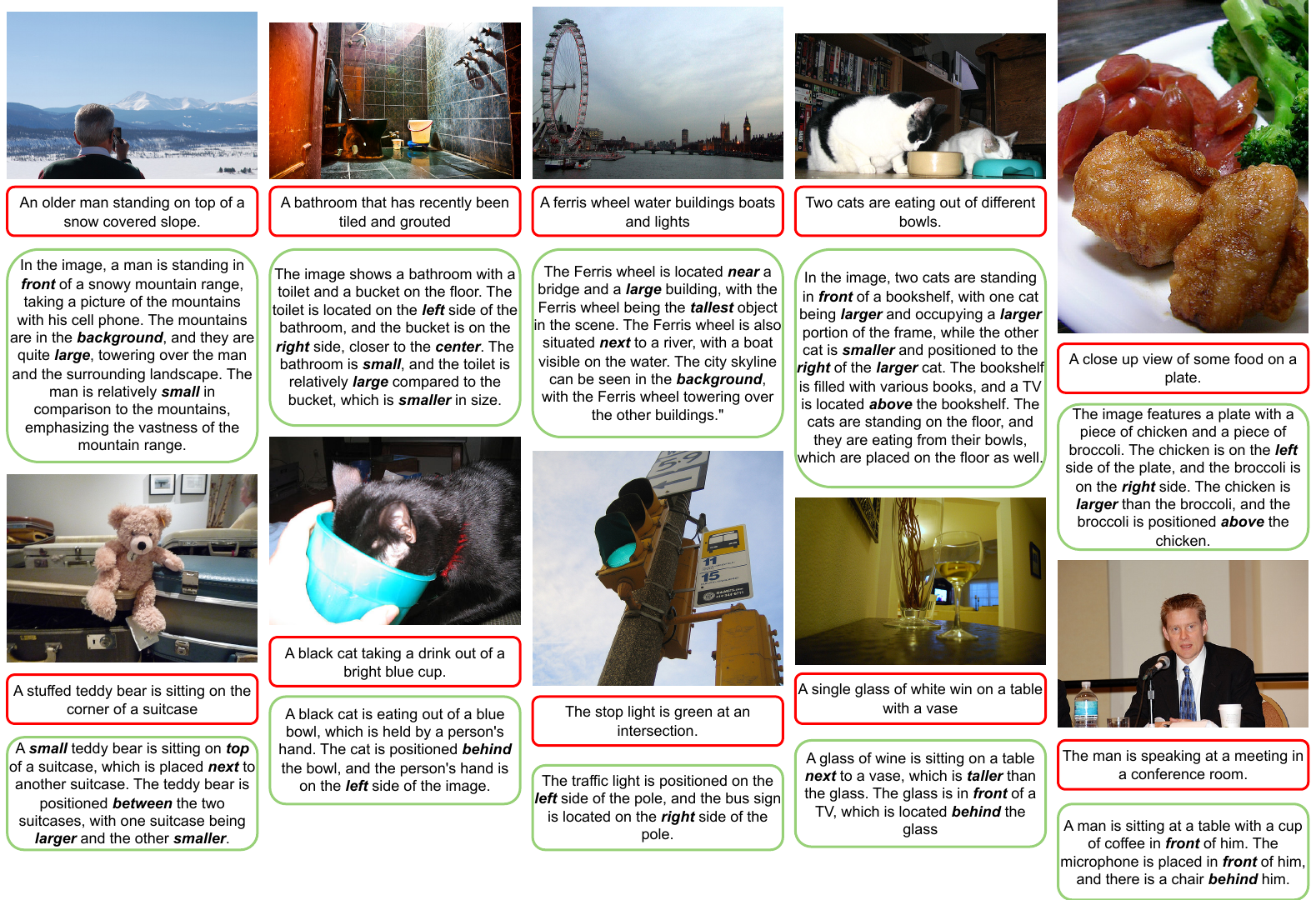} 
	\caption{The existing vision-language datasets do not capture spatial relationships well. To alleviate this shortcoming, we synthetically re-caption
6M images with a spatial focus, and create the SPRIGHT (SPatially RIGHT)\cite{chatterjee2025getting} dataset. Shown above are samples from the COCO Validation Set, where text in denotes ground-truth captions and text in are corresponding captions from SPRIGHT.}
	\label{spright}
        \vspace{-0.4cm}
\end{figure*}

SynGen \cite{rassin2024linguistic} improves attribute binding consistency by aligning attention maps, thereby reducing semantic errors. Training-Free Structured Diffusion Guidance \cite{feng2022training} enhances T2I synthesis with structured representations, improving the accuracy of attribute binding. ParaDiffusion \cite{wu2023paragraph} focuses on generating images from long paragraphs, using the semantic understanding capabilities of large language models to enhance alignment between text and image feature spaces.

StyleT2I \cite{li2022stylet2i} uses a contrastive loss guided by CLIP to enhance the compositionality of T2I synthesis. Bounded Attention \cite{dahary2024yourself} limits the flow of information to prevent theme leakage, ensuring consistency across multiple themes during generation.

DAC \cite{song2024doubly} introduces an editability recovery framework using a dual counterfactual inference approach, enhancing text-image consistency. DDS \cite{hertz2023delta} employs a text-based image editing scoring function to guide images towards the intended direction of text descriptions, ensuring alignment. CDS \cite{nam2024contrastive} uses a contrastive denoising scoring method to achieve structural correspondence during image translation and editing, ensuring content controllability.

MIGC (Multi-Instance Generation Controller) \cite{zhou2024migc} ensures global alignment among multiple instances by managing controls such as instance quantity, position, attributes, and interactions. Adma-GAN \cite{wu2022adma} introduces an attribute-driven memory-augmented GAN, which enhances generation quality by incorporating attribute information. AtHom \cite{shi2022athom} utilizes two attention modules to extract relationships between independent and unified modalities. DSE-GAN \cite{huang2022dse} reorganizes text features within an adversarial multistage architecture to improve the coherence of generated outputs. Phung et al. \cite{phung2024grounded} propose a method to generate more semantically accurate compositions by restricting the attention area of each token in the prompt to the corresponding image section. Token Merging (Tome) \cite{hu2024token} enhances semantic binding by merging related tokens into a single composite token, ensuring that the object, its attributes, and subobjects share the same cross-attention map.
Lastly, StableDrag \cite{cui2024stabledrag} offers a stable and precise point-dragging editing framework, enhancing stability for long-distance manipulations. FreeDrag \cite{ling2024freedrag} is a feature-dragging method that reduces the complexity of point tracking, while RegionDrag \cite{lu2024regiondrag} is a region-based editing technique that allows users to control the editing process with greater precision.

With these diverse technical approaches, researchers have made significant progress in improving text-image consistency, driving advances in T2I generation technology.

\begin{figure*}[h]  
	\centering
    \includegraphics[width=0.9\linewidth]{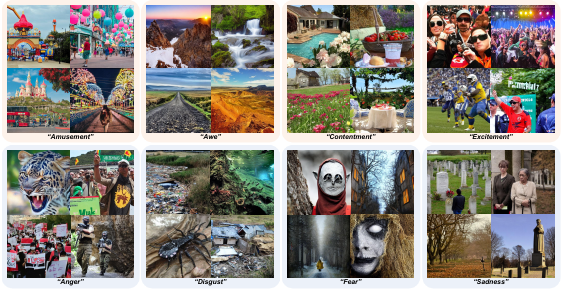}
	\caption{Emotional Image Content Generation (EICG). Given an emotion category, The network EmoGen\cite{yang2024emogen} produces images that exhibit unambiguous
meanings (semantic-clear), reflect the intended emotion (emotion-faithful) and incorporate varied semantics (semantic-diverse).} 
	\label{emogen}
        \vspace{-0.4cm}
\end{figure*}

\subsection{Spatial Consistency Between Text and Image}
\label{Spatial}
In the current field of T2I generation models, maintaining spatial consistency is a significant research focus. Although models like Stable Diffusion \cite{rombach2022high} and DALL-E \cite{ramesh2021zero} have made substantial progress in generating high-resolution, realistic images, they still struggle to accurately represent the spatial relationships described in text prompts. This limitation often results in generated images that do not reflect the correct positions and relative relationships of objects as described in the input text.

To address this issue, researchers have developed various methods aimed at enhancing spatial consistency through improvements in datasets, model architectures, and training strategies. For example, the SPRIGHT dataset \cite{chatterjee2025getting} provides detailed descriptions of spatial relationships, helping models learn how to transform text into images with accurate spatial layouts. As illustrated in Figure \ref{spright}, these descriptions not only include basic spatial vocabulary but also detail the relative sizes of objects and other relationships, significantly advancing the research on spatial consistency in models. CoMat \cite{jiang2024comat} improves the activation of text tokens by introducing the adversarial loss Fidelity Preservation, thereby encouraging the correct activation of image regions.

For user interaction, TCTIG \cite{yan2022trace} offers users a new way to control the generation process through mouse trajectory control, allowing them to draw paths while describing images, thus improving spatial control during generation. Meanwhile, the instance decomposition and occlusion information provided by the MULAN dataset \cite{tudosiu2024mulan} introduces new possibilities for image generation research.

PLACE \cite{lv2024place} introduces an adaptive layout-semantic fusion mechanism that ensures the effective integration of layout information and semantic features, enhancing both the quality and consistency of generated images. Using backward guidance \cite{chen2024training}, researchers can leverage user-specified inputs to better control layouts, uncovering various factors that influence layout during the image generation process.

Finally,  SSMG \cite{jia2024ssmg} improves generation quality and controllability through spatial-semantic map guidance, combined with relationship- and position-sensitive attention mechanisms. These studies and methods have made significant progress in improving T2I spatial consistency, advancing T2I model research, while providing crucial foundations and insights for future studies. By continuously improving models and datasets, researchers aim to achieve higher levels of text-image consistency to meet real-world application needs.

\begin{figure*}[h]    
	\centering
    \includegraphics[width=0.9\linewidth]{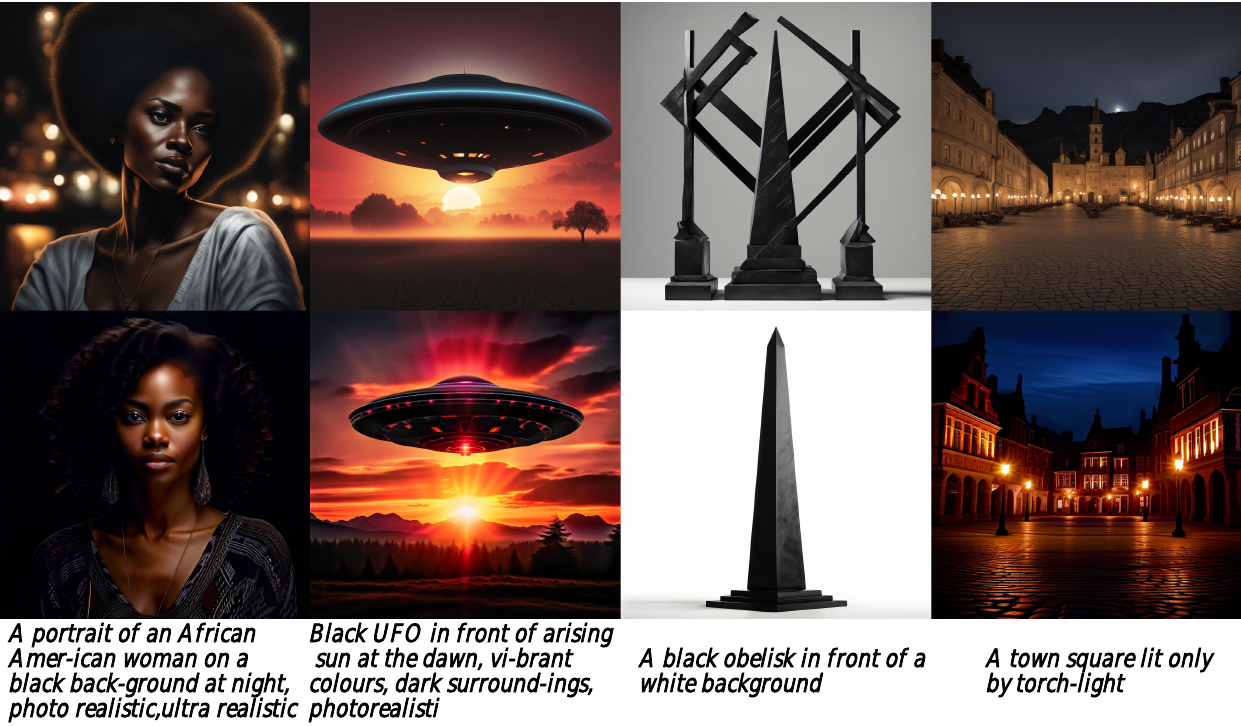}
	\caption{Colors and contrast. The top row is SDXL \cite{podell2023sdxl}, bottom row is Playground v2.5. The model Playground v2.5 \cite{li2024playground} can generate
samples with more vibrant colors and contrast.}
	\label{playgroundV2}
        \vspace{-0.4cm}
\end{figure*}

\subsection{Specific Content Generation}
\label{Specific}
In the field of T2I generation, specialized content generation is a key research direction that includes applications such as high-fidelity human imagery, emotional expression, and personalized image generation. To address these challenges, researchers have proposed various innovative methods. One of these is EmoGen \cite{yang2024emogen}, which introduces emotional image content generation (EICG), a novel task that focuses on creating images with clear semantic meaning and faithful emotional expression based on specified emotional categories. EmoGen ensures accurate emotional depiction by constructing an emotional space and aligning it with the CLIP space, as shown in Figure \ref{emogen}.

The CosmicMan model \cite{li2024cosmicman} targets high-fidelity human image generation, effectively capturing the requirements of text descriptions. Text2Human \cite{jiang2022text2human} uses a text-driven, controllable framework to synthesize complete human images by converting body poses into parsing maps. In contrast, Cross Initialization for Face Personalization \cite{pang2023cross} focuses on personalized face generation through cross-initialization, thereby improving visual fidelity. The HanDiffuser model \cite{narasimhaswamy2024handiffuser} excels in generating realistic human hand images by incorporating hand embeddings during the generation process. Furthermore, Wang et al. \cite{wang2024towards} proposed a human-centric priors (HcP) layer to improve the alignment between cross-attention maps and human-related textual information in prompts. This approach strengthens human structure information within cross-attention maps derived from text prompts, thereby improving the quality of generated images. The Face-Adapter \cite{han2024face} offers high-precision facial editing capabilities for pretrained diffusion models.

ZS-SBIR \cite{lin2023zero} uses diffusion models as feature extractors to tackle challenges in zero-shot sketch-based image retrieval. Text2Street \cite{su2024text2street} allows controllable street view generation by combining road topology and object layout generation strategies. PanFusion \cite{zhang2024taming} generates 360-degree images from text prompts by integrating stable diffusion models with panoramic branches to achieve cohesive image generation. StoryGen \cite{liu2024intelligent} creates current frames using text prompts and previous images, allowing the production of sequential images and story-like visuals. SVGDreamer \cite{xing2024svgdreamer} improves image editability by employing a semantic image vectorization process and introduces a scoring distillation method based on vector particles to enhance aesthetic appeal. StoryDALL-E \cite{maharana2022storydall} focuses on story continuation by generating visual stories from source images and enhancing pretrained models for sequential image generation.

Finally, TexFit \cite{wang2024texfit} and EditWorld \cite{yang2024editworld} have made notable progress in fashion image editing and world instruction-based image editing, respectively. These studies illustrate how T2I generation technology has made significant advances in specialized content generation, offering new opportunities for various future applications.

\subsection{Fine-grained Control in Generation}
\label{Fine-Grained}
In the field of fine-grained attention generation, several innovative methods have been proposed to improve the ability of T2I generation models to control details and allow personalization.

Playground v2.5 \cite{li2024playground} aims to improve the aesthetic quality of the generated images by addressing key challenges such as improving color and contrast, generating different aspect ratios, and refining human-centered details. Playground v2.5 highlights the importance of noise scheduling in achieving realism and visual fidelity in generated images. It also suggests that preparing a balanced bucketed dataset can help handle different aspect ratios during image generation and ensure that model outputs align with human aesthetic preferences, ultimately meeting perceptual expectations (as shown in Figure \ref{playgroundV2}).

Continuous 3D Words \cite{cheng2024learning} introduces a new three-dimensional perception method that learns continuous attribute control, allowing adaptation to various new conditions and improving object viewpoint changes. Localizing Object-level Shape Variations with T2I Diffusion Models \cite{patashnik2023localizing} achieves object-level shape exploration by generating variations in specific object shapes, using self-attention and cross-attention layers to precisely localize operations on these shapes. FiVA is the first fine-grained visual attribute dataset. Building on FiVA, FiVA-Adapter \cite{wufiva} decomposes image aesthetics into specific visual attributes, enabling users to apply features like lighting, texture, and dynamics from different images. This approach allows decoupling visual attributes from one or more source images and adapting them to the generated image, thereby controlling fine-grained visual attributes during generation.

\begin{figure*}[h] 
	\centering
    \includegraphics[width=0.9\linewidth]{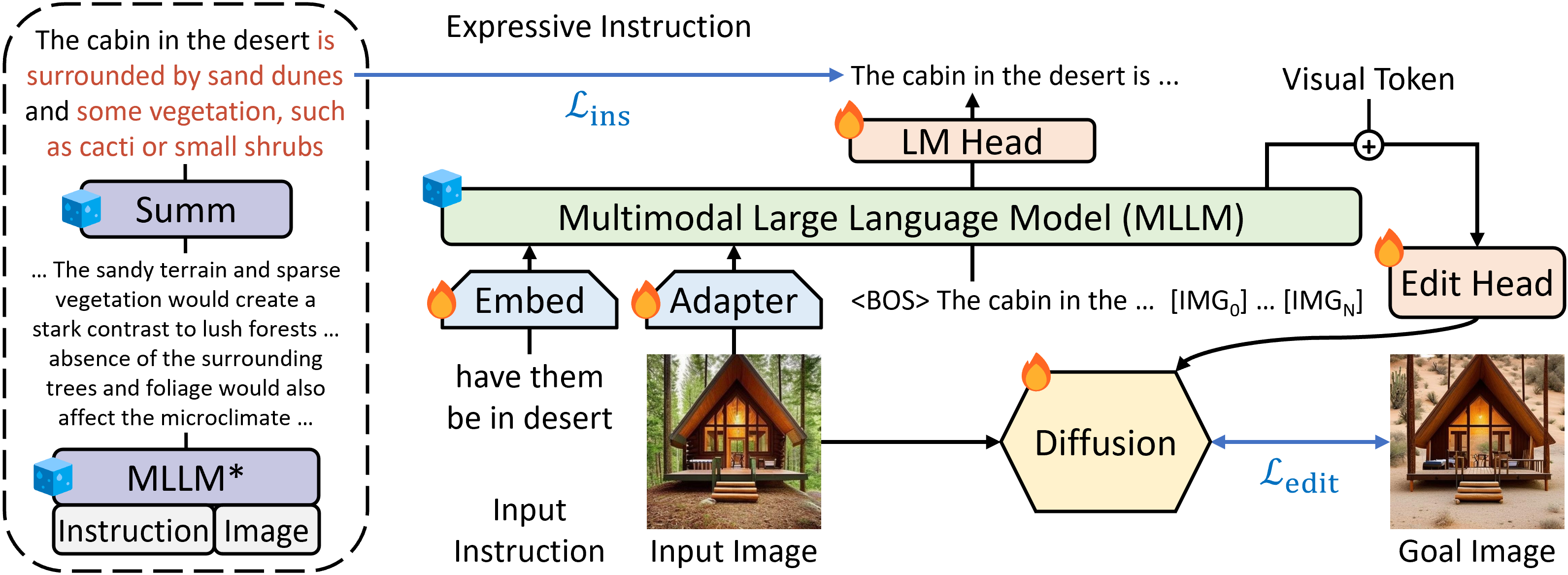}
	\caption{Overview of MLLM-Guided Image Editing (MGIE) \cite{fu2023guiding}, which leverages MLLMs to enhance instruction-based image editing. MGIE learns to derive concise and expressive instructions while providing explicit visual-related guidance for the intended goal. The diffusion model jointly trains and achieves image editing with latent imagination through the edit head in an end-to-end manner, showing that the module is trainable or frozen, respectively.}  
	\label{MGIE}
        \vspace{-0.4cm}
\end{figure*}

To address challenges in editing object layout, position, pose, and shape, Motion Guidance \cite{geng2024motion} employs a zero-shot technique that allows users to specify complex motion fields, guiding the diffusion model to achieve the desired effects. GLIGEN \cite{li2023gligen} introduces multimodal inputs, supporting not only textual prompts but also bounding boxes and reference images for more precise concept localization and generation control.

InteractDiffusion \cite{hoe2024interactdiffusion} incorporates interaction information into existing T2I models as an additional training condition, improving the accuracy of interactions in generated images, and improving the representation of complex relationships. NoiseCollage \cite{shirakawa2024noisecollage} independently estimates noise for objects using a noise collage technique, involving cropping and merging during the denoising process to optimize control over object layouts. Concept Weaver \cite{kwon2024concept} employs a two-step approach to customized T2I generation, enhancing identity fidelity by generating personalized targets.

ConForm \cite{meral2024conform} proposes a contrastive framework to address challenges related to high-fidelity generation, focusing on comparing object attributes to improve image quality. LayerDiffuse \cite{zhang2024transparent} allows pretrained latent diffusion models to generate transparent images by adjusting latent transparency while maintaining overall quality.

SingDiffusion \cite{zhang2024tackling} addresses the issue of initial singularities in time steps within diffusion models, improving performance during generation, and reducing FID scores. CFLD \cite{lu2024coarse} introduces a new training paradigm that controls the generation process using images, gradually refining learnable queries to separate fine-grained appearance and pose information. Finally, PreciseControl \cite{parihar2025precisecontrol} combines large T2I models with StyleGAN2 to achieve fine-grained attribute editing in the W+ latent space, allowing rough edits through text prompts.

\subsection{LLM-assisted T2I}
\label{LLM-Assisted}
In the field of LLM-assisted T2I, several significant research efforts have been made to improve the accuracy and flexibility of generative models. MGIE \cite{fu2023guiding} explores how multimodal large language models (MLLM) \cite{yin2023survey} can facilitate editing instructions, providing clear guidance to enhance image editing results, as shown in Figure \ref{MGIE}.

LLM-grounded Diffusion \cite{lian2023llm} and VPGen \cite{cho2024visual} leverage large language models (LLMs) to infer object positions using carefully crafted system prompts. LayoutGPT \cite{feng2024layoutgpt} integrates language understanding with graphic generation technologies, utilizing pretrained language models like GPT-4 \cite{achiam2023gpt} to interpret complex layout instructions and generate the required 2D and 3D layouts. It also incorporates GLIGEN \cite{li2023gligen} for image layout generation, GLIP \cite{li2022grounded} for image evaluation, and ATISS for 3D scene synthesis, forming a complete end-to-end process.

LayoutLLM-T2I \cite{qu2023layoutllm} is the first to explore layout planning in complex natural scenes using large language models and diffusion models. It proposes a feedback-based sampler learning paradigm and a layout-guided object interaction scheme. DiagrammerGPT \cite{zala2023diagrammergpt} uses GPT-4 as a ``planner'' to generate layout information for diagrams and as an ``auditor'' for detail optimization. SmartEdit \cite{huang2024smartedit} improves the understanding and reasoning abilities for complex instructions using MLLM \cite{yin2023survey}. It introduces a bidirectional interaction module (BIM) to facilitate comprehensive information exchange between input images and MLLM outputs.

AutoStory \cite{wang2023autostory} aims to generate story images that meet specific user needs with minimal human input, align text, and produce high-quality story visuals by combining LLM with diffusion models. CAFE \cite{zhou2024customization} offers rapid customization without additional tuning, handling diverse inputs and inferring user intentions. Ranni \cite{feng2024ranni} introduces a semantic panel as middleware to enhance the T2I generation process, improving the model's ability to follow instructions.

RFNet \cite{yao2025fabrication} explores generating images from complex prompts that require artistic creativity by presenting a Reality-Fantasy Network. This network combines diffusion models with large language models (LLMs) without requiring additional training. TIAC \cite{liao2024text} transforms abstract concepts into tangible physical objects by integrating information to generate prompts for T2I models. RPG \cite{yang2024mastering} uses multimodal LLMs to enhance the combinatorial capacity of T2I generation, breaking the generation process into multiple simpler tasks. ELLA \cite{hu2024ella} introduces an efficient LLM adapter to improve text alignment without the need to train additional models.

DialogGen \cite{huang2024dialoggen} establishes a multimodal interactive dialogue system for multi-turn T2I generation, ensuring effective prompt alignment and data organization. Lastly, MoMA \cite{song2024moma} utilizes multimodal LLMs as feature extractors and generators, enhancing the similarity of target objects in the generated images. These studies provide significant theoretical foundations and practical insights for LLM-assisted T2I generation, advancing the development of the field.
\section{ Datasets and Evaluation Criteria}
\label{section5}

\subsection{Datasets}
The COCO (Common Objects in Context) dataset \cite{lin2014microsoft} contains a diverse collection of more than 300,000 images, each paired with five human-generated descriptions that capture a wide variety of objects and scenes. This dataset is widely used in tasks related to image annotation, generation, and retrieval. Images are collected through crowdsourcing, and the descriptions are authored by professionals, ensuring both diversity and accuracy.

COCO Captions \cite{chen2015microsoft} is a dataset derived from the COCO dataset, which itself is a large-scale collection of images obtained from Flickr, showcasing multiple objects in natural environments. COCO Captions extends MS-COCO by providing 1,026,459 captions for 164,062 images. Each image in the MS-COCO dataset has five captions, and a subset of 5,000 images includes annotations with 40 reference sentences.

The Flickr30k dataset \cite{young2014image,plummer2015flickr30k} contains 30,000 images, each paired with five textual descriptions spanning various everyday scenarios. It is primarily used for image generation and image-to-text retrieval tasks, reflecting the diversity of real-world scenes. Images are sourced from the Flickr platform, with descriptions provided by users.

The Flickr8k dataset \cite{hodosh2013framing} includes 8,000 images, each accompanied by five distinct descriptions. It is commonly used for image annotation and generation tasks and is well-suited for training image-to-text generation models as well as assessing the descriptive capabilities of models.

The Oxford 102 Flower Dataset \cite{nilsback2008automated} includes images of 102 flower species, with 40 to 258 images per species, each accompanied by detailed descriptions. This dataset is commonly used for fine-grained image generation and classification tasks. The images are captured through field photography and manually annotated, ensuring accuracy and detail. It also serves as a benchmark for previous T2I studies.

The CelebA dataset \cite{liu2018large} consists of 200,000 images of celebrity faces, each tagged with 40 attribute labels. It is suitable for generating face images with specific features. The dataset is created by collecting and manually annotating celebrity images from the Internet, ensuring diversity and reliability of the attributes.

The Conceptual Captions dataset \cite{sharma2018conceptual} contains three million images, each paired with a natural language description. It focuses on conceptual understanding and aims to promote multimodal learning. The images are gathered from the web and the descriptions are generated by computational algorithms, reflecting the contextual information of the images. Additionally, the Conceptual 12M dataset \cite{changpinyo2021conceptual} includes 12 million images with concept-related descriptions.

LAION-400M \cite{schuhmann2021laion} is a large-scale multimodal dataset comprising approximately 400 million filtered image-text pairs collected from the web. It has been used as a resource for training and analyzing large-scale vision-language and generative models; it should not be described as a dataset for training CLIP alone. LAION-5B \cite{schuhmann2022laion} extends this web-scale collection to billions of image-text pairs. Occupational gender-bias subsets or analyses derived from such resources should be cited separately, rather than treating LAION-5B itself as a bias-specific dataset.

The Visual Genome dataset \cite{krishna2017visual} includes images along with detailed regional descriptions, supporting tasks such as object detection and image generation. The dataset provides detailed annotations generated through a combination of manual annotation and CV techniques, aiding in the understanding of complex image content.

The SBU Captions dataset \cite{ordonez2011im2text} contains over one million images paired with corresponding descriptions, primarily used to train and evaluate image generation models. The images in this dataset are collected from Flickr, and the descriptions are generated based on the context of the images, ensuring diversity and comprehensiveness.

DiffusionDB \cite{wang2022diffusiondb} is a large-scale prompt gallery dataset that provides extensive prompt data for T2I generation models, primarily gathered through automated collection and organization. PaintSkills \cite{cho2023dall} is a dataset focused on painting skills, emphasizing object recognition, quantity, and spatial relationships between objects in images and text. It is mainly used for researching models that generate artistic works and provides a wealth of painting samples to enhance the model's artistic capabilities.

The Pinterest40M dataset \cite{mao2016training} contains 40 million images along with associated texts, such as user comments and tags. It is intended for multimodal learning and recommendation systems, supporting tasks related to image content understanding and the training of recommendation algorithms. The dataset helps models learn associations between images and text, making it suitable for image retrieval, personalized recommendations, and advertising optimization tasks.

Nocaps \cite{agrawal2019nocaps} is a novel-object image-captioning benchmark. It evaluates caption generation for images containing object classes that are novel relative to the training captions, testing whether a model can describe objects it has not seen paired with text during training; it is not a general visual-question-answering dataset.

The UIUC Pascal Sentence Dataset \cite{rashtchian2010collecting} combines images with corresponding sentences and is primarily used for research in image understanding and description generation. Each image is accompanied by multiple sentence descriptions, supporting diverse generation tasks and NLP applications.

PASCAL-50S \cite{vedantam2015cider} provides images from the PASCAL VOC domain, each paired with 50 human-written sentences together with human relatedness judgments. The ``50S'' notation refers to the 50 descriptions per image, not to 50 object categories; it is mainly used to study caption-quality metrics and their agreement with human judgments.

ABSTRACT-50S \cite{vedantam2015cider} provides abstract (clip-art) scenes, each paired with multiple human descriptions and relatedness judgments. It is useful for studying caption quality and semantic agreement in a controlled visual domain, but it is not itself a general-purpose T2I training set.

Visual Question Answering (VQA) datasets \cite{antol2015vqa,li2023blip,achiam2023gpt} allow users to ask questions about images and receive corresponding responses. It supports the training and evaluation of models, helping to improve their visual understanding and reasoning capabilities. VQAv2.0 \cite{goyal2017making} is an updated version of the VQA dataset, featuring a wider range of question types and more diverse image data, increasing thus the complexity of the questions. This dataset is used to train and evaluate the performance of question-answering models.

The VCR (Visual Commonsense Reasoning) dataset \cite{zellers2019recognition} combines visual and textual information to support multiple-choice commonsense reasoning tasks, evaluating a model's ability to both answer a question and justify the answer. Winoground \cite{thrush2022winoground} probes compositional image-text understanding: it contains 400 examples, each with two images and two captions that use the same words in a different order, and a model must match them correctly. Its task design should not be conflated with VCR or with a T2I generation benchmark.

DrawBench \cite{saharia2022photorealistic}, developed alongside the Imagen model, includes 200 challenging prompts designed to assess the performance of various T2I models, covering 11 different categories. DrawBench serves as a valuable tool for T2I research, promoting the development of generative models in a variety of application scenarios.

ABC-6K \cite{feng2022training} contains 6,000 images with descriptions, suitable for image generation and understanding tasks, facilitating rapid experimentation. This dataset provides a straightforward environment for testing and validating image processing algorithms. Similarly, the CC-500 \cite{feng2022training} dataset includes 500 images with descriptions, mainly used for quick experimentation and validation, making it suitable for developing and testing preliminary models.

I2P (Inappropriate Image Prompts) \cite{schramowski2023safe} is a safety-evaluation set of real user-written prompts that tend to elicit inappropriate or unsafe content in text-to-image generation. It is used to measure and mitigate unsafe generation; it is not an image-to-image translation dataset.

RichHF-18K \cite{liang2024rich} contains 18,000 images with detailed descriptions and is primarily used for multimodal learning and evaluation. It offers complex annotations to support the fine-tuning of scoring models based on human feedback, providing valuable resources for evaluating generative models through detailed annotations and intricate feedback mechanisms.

T2I-CompBench \cite{huang2023t2i} is a dataset designed to investigate complex prompt combinations in image generation models. It is used to evaluate the performance of different models when handling complex textual prompts and provides various evaluation criteria to help researchers analyze the generative capabilities of the models.

These datasets enable researchers to comprehensively evaluate and optimize the performance of T2I generation models, advancing research in this field. Table~\ref{dataset} summarizes the representative datasets discussed above, together with their year, size, and source. We note that the ``Size'' column uses the unit reported by each resource (e.g., images, image-text pairs, prompts, or QA items), which are not directly comparable across rows.

\begin{table*}[ht]
    \centering
    \caption{Datasets in the T2I task. The ``Size'' column follows the unit reported by each resource (images, image-text pairs, prompts, or QA items) and is not directly comparable across rows.}
    \label{dataset}
    \begin{tabular}{@{}llll@{}}
        \toprule
        \textbf{Dataset} & \textbf{Year} & \textbf{Size} & \textbf{Source} \\ \midrule
        Oxford 102 Flower Dataset \cite{nilsback2008automated} & 2008 & 8,189 & University of Oxford \\
        UIUC Pascal Sentence \cite{rashtchian2010collecting} & 2010 & 1,000 & VOC2008 \\
        SBU Captions \cite{ordonez2011im2text} & 2011 & 1,000,000 & Flickr.com \\
        Flickr8K \cite{hodosh2013framing} & 2013 & 8,092 & Flickr.com \\        COCO \cite{lin2014microsoft} & 2014 & 330,000+ & Microsoft \\
        ABSTRACT-50S \cite{vedantam2015cider} & 2015 & 500 & ASD \\
        Flickr30k \cite{plummer2015flickr30k} & 2015 & 31,783 & Flickr.com \\
        COCO Captions \cite{chen2015microsoft} & 2015 & 204,721 & Microsoft \\
        PASCAL-50S \cite{vedantam2015cider} & 2015 & 1,000 & VOC2008 \\
        vQA \cite{antol2015vqa} & 2015 & 254,721 & MS COCO \& Abstract Images \\        Pinterest40M\cite{mao2016training} & 2016 & 40,000,000 & Pinterest.com \\
        Visual Genome \cite{krishna2017visual} & 2017 & 108,000 & Stanford University \\
        vQAv2.0 \cite{goyal2017making} & 2017 & 204,721 & MS COCO \\        CelebA \cite{liu2018large} & 2018 & 202,599 & Chinese University of Hong Kong \\
        Nocaps \cite{agrawal2019nocaps} & 2019 & 15,100 & Open Images V4 (Flickr.com) \\
        VCR \cite{zellers2019recognition} & 2019 & 110 & LSMDC \& YT \\
        Conceptual Captions \cite{sharma2018conceptual} & 2018 & 3,369,218 & Google Research \\
        LAION-400M \cite{schuhmann2021laion} & 2021 & 400,000,000 & LAION (Large-scale AI Open Network) \\
        CC-500 \cite{feng2022training} & 2022 & 500 & Synthetic prompts \\
        Conceptual 12M \cite{changpinyo2021conceptual} & 2021 & 12,423,374 & World Wide Web \\
        LAION-5B \cite{schuhmann2022laion} & 2022 & 5,000,000,000 & LAION \\
        DiffusionDB \cite{wang2022diffusiondb} & 2022 & 2,000,000 & Open-source contributions \\
        Winoground \cite{thrush2022winoground} & 2022 & 800 & Getty Images API \\
        DrawBench \cite{saharia2022photorealistic} & 2022 & 200 & DALL-E \& Reddit \\
        ABC-6K \cite{feng2022training} & 2022 & 6,400 & MS COCO \\
        I2P \cite{schramowski2023safe} & 2023 & 4,703 & User generated prompts \\
        T2I-CompBench \cite{huang2023t2i} & 2023 & 6,000 & Generated prompts by GPT \\
        PaintSkills \cite{cho2023dall} & 2023 & 65,535 & Synthetic prompts \\
        RichHF-18K \cite{liang2024rich} & 2024 & 18,000 & Pick-a-Pic \\
        \bottomrule
    \end{tabular}
        \vspace{-0.4cm}
\end{table*}

\subsection{Data Augmentation}
In T2I generation tasks, data augmentation is an important technique to improve model performance and generation quality. Common image enhancement techniques include rotation, translation, cropping, resizing, flipping, adding noise, and adjusting brightness, contrast, saturation, and random erasure. These techniques help maintain visual consistency and improve model robustness.
For text data augmentation, common methods include synonym replacement (replacing words in text descriptions to increase the diversity of training samples), text reconstruction (modifying sentence structures or using different descriptions while preserving the original meaning to enrich the dataset) and random insertion or deletion (randomly inserting or deleting words in the text to generate variations) \cite{frolov2021adversarial}.

In multimodal tasks, T2I data can be augmented by generating new images based on existing text descriptions and creating new text descriptions from existing images \cite{dong2017learning}. Additionally, combining these data augmentation methods can lead to even better results.

Label smoothing \cite{zhang2021delving} is also widely used to improve the robustness and accuracy of models during training. The Mixup technique \cite{zhang2017mixup} involves using convex combinations of samples and their labels to train neural networks. However, because interpolated samples do not introduce new information, relying solely on this method may not effectively address data scarcity. Therefore, retrieving relevant training data from external databases for re-imaging is crucial to further enhance model performance.

LAFITE \cite{zhou2022towards} is the first work to implement language-free training for T2I generation. This approach leverages the powerful pretrained CLIP model \cite{radford2021learning} to use its multimodal semantic space, thus reducing the reliance on explicit text conditions by generating text features from image features.

In recent T2I research, linear extrapolation has been used to enhance text features in small datasets. This means that new image data can be retrieved from search engines using text descriptions to expand the existing dataset \cite{ye2024data}. Furthermore, K-means clustering \cite{dhanachandra2015image} and fine-grained classifiers can be used to identify and remove irrelevant images, ensuring the quality of the new data. Furthermore, introducing a NULL condition guidance mechanism \cite{bodla2018semi,yan2016attribute2image,van2016conditional} improves the score estimation for image generation by using prompts with no physical meaning. These methods are effective in increasing data volume for small datasets.

The Semantic-aware Data Augmentation (SADA) framework \cite{tan2024semantic} is a new data augmentation approach specifically designed for T2I synthesis. It introduces an implicit text semantic preservation enhancement method to improve text representation in the semantic space and combines it with a specially designed image semantic regularization loss to ensure semantic consistency in the generated images. This approach effectively addresses issues such as semantic mismatches and collapse. Theoretically, it has been shown that implicit text semantic preservation enhancement can improve text-image consistency, while image semantic regularization loss helps maintain semantic integrity in generated images, preventing semantic collapse, and enhancing image quality.

\subsection{Evaluation Metrics}
C2ST (Cumulative 2-Sample Test) \cite{lopez2016revisiting} uses statistical methods to assess the similarity between generated images and real images, focusing primarily on comparing the distributions of the two samples. KPR (Kernel Precision Recall) \cite{kynkaanniemi2019improved} combines kernel methods with precision and recall to analyze the comprehensiveness of the generated content. SSIM (Structural Similarity Index) \cite{wang2004image} evaluates image quality by analyzing the similarity between generated and reference images in terms of luminance, contrast, and structural information. PSNR (Peak Signal-to-Noise Ratio) \cite{hore2010image} assesses image quality by calculating the mean squared error, providing a measure of the signal-to-noise ratio.
The Inception Score (IS) \cite{salimans2016improved} employs a pretrained Inception model to evaluate the quality of generated images by calculating their class distribution and diversity, focusing on the recognizability of the images. The Fréchet Inception Distance (FID) \cite{heusel2017gans} is widely used in image generation tasks and measures quality by calculating the distance between generated images and real images in the feature space.  Modified FID (MiFID) \cite{bai2021training} improves FID by incorporating multiple indicators to assess image quality, including diversity and structural features. The Kernel Inception Distance (KID) \cite{binkowski2018demystifying} enhances the robustness of FID through the use of kernel methods, making it particularly suitable for small sample scenarios.
R-Precision \cite{park2021benchmark} is a classic metric used in information retrieval, evaluating the accuracy of image and text matching by examining the proportion of correct matches. PickScore \cite{kirstain2023pick} is commonly used for GAN models to assess the diversity and quality of generated images, usually achieved through feature extraction and comparison.

The Holistic Evaluation of Text-to-Image Models (HEIM) \cite{lee2024holistic} combines self-curated prompts with MS-COCO data \cite{lin2014microsoft} to propose an evaluation benchmark that includes biases and fairness subcategories, providing a comprehensive assessment of text-to-image generation models.

Using OpenAI's CLIP model, CLIPScore \cite{hessel2021clipscore} calculates the similarity between images and text by embedding them in the same feature space and evaluating their relevance using cosine similarity. CLIP-R-Precision \cite{park2021benchmark} assesses the quality of generated content by calculating the matching accuracy between images and text. NegCLIP \cite{yuksekgonul2022and} introduces negative samples for contrastive learning, enhancing image-text alignment. MosaiCLIP \cite{singh2023coarse} evaluates overall performance by concatenating and comparing multimodal features. CLove \cite{castro2024clove} quantifies the similarity between generated images and their textual descriptions, while CLIP-IOA (CLIP Image Output Assessment) \cite{wang2023exploring} measures the degree of alignment between generated images and text descriptions.

The BLIP model also provides evaluation metrics. BLIP-ITC \cite{li2022blip} uses contrastive learning in image-text pairs to calculate similarity and consistency. BLIP-ITM \cite{li2022blip} focuses on direct image-text matching, evaluating alignment using a contrastive learning framework. BLIP2-ITC \cite{li2023blip} is an improved version of BLIP, designed to enhance image-text matching performance by optimizing contrastive learning for more efficient similarity scoring. BLIP2-ITM \cite{li2023blip} uses enriched image and text features to match, assessing the relevance and consistency of the generated content.

ImageReward \cite{xu2024imagereward} is a human-preference reward model and evaluation resource for text-to-image generation; its score reflects the preferences represented in its training data and should not be treated as a universal or model-family-specific image-quality metric. RAHF (Rich Human Feedback) \cite{liang2024rich} provides finer-grained supervision, such as region-level and keyword-level feedback, for diagnosing generation errors. PAL4VST \cite{zhang2023perceptual} emphasizes perceptual evaluation, focusing on the subjective quality of the generated content.

COBRA \cite{ma2024cobra} evaluates the alignment between objects in images and text using contrastive learning. HPSv1 \cite{wu2023human} and its improved version, HPSv2 \cite{wu2023human}, assess the consistency and quality of generated content through multimodal contrastive learning, with HPSv2 offering enhanced evaluation accuracy. vISORcond \cite{gokhale2022benchmarking} uses conditional contrastive learning to understand the dependencies between images and text, thus assessing the quality of the generated content. VIFIDEL \cite{madhyastha2019vifidel} integrates visual and textual features, using contrastive learning to analyze the relationship between images and text while calculating similarity scores. ViCE \cite{betti2023let} uses contrastive learning to evaluate the embedding space of images and text, optimizing similarity through a contrastive loss function.

ViLBERTScore \cite{lee2020vilbertscore}, based on the ViLBERT model, assesses the correlation between images and text by calculating joint embeddings to evaluate the match between generated content and its descriptions, using contrastive learning to optimize scores. vISOR \cite{gokhale2022benchmarking} evaluates generation quality by comparing the similarity between images and text, calculating matching scores that combine visual and linguistic features, making it well-suited for analyzing complex image-text relationships. vISORN \cite{gokhale2022benchmarking} is an enhanced version of vISOR that incorporates negative sample contrastive learning to improve understanding of image-text relationships, strengthening its ability to distinguish between correct and incorrect matches.

VNLI \cite{yarom2024you} emphasizes reasoning capabilities by assessing the relationships between images and text through natural language inference, analyzing whether the image content supports the text description, typically using inference models to generate evaluation results. The DINO Metric \cite{ruiz2023dreambooth}, based on self-supervised learning, evaluates generation quality by comparing image and text features, using the DINO model to extract features and compute similarity scores. MID \cite{kim2022mutual} assesses the relevance between multimodal features by extracting and comparing them. MINT-IQA \cite{wang2024understanding} evaluates the quality of images and text using inference methods, combining multimodal information to analyze the consistency of generated content.

DreamSim \cite{fu2023dreamsim} simulates the evaluation process to calculate the correlation between generated images and their textual descriptions. TIGEr \cite{jiang2019tiger} analyzes the generation processes of text and images, assessing the quality of generated content by integrating various features and models to provide a comprehensive score. DA-Score \cite{singh2023divide} evaluates both the diversity and accuracy of generated content, quantifying how well the generated results match the textual descriptions. The 3-in-1 metric \cite{huang2023t2i} combines three aspects—generation quality, relevance, and diversity—to provide a holistic assessment.

PA \cite{dinh2022tise} is a metric for evaluating precision and recall in multimodal tasks, focusing on the relevance of generated content. PRD \cite{sajjadi2018assessing} combines precision and recall to assess performance by calculating the overlap between generated content and reference content. I-PRD \cite{kynkaanniemi2019improved} enhances PRD, aiming to improve the accuracy and robustness of the evaluation.

CA \cite{dinh2022tise} examines how well the generated content covers the textual description, determining whether the generated images include all significant elements mentioned in the text by calculating feature coverage. LEIC \cite{cui2018learning} assesses the similarity between images and text by comparing their embedded features using cosine similarity. CIDEr \cite{vedantam2015cider} computes weighted TF-IDF values of N-Grams to evaluate the similarity between generated descriptions and reference descriptions, with a particular emphasis on the consensus of the descriptions.

TIAM \cite{grimal2024tiam} evaluates the alignment between textual descriptions and images by analyzing semantic consistency, with attention to whether specified objects and attributes appear. TIFA \cite{hu2023tifa} takes a question-answering approach: it generates questions from the prompt and uses a visual-question-answering model to check whether the generated image answers them correctly, making it more interpretable than a single embedding similarity, though it depends on question coverage and VQA reliability. GMM-GIOA \cite{kolouri2018sliced} uses a Gaussian mixture formulation to evaluate the distributional differences between the generated and reference images.

VIEScore \cite{ku2023viescore} calculates the similarity between visual and linguistic features by analyzing the embedded characteristics of images and text, providing matching scores suitable for fine-grained assessments in multimodal learning. MQ \cite{gordon2025mismatch} integrates image quality and text quality, consolidating various features to provide a comprehensive evaluation. UniDet \cite{huang2023t2i} assesses the relationship between images and text using unified feature extraction and comparison methods, yielding an overall matching score.

The Aesthetic Predictor \cite{schuhmann2022laion} evaluates the quality of generated content by analyzing the aesthetic characteristics of images, typically using a pretrained aesthetic model to compute aesthetic scores. PAL4InPaint \cite{zhang2022perceptual} specializes in image inpainting tasks, combining multiple features to assess the quality of generated images, including semantic consistency and visual quality.  Semantic Object Accuracy (SOA) \cite{hinz2020semantic} metric is employed for the quantitative evaluation of T2I tasks, utilizing a pre-trained object detector to assess whether the generated images contain the objects mentioned in the corresponding captions.

These advancements in evaluation metrics for T2I tasks will allow for a more comprehensive assessment of the quality and applicability of generated content, ultimately promoting further progress in T2I generation technology.

Table~\ref{table:evaluation} summarizes the metrics discussed above. In the table, ``Human Evaluated'' indicates whether the metric is designed to incorporate or be calibrated against human judgments (as opposed to being computed purely from model features or image statistics), and ``Positive Correlation'' indicates whether the original work reports a positive correlation between the metric and human preference. These two attributes are coarse descriptors drawn from each metric's own reporting rather than results of a unified re-evaluation, and should be read as guidance rather than a ranking.

\begin{table}[!t]
    \centering
    \caption{Evaluation metrics in the T2I task. ``Human Evaluated'' and ``Positive Correlation'' are coarse descriptors taken from each metric's own reporting, not a unified re-evaluation.}
    			\resizebox{1\linewidth}{!}{%
    \begin{tabular}{lccc}
        \toprule
        \textbf{Metric} & \textbf{Year} & \tht{c}{\textbf{Human} \\ \textbf{Evaluated}} & \tht{c}{\textbf{Positive} \\ \textbf{Correlation}} \\
        \midrule
        SSIM \cite{wang2004image} & 2004 & \textcolor{red}{\large $\checkmark$} & \textcolor{red}{\large $\checkmark$} \\
        PSNR \cite{hore2010image} & 2010 & \textcolor{red}{\large $\checkmark$} & \textcolor{red}{\large $\checkmark$} \\
        CIDEr \cite{vedantam2015cider} & 2015 & \textcolor{red}{\large $\checkmark$} & \textcolor{red}{\large $\checkmark$} \\
        IS \cite{salimans2016improved} & 2016 & \textcolor{green}{\large $\times$} & \textcolor{red}{\large $\checkmark$} \\
        C2ST \cite{lopez2016revisiting} & 2016 & \textcolor{green}{\large $\times$} & \textcolor{red}{\large $\checkmark$} \\
        FID \cite{heusel2017gans} & 2017 & \textcolor{green}{\large $\times$} & \textcolor{green}{\large $\times$} \\
        KID \cite{binkowski2018demystifying} & 2018 & \textcolor{green}{\large $\times$} & \textcolor{red}{\large $\checkmark$} \\
        PRD \cite{sajjadi2018assessing} & 2018 & \textcolor{green}{\large $\times$} & \textcolor{green}{\large $\times$} \\
        LEIC \cite{cui2018learning} & 2018 & \textcolor{green}{\large $\times$} & \textcolor{green}{\large $\times$} \\
        GMM-GIOA \cite{kolouri2018sliced} & 2018 & \textcolor{green}{\large $\times$} & \textcolor{green}{\large $\times$} \\
        KPR \cite{kynkaanniemi2019improved} & 2019 & \textcolor{green}{\large $\times$} & \textcolor{red}{\large $\checkmark$} \\
        VIFIDEL \cite{madhyastha2019vifidel} & 2019 & \textcolor{red}{\large $\checkmark$} & \textcolor{red}{\large $\checkmark$} \\
        TIGEr \cite{jiang2019tiger} & 2019 & \textcolor{red}{\large $\checkmark$} & \textcolor{red}{\large $\checkmark$} \\
        I-PRD \cite{kynkaanniemi2019improved} & 2019 & \textcolor{green}{\large $\times$} & \textcolor{green}{\large $\times$} \\
        ViLBERTScore \cite{lee2020vilbertscore} & 2020 & \textcolor{red}{\large $\checkmark$} & \textcolor{red}{\large $\checkmark$} \\
        SOA \cite{hinz2020semantic} & 2020 & \textcolor{red}{\large $\checkmark$} & \textcolor{red}{\large $\checkmark$} \\
        MiFID \cite{bai2021training} & 2021 & \textcolor{green}{\large $\times$} & \textcolor{green}{\large $\times$} \\
        R-Precision \cite{park2021benchmark} & 2021 & \textcolor{red}{\large $\checkmark$} & \textcolor{red}{\large $\checkmark$} \\
        CLIPScore \cite{hessel2021clipscore} & 2021 & \textcolor{red}{\large $\checkmark$} & \textcolor{red}{\large $\checkmark$} \\
        CLIP-R-Precision \cite{park2021benchmark} & 2021 & \textcolor{red}{\large $\checkmark$} & \textcolor{red}{\large $\checkmark$} \\
        NegCLIP \cite{yuksekgonul2022and} & 2022 & \textcolor{green}{\large $\times$} & \textcolor{green}{\large $\times$} \\
        BLIP-ITC \cite{li2022blip} & 2022 & \textcolor{green}{\large $\times$} & \textcolor{red}{\large $\checkmark$} \\
        BLIP-ITM \cite{li2022blip} & 2022 & \textcolor{green}{\large $\times$} & \textcolor{red}{\large $\checkmark$} \\
        vISORcond \cite{gokhale2022benchmarking} & 2022 & \textcolor{red}{\large $\checkmark$} & \textcolor{red}{\large $\checkmark$} \\
        vISOR \cite{gokhale2022benchmarking} & 2022 & \textcolor{red}{\large $\checkmark$} & \textcolor{green}{\large $\times$} \\
        vISORN \cite{gokhale2022benchmarking} & 2022 & \textcolor{red}{\large $\checkmark$} & \textcolor{green}{\large $\times$} \\
        MID \cite{kim2022mutual} & 2022 & \textcolor{red}{\large $\checkmark$} & \textcolor{red}{\large $\checkmark$} \\
        PA \cite{dinh2022tise} & 2022 & \textcolor{red}{\large $\checkmark$} & \textcolor{red}{\large $\checkmark$} \\
        CA \cite{dinh2022tise} & 2022 & \textcolor{red}{\large $\checkmark$} & \textcolor{green}{\large $\times$} \\
        Aesthetic Predictor \cite{schuhmann2022laion} & 2022 & \textcolor{green}{\large $\times$} & \textcolor{red}{\large $\checkmark$} \\
        PAL4InPaint \cite{zhang2022perceptual} & 2022 & \textcolor{green}{\large $\times$} & \textcolor{green}{\large $\times$} \\
        PickScore \cite{kirstain2023pick} & 2023 & \textcolor{red}{\large $\checkmark$} & \textcolor{red}{\large $\checkmark$} \\
        MosaiCLIP \cite{singh2023coarse} & 2023 & \textcolor{green}{\large $\times$} & \textcolor{red}{\large $\checkmark$} \\
        CLIP-IOA \cite{wang2023exploring} & 2023 & \textcolor{red}{\large $\checkmark$} & \textcolor{green}{\large $\times$} \\
        BLIP2-ITC \cite{li2023blip} & 2023 & \textcolor{green}{\large $\times$} & \textcolor{red}{\large $\checkmark$} \\
        BLIP2-ITM \cite{li2023blip} & 2023 & \textcolor{green}{\large $\times$} & \textcolor{red}{\large $\checkmark$} \\
        HPSv1 \cite{wu2023human} & 2023 & \textcolor{red}{\large $\checkmark$} & \textcolor{red}{\large $\checkmark$} \\
        HPSv2 \cite{wu2023human} & 2023 & \textcolor{red}{\large $\checkmark$} & \textcolor{red}{\large $\checkmark$} \\
        ViCE \cite{betti2023let} & 2023 & \textcolor{red}{\large $\checkmark$} & \textcolor{red}{\large $\checkmark$} \\
        DINO Metric \cite{ruiz2023dreambooth} & 2023 & \textcolor{green}{\large $\times$} & \textcolor{red}{\large $\checkmark$} \\
        DreamSim \cite{fu2023dreamsim} & 2023 & \textcolor{red}{\large $\checkmark$} & \textcolor{red}{\large $\checkmark$} \\
        DA-Score \cite{singh2023divide} & 2023 & \textcolor{green}{\large $\times$} & \textcolor{red}{\large $\checkmark$} \\
        3-in-1 \cite{huang2023t2i} & 2023 & \textcolor{red}{\large $\checkmark$} & \textcolor{red}{\large $\checkmark$} \\
        TIFA \cite{hu2023tifa} & 2023 & \textcolor{green}{\large $\times$} & \textcolor{green}{\large $\times$} \\
        VIEScore \cite{ku2023viescore} & 2023 & \textcolor{green}{\large $\times$} & \textcolor{red}{\large $\checkmark$} \\
        UniDet \cite{huang2023t2i} & 2023 & \textcolor{green}{\large $\times$} & \textcolor{green}{\large $\times$} \\
        COBRA \cite{ma2024cobra} & 2024 & \textcolor{red}{\large $\checkmark$} & \textcolor{red}{\large $\checkmark$} \\
        VNLI \cite{yarom2024you} & 2024 & \textcolor{red}{\large $\checkmark$} & \textcolor{green}{\large $\times$} \\
        MINT-IQA \cite{wang2024understanding} & 2024 & \textcolor{red}{\large $\checkmark$} & \textcolor{red}{\large $\checkmark$} \\
        TIAM \cite{grimal2024tiam} & 2024 & \textcolor{red}{\large $\checkmark$} & \textcolor{red}{\large $\checkmark$} \\
        CLove \cite{castro2024clove} & 2024 & \textcolor{green}{\large $\times$} & \textcolor{green}{\large $\times$} \\
        HEIM \cite{lee2024holistic} & 2024 & \textcolor{red}{\large $\checkmark$} & \textcolor{red}{\large $\checkmark$} \\
        MQ \cite{gordon2025mismatch} & 2024 & \textcolor{green}{\large $\times$} & \textcolor{red}{\large $\checkmark$} \\
        \bottomrule
    \end{tabular}}
    \label{table:evaluation}
        \vspace{-0.4cm}
\end{table}

\subsection{Recent Benchmarks and Evaluation Reliability (2025--2026)}
\label{sec:recent-eval}
The metrics above are increasingly complemented by capability- and scenario-specific benchmarks that move beyond a single fidelity or alignment score. Rather than proposing one new number, these resources target particular axes: unified understanding-and-generation, long-prompt and knowledge-grounded generation, creator-oriented quality, and the reliability of automatic judges themselves. Table~\ref{tab:recent_bench} lists representative examples together with what each is designed to measure and its main limitation. We organize them by evaluation target rather than by a single leaderboard, because their inputs, scoring protocols, and model pools differ and their scores are not directly comparable.

\begin{table*}[t]
\centering\small
\caption{Representative 2025--2026 T2I benchmarks, organized by evaluation target. Scores across these resources are not directly comparable because inputs, judges, and model pools differ.}
\label{tab:recent_bench}
\begin{tabular}{p{0.17\textwidth}p{0.30\textwidth}p{0.26\textwidth}p{0.17\textwidth}}
\toprule
Benchmark & What it evaluates & Scoring & Main limitation \\
\midrule
MME-Unify \cite{xie2025mmeunify} & Unified understanding, generation, and mixed-modality interaction & Multi-task, model-based & Mixed-task protocols vary; interaction gains hard to isolate \\
KontextBench \cite{labs2025flux} & In-context generation and single-/multi-turn editing (1{,}026 image-prompt pairs) & Human and automatic & Introduced with a specific model family \\
LongT2IBench \cite{yang2025longtibench} & Long-prompt generation with graph-structured annotations & Structured, automatic & Focused on long-prompt setting \\
WISE \cite{niu2025wise} & World-knowledge-informed semantic correctness & Knowledge-grounded checks & Depends on knowledge coverage \\
GenExam \cite{wang2025genexam} & Multidisciplinary ``exam-style'' prompts & Rubric/answer-based & Subject coverage is finite \\
Qwen-Image-Bench \cite{li2026qwenimagebench} & Creator-oriented real-world fidelity and creative generation & Human and model-based & Creator pool and judge may share source bias \\
FineGRAIN \cite{hayes2025finegrain} & Failure modes of vision-language-model judges & Diagnostic, judge-focused & Audits judges, not generators directly \\
\bottomrule
\end{tabular}
    \vspace{-0.4cm}
\end{table*}

Two reliability concerns deserve emphasis. First, many recent benchmarks use large vision-language models as automatic judges, yet such judges have characteristic failure modes \cite{hayes2025finegrain}; when the same family of model is used both to generate and to judge, shared biases can inflate scores. Second, a preference-reward score such as ImageReward \cite{xu2024imagereward} or PickScore \cite{kirstain2023pick} reflects the preferences in its training data and is not a universal measure of quality, diversity, or factual correctness. Consequently, a single metric or benchmark is insufficient; reported results should state the input prompts, the judge, and the model pool, and preference, compositional, editing, and safety dimensions should be assessed separately rather than merged into one ranking.
\section{Applications}
\label{section6}
In this digital age, artificial intelligence (AI) has become a major driving force behind technological innovation. In particular, T2I tools have emerged as a crucial element of AI-generated content (AIGC) \cite{wu2023ai}. Section \ref{section3} discusses the latest advancements in T2I technology. Based on this research, T2I technology has shown broad practical applications in various fields, significantly improving creativity and production efficiency.

\begin{figure}[!t]    
	\centering
    \includegraphics[width=1\linewidth]{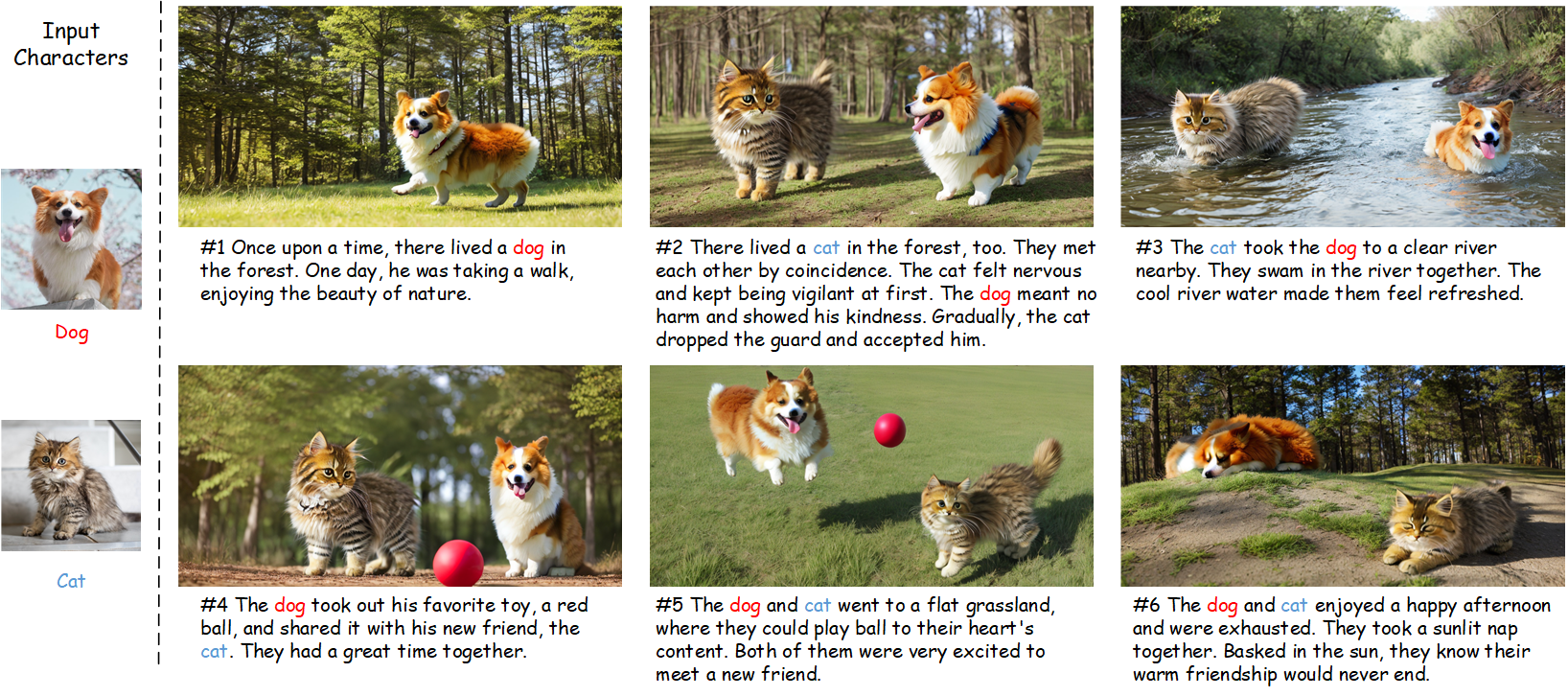} 
	\caption{Example storytelling images generated by the method AutoStory\cite{wang2023autostory}. it can generate text-aligned, identity-consistent, and high-quality story images from user-input stories and characters (the dog and cat on the left, specified by about 5 images per character), without additional
inputs like sketches. } 
	\label{autostory}
        \vspace{-0.4cm}
\end{figure}

\subsection{T2I for Artistic Creation}
In artistic creation, models such as DALL-E \cite{ramesh2021zero}, Stable Diffusion \cite{rombach2022high}, Imagen \cite{saharia2022photorealistic}, and DreamBooth \cite{ruiz2023dreambooth} let artists turn a textual description into images and explore styles rapidly, building directly on the controllable and personalized generation techniques of Section~\ref{section4}. For narrative content, tools such as TaleCraft \cite{gong2023talecrafter}, Custom Diffusion \cite{kumari2023multi}, Paint-by-Example \cite{yang2023paint}, Make-a-Story \cite{rahman2023make}, and AutoStory \cite{wang2023autostory} rely on subject personalization and identity preservation to keep characters consistent across panels (Figure~\ref{autostory}). The same pipeline supports concept art for game design and immersive VR/AR backdrops. These uses can speed up early iteration and exploration, but any quantitative claim about labor or cost savings would require a task-specific study.

\subsection{T2I for E-Commerce}
\begin{figure}[!t]  
	\centering
    \includegraphics[width=1\linewidth]{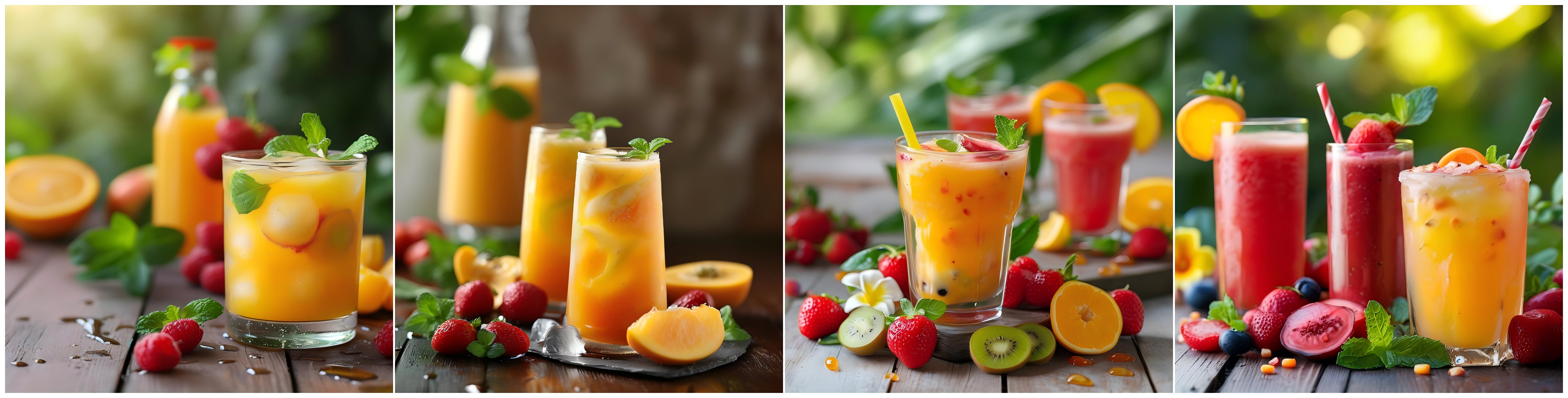} 
	\caption{Example images generated by Kolors \cite{kolors2024github}, with the text description ``refreshing summer juice drink''. Kolors can produce a variety of clear image results. } 
	\label{kolors1}
        \vspace{-0.4cm}
\end{figure}

Companies use tools such as Stable Diffusion \cite{rombach2022high} and DALL-E \cite{ramesh2021zero} to create product visuals from descriptions; Figure \ref{kolors1} shows an example produced with the Kolors \cite{kolors2024github} platform for the prompt ``refreshing summer juice drink''.

Two technical capabilities are especially relevant in this setting. First, inpainting together with subject and identity preservation (Section~\ref{section4}) allows the same garment to be shown on diverse virtual models for cross-border fashion, supporting early visual prototyping rather than replacing photoshoots outright. Second, spatial control and text rendering support packaging and advertisement design, for example via ControlNet \cite{zhang2023adding} with layout templates, enabling rapid iteration over design alternatives. T2I thus offers faster visual prototyping in advertising and marketing; claims about conversion, market response, or commercial impact, however, require evidence from the relevant workflow and user population, along with checks for product identity, garment geometry, and cultural representation.

\subsection{T2I for Education and Healthcare}
\begin{figure}[!t]     
	\centering
    \includegraphics[width=1\linewidth]{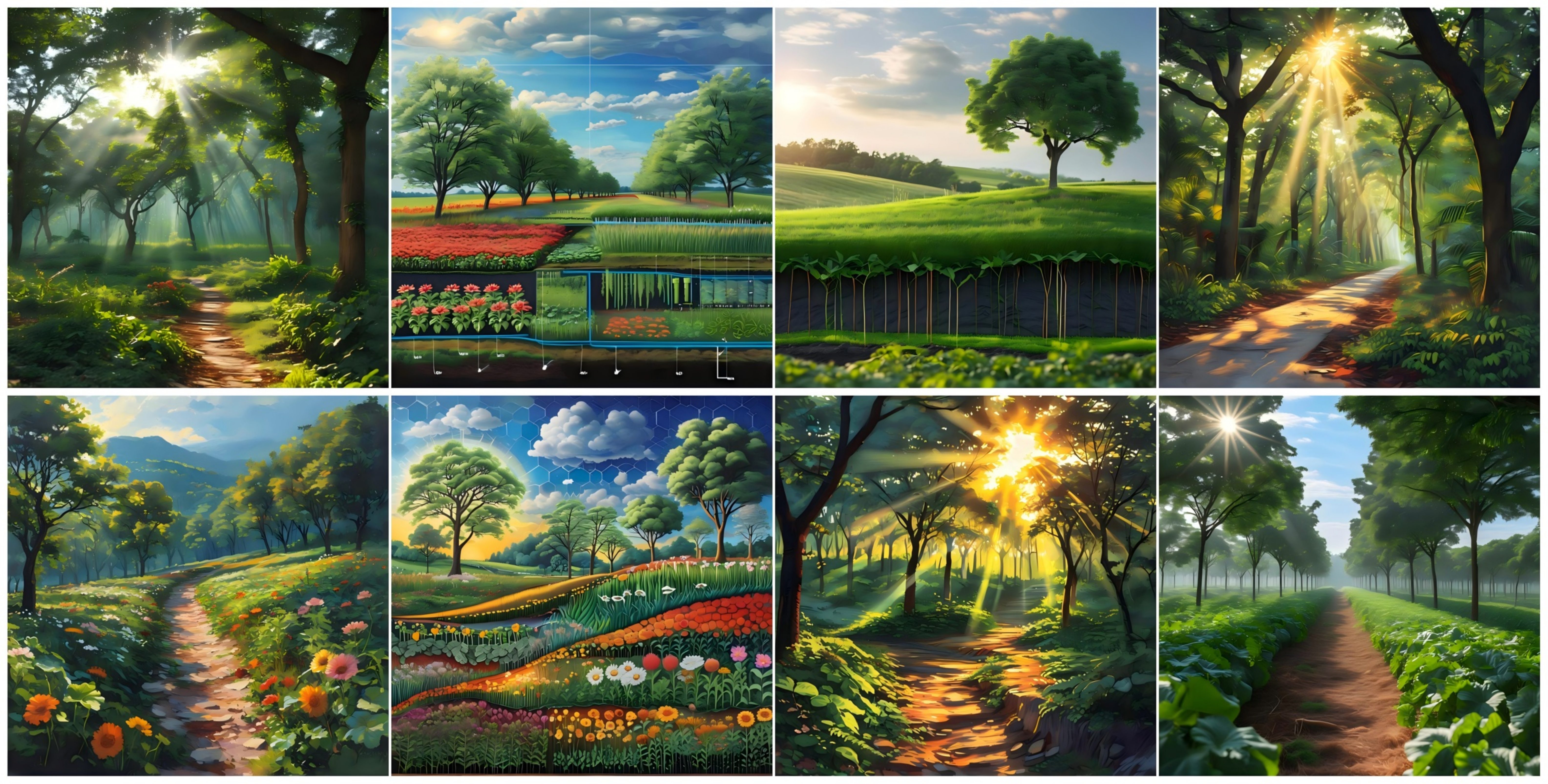} 
	\caption{Given Kolors \cite{kolors2024github} the content ``explain the process of photosynthesis'', it can generate a series of images related to photosynthesis to enhance educational significance.}  
	\label{kolors2}
        \vspace{-0.4cm}
\end{figure}

In education, a T2I system such as Kolors \cite{kolors2024github} can turn a prompt like ``explain the process of photosynthesis'' into illustrative material (Figure \ref{kolors2}), whereas CLIP-style models \cite{radford2021learning} are vision-language representations used for retrieval and alignment rather than image generators. Such illustrations may aid explanation, but improved learning outcomes must be established by a dedicated educational study rather than inferred from a generated example.

In healthcare, generative models have been explored for educational or research visualizations of medical content \cite{skandarani2023gans,zhang2017stackgan,gou2020segattngan}. However, a general T2I sample is not a validated diagnostic image: any use in diagnosis or treatment planning requires domain-specific data, clinical validation, and expert review, which lie outside the evidence of this survey.

\begin{figure}[!t]   
	\centering
    \includegraphics[width=1\linewidth]{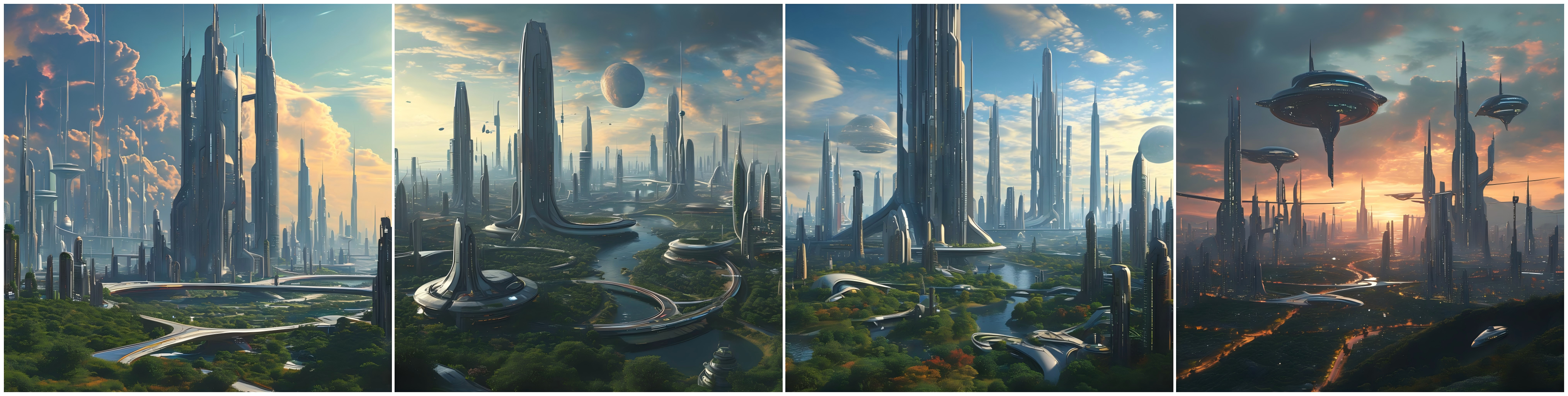} 
	\caption{Given the text description ``a futuristic city'', a series of images can be generated to capture users' attention. This example was generated by Kolors \cite{kolors2024github}.}  
	\label{kolors3}
        \vspace{-0.4cm}
\end{figure}

\subsection{T2I for Social Media}
On social media, creators use tools such as Stable Diffusion 3 \cite{esser2024scaling}, DALLE-3 \cite{betker2023improving}, and Kolors to turn text into personalized visuals that complement posts, articles, or videos (Figure \ref{kolors3}). The enabling techniques here are identity and character consistency, instruction-based editing, and publishing-time safety (Section~\ref{section4}). T2I can lower the effort of producing on-brand visual content, but its effect on engagement remains an empirical question that depends on the audience and platform.

Overall, T2I is reshaping creative practice across art, advertising, game development, education, and social media by providing flexibility and faster visual iteration. Its effects on efficiency, engagement, and commercial outcomes, however, remain empirical questions that depend on the workflow, the users, and the safeguards in each setting.

\section{Challenges and Outlook}
\label{section7}

\subsection{Challenges}
Despite the significant potential of T2I technology in various fields, it still faces several challenges.

\subsubsection{Data and Computational Resource Challenges}
Training large T2I models depends on massive text-image corpora, which carry demographic and cultural biases: models can under-represent minority groups or reproduce stereotypes (for example, defaulting to male figures for ``a professional''), raising fairness concerns \cite{d2024openbias,li2024self,wan2024survey,vice2023quantifying}. Most systems assume English prompts, disadvantaging other languages and losing cultural nuance. Mitigations include more balanced and multilingual data \cite{jha2024visage,friedrich2024multilingual,lee2024holistic} and bias-correction techniques \cite{palash2021fine}, but these remain partial solutions rather than guarantees.

Compute is the second constraint. Leading models are trained on hundreds of millions of image-text pairs \cite{ramesh2022hierarchical,nichol2021glide,saharia2022photorealistic}, concentrating capability in a few large organizations and limiting participation by smaller institutions. At inference, large backbones are difficult to deploy on edge or mobile devices, motivating compression and few-step sampling---pruning, quantization, distillation, and the efficiency methods of Section~\ref{Performance} \cite{patel2024eclipse,chen2023pixart}. Progress therefore depends as much on open, efficient, and reproducible recipes as on raw scale.

\subsubsection{Copyright and Ethical Issues}
Copyright remains unresolved: training on unlicensed content and the lack of data-provenance transparency make infringement hard to attribute or prevent \cite{kim2024automatic,li2024va3}. Technical protections and concept-erasure methods are covered in Sections~\ref{Safety} and~\ref{Copyright}; here we only note that they \emph{mitigate} but do not \emph{settle} the underlying legal questions, which lie outside the scope of this survey. Related risks---deepfakes, misinformation, and privacy violations---call for content moderation, provenance or watermarking, and transparent usage policies, alongside the bias mitigations discussed above \cite{liu2024countering,wang2024simac,zhang2024steerdiff,d2024openbias}.

\subsubsection{Understanding Complex Text and Generative Diversity}
Current models still struggle with the harder parts of language understanding: attribute binding, counting, spatial relations, long or compositional prompts, and implicit or cultural references. These failures connect directly to the consistency techniques of Section~\ref{Consistency} and to the compositional benchmarks of Section~\ref{section5}, which localize \emph{where} errors occur rather than reporting a single score. A second issue is limited output diversity, which can amplify dataset biases. Addressing both requires better-balanced data and training signals \cite{schuhmann2022laion,chatterjee2025getting,liang2024rich} together with evaluation that measures compositional competence and diversity separately, rather than assuming that a higher overall preference score implies broader capability.

\subsection{Outlook}

\subsubsection{Social Responsibility and Accessibility}
As T2I is adopted more widely, responsible development---bias and safety auditing, content provenance, and transparent usage standards---should be a design requirement rather than an afterthought. In parallel, falling inference cost, open training recipes, and reproducible evaluation are likely to broaden access for smaller creators and institutions. The open question is how to make models cheaper and more auditable without sacrificing safety or reproducibility.

\subsubsection{Beyond Superficial Realism: the ``AI-generated Look''}
Even high-fidelity outputs often retain a recognizable ``AI'' appearance. Rather than treating this as a single quality score, we suggest decomposing it into measurable factors: local artifacts (hands, text, repeated textures), implausible lighting and material response, global composition and perspective errors, and stylistic homogenization across samples. Each factor calls for targeted data and metrics; importantly, current preference rewards do not by themselves certify that the ``AI look'' has been removed, and a higher reward can even coexist with reduced diversity. This is an analytical agenda for future evaluation rather than a solved problem with an agreed metric.

\subsubsection{Stronger Semantic Reasoning and Unified Feedback}
A central direction is tighter coupling between generation and reasoning. Two mechanisms should be distinguished: \emph{external} planning or prompt rewriting by a separate language model (Section~\ref{LLM-Assisted}), and \emph{internal} reasoning in unified understanding-and-generation models (Section~\ref{sec:unified}), where one model both interprets and renders. The latter is promising because the understanding component could, in principle, check and correct the generated image; however, as noted in Section~\ref{sec:unified}, shared competence is not yet evidence of cross-task benefit, and mixed-modality benchmarks such as MME-Unify \cite{xie2025mmeunify} are needed to test it. The key open problem is a reliable feedback loop in which reasoning, generation, and evaluation improve together without collapsing onto a single judge. This T2I-specific question, rather than an open-ended ``agent'' agenda, is what we see as most consequential for the field.

As technology serves humanity, by addressing these challenges and opportunities, T2I is set to play an even more significant role across creative and practical domains.

\section*{Acknowledgements}
This work is supported by the Fundamental Research Funds for the Central Universities, Peking University.

\section*{Data Availability Statement}
The data supporting the findings of this study are available from the corresponding author upon reasonable request.

\small
\bibliographystyle{plain}
\bibliography{ref}

@article{watkins1989learning,
  title={Learning from delayed rewards},
  author={Watkins, Christopher John Cornish Hellaby},
  year={1989},
  publisher={King's College, Cambridge United Kingdom}
}

@inproceedings{bengio2000neural,
  title={A neural probabilistic language model},
  author={Bengio, Yoshua and Ducharme, R{\'e}jean and Vincent, Pascal},
  booktitle={NeurIPS},
  year={2000}
}

@article{wang2004image,
  title={Image quality assessment: from error visibility to structural similarity},
  author={Wang, Zhou and Bovik, Alan C and Sheikh, Hamid R and Simoncelli, Eero P},
  journal={IEEE TIP},
  year={2004}
}

@book{goertzel2007artificial,
  title={Artificial general intelligence},
  author={Goertzel, Ben and Pennachin, Cassio},
  volume={2},
  year={2007},
  publisher={Springer}
}

@inproceedings{nilsback2008automated,
  title={Automated flower classification over a large number of classes},
  author={Nilsback, Maria-Elena and Zisserman, Andrew},
  booktitle={Sixth Indian conference on computer vision, graphics \& image processing},
  pages={722--729},
  year={2008}
}

@book{marr2010vision,
  title={Vision: A computational investigation into the human representation and processing of visual information},
  author={Marr, David},
  year={2010},
  publisher={MIT press}
}

@inproceedings{rashtchian2010collecting,
  title={Collecting image annotations using amazon’s mechanical turk},
  author={Rashtchian, Cyrus and Young, Peter and Hodosh, Micah and Hockenmaier, Julia},
  booktitle={Proceedings of the NAACL HLT 2010 workshop on creating speech and language data with Amazon’s Mechanical Turk},
  pages={139--147},
  year={2010}
}

@inproceedings{hore2010image,
  title={Image quality metrics: PSNR vs. SSIM},
  author={Hore, Alain and Ziou, Djemel},
  booktitle={ICPR},
  year={2010}
}

@inproceedings{ordonez2011im2text,
  title={Im2text: Describing images using 1 million captioned photographs},
  author={Ordonez, Vicente and Kulkarni, Girish and Berg, Tamara},
  booktitle={NeurIPS},
  year={2011}
}

@book{bartholomew2011latent,
  title={Latent variable models and factor analysis: A unified approach},
  author={Bartholomew, David J and Knott, Martin and Moustaki, Irini},
  year={2011},
  publisher={John Wiley \& Sons}
}

@inproceedings{krizhevsky2012imagenet,
  title={Imagenet classification with deep convolutional neural networks},
  author={Krizhevsky, Alex and Sutskever, Ilya and Hinton, Geoffrey E},
  booktitle={NeurIPS},
  year={2012}
}

@article{kingma2013auto,
  title={Auto-encoding variational bayes},
  author={Kingma, Diederik P},
  journal={arXiv preprint arXiv:1312.6114},
  year={2013}
}

@article{hodosh2013framing,
  title={Framing image description as a ranking task: Data, models and evaluation metrics},
  author={Hodosh, Micah and Young, Peter and Hockenmaier, Julia},
  journal={Journal of Artificial Intelligence Research},
  volume={47},
  pages={853--899},
  year={2013}
}

@inproceedings{goodfellow2014generative,
  title={Generative adversarial nets},
  author={Goodfellow, Ian and Pouget-Abadie, Jean and Mirza, Mehdi and Xu, Bing and Warde-Farley, David and Ozair, Sherjil and Courville, Aaron and Bengio, Yoshua},
  booktitle={NeurIPS},
  year={2014}
}

@article{mirza2014conditional,
  title={Conditional generative adversarial nets},
  author={Mirza, Mehdi},
  journal={arXiv preprint arXiv:1411.1784},
  year={2014}
}

@article{goertzel2014artificial,
  title={Artificial general intelligence: concept, state of the art, and future prospects},
  author={Goertzel, Ben},
  journal={Journal of Artificial General Intelligence},
  volume={5},
  number={1},
  pages={1},
  year={2014},
  publisher={De Gruyter Poland}
}

@article{chen2015microsoft,
  title={Microsoft coco captions: Data collection and evaluation server},
  author={Chen, Xinlei and Fang, Hao and Lin, Tsung-Yi and Vedantam, Ramakrishna and Gupta, Saurabh and Doll{\'a}r, Piotr and Zitnick, C Lawrence},
  journal={arXiv preprint arXiv:1504.00325},
  year={2015}
}

@article{young2014image,
  title={From image descriptions to visual denotations: New similarity metrics for semantic inference over event descriptions},
  author={Young, Peter and Lai, Alice and Hodosh, Micah and Hockenmaier, Julia},
  journal={Transactions of the Association for Computational Linguistics},
  volume={2},
  pages={67--78},
  year={2014}
}

@inproceedings{lin2014microsoft,
  title={Microsoft coco: Common objects in context},
  author={Lin, Tsung-Yi and Maire, Michael and Belongie, Serge and Hays, James and Perona, Pietro and Ramanan, Deva and Doll{\'a}r, Piotr and Zitnick, C Lawrence},
  booktitle={ECCV},
  year={2014}
}

@inproceedings{vinyals2015show,
  title={Show and tell: A neural image caption generator},
  author={Vinyals, Oriol and Toshev, Alexander and Bengio, Samy and Erhan, Dumitru},
  booktitle={CVPR},
  year={2015}
}

@inproceedings{antol2015vqa,
  title={Vqa: Visual question answering},
  author={Antol, Stanislaw and Agrawal, Aishwarya and Lu, Jiasen and Mitchell, Margaret and Batra, Dhruv and Zitnick, C Lawrence and Parikh, Devi},
  booktitle={ICCV},
  year={2015}
}

@article{lecun2015deep,
  title={Deep learning},
  author={LeCun, Yann and Bengio, Yoshua and Hinton, Geoffrey},
  journal={nature},
  volume={521},
  number={7553},
  pages={436--444},
  year={2015}
}

@inproceedings{plummer2015flickr30k,
  title={Flickr30k entities: Collecting region-to-phrase correspondences for richer image-to-sentence models},
  author={Plummer, Bryan A and Wang, Liwei and Cervantes, Chris M and Caicedo, Juan C and Hockenmaier, Julia and Lazebnik, Svetlana},
  booktitle={ICCV},
  year={2015}
}

@inproceedings{ronneberger2015u,
  title={U-net: Convolutional networks for biomedical image segmentation},
  author={Ronneberger, Olaf and Fischer, Philipp and Brox, Thomas},
  booktitle={MICCAI},
  year={2015}
}

@article{dhanachandra2015image,
  title={Image segmentation using K-means clustering algorithm and subtractive clustering algorithm},
  author={Dhanachandra, Nameirakpam and Manglem, Khumanthem and Chanu, Yambem Jina},
  journal={Procedia Computer Science},
  volume={54},
  pages={764--771},
  year={2015},
  publisher={Elsevier}
}

@inproceedings{vedantam2015cider,
  title={Cider: Consensus-based image description evaluation},
  author={Vedantam, Ramakrishna and Lawrence Zitnick, C and Parikh, Devi},
  booktitle={CVPR},
  year={2015}
}

@inproceedings{romera2015embarrassingly,
  title={An embarrassingly simple approach to zero-shot learning},
  author={Romera-Paredes, Bernardino and Torr, Philip},
  booktitle={ICML},
  year={2015}
}

@inproceedings{sohl2015deep,
  title={Deep unsupervised learning using nonequilibrium thermodynamics},
  author={Sohl-Dickstein, Jascha and Weiss, Eric and Maheswaranathan, Niru and Ganguli, Surya},
  booktitle={ICML},
  year={2015}
}

@inproceedings{mao2016training,
  title={Training and evaluating multimodal word embeddings with large-scale web annotated images},
  author={Mao, Junhua and Xu, Jiajing and Jing, Kevin and Yuille, Alan L},
  booktitle={NeurIPS},
  year={2016}
}

@article{lopez2016revisiting,
  title={Revisiting classifier two-sample tests},
  author={Lopez-Paz, David and Oquab, Maxime},
  journal={arXiv preprint arXiv:1610.06545},
  year={2016}
}

@inproceedings{van2016conditional,
  title={Conditional image generation with pixelcnn decoders},
  author={Van den Oord, Aaron and Kalchbrenner, Nal and Espeholt, Lasse and Vinyals, Oriol and Graves, Alex and others},
  booktitle={NeurIPS},
  year={2016}
}

@inproceedings{salimans2016improved,
  title={Improved techniques for training gans},
  author={Salimans, Tim and Goodfellow, Ian and Zaremba, Wojciech and Cheung, Vicki and Radford, Alec and Chen, Xi},
  booktitle={NeurIPS},
  year={2016}
}

@inproceedings{yan2016attribute2image,
  title={Attribute2image: Conditional image generation from visual attributes},
  author={Yan, Xinchen and Yang, Jimei and Sohn, Kihyuk and Lee, Honglak},
  booktitle={ECCV},
  year={2016}
}

@article{mansimov2015generating,
  title={Generating images from captions with attention},
  author={Mansimov, Elman and Parisotto, Emilio and Ba, Jimmy Lei and Salakhutdinov, Ruslan},
  journal={arXiv preprint arXiv:1511.02793},
  year={2015}
}

@article{rolfe2016discrete,
  title={Discrete variational autoencoders},
  author={Rolfe, Jason Tyler},
  journal={arXiv preprint arXiv:1609.02200},
  year={2016}
}

@inproceedings{reed2016generative,
  title={Generative adversarial text to image synthesis},
  author={Reed, Scott and Akata, Zeynep and Yan, Xinchen and Logeswaran, Lajanugen and Schiele, Bernt and Lee, Honglak},
  booktitle={ICML},
  year={2016}
}

@inproceedings{croce2020gan,
  title={GAN-BERT: Generative adversarial learning for robust text classification with a bunch of labeled examples},
  author={Croce, Danilo and Castellucci, Giuseppe and Basili, Roberto},
  booktitle={ACL},
  pages={2114--2119},
  year={2020}
}

@inproceedings{xu2022predict,
  title={Predict, prevent, and evaluate: Disentangled text-driven image manipulation empowered by pre-trained vision-language model},
  author={Xu, Zipeng and Lin, Tianwei and Tang, Hao and Li, Fu and He, Dongliang and Sebe, Nicu and Timofte, Radu and Van Gool, Luc and Ding, Errui},
  booktitle={CVPR},
  year={2022}
}

@article{wu2022cross_tmm,
  title={Cross-view panorama image synthesis},
  author={Wu, Songsong and Tang, Hao and Jing, Xiao-Yuan and Zhao, Haifeng and Qian, Jianjun and Sebe, Nicu and Yan, Yan},
  journal={IEEE TMM},
  volume={25},
  pages={3546--3559},
  year={2022}
}

@inproceedings{tao2023net,
  title={De-net: Dynamic text-guided image editing adversarial networks},
  author={Tao, Ming and Bao, Bing-Kun and Tang, Hao and Wu, Fei and Wei, Longhui and Tian, Qi},
  booktitle={AAAI},
  year={2023}
}

@inproceedings{tang2023graph,
  title={Graph transformer gans for graph-constrained house generation},
  author={Tang, Hao and Zhang, Zhenyu and Shi, Humphrey and Li, Bo and Shao, Ling and Sebe, Nicu and Timofte, Radu and Van Gool, Luc},
  booktitle={CVPR},
  year={2023}
}

@inproceedings{zhang2024ssr,
  title={Ssr-encoder: Encoding selective subject representation for subject-driven generation},
  author={Zhang, Yuxuan and Song, Yiren and Liu, Jiaming and Wang, Rui and Yu, Jinpeng and Tang, Hao and Li, Huaxia and Tang, Xu and Hu, Yao and Pan, Han and others},
  booktitle={CVPR},
  year={2024}
}

@article{tang2023edge,
  title={Edge guided GANs with multi-scale contrastive learning for semantic image synthesis},
  author={Tang, Hao and Sun, Guolei and Sebe, Nicu and Van Gool, Luc},
  journal={IEEE TPAMI},
  volume={45},
  number={12},
  pages={14435--14452},
  year={2023}
}

@article{tang2024graph,
  title={Graph Transformer GANs with Graph Masked Modeling for Architectural Layout Generation},
  author={Tang, Hao and Shao, Ling and Sebe, Nicu and Van Gool, Luc},
  journal={IEEE TPAMI},
  year={2024}
}

@inproceedings{zhang2017stackgan,
  title={Stackgan: Text to photo-realistic image synthesis with stacked generative adversarial networks},
  author={Zhang, Han and Xu, Tao and Li, Hongsheng and Zhang, Shaoting and Wang, Xiaogang and Huang, Xiaolei and Metaxas, Dimitris N},
  booktitle={ICCV},
  year={2017}
}

@article{zhang2017mixup,
  title={mixup: Beyond empirical risk minimization},
  author={Zhang, Hongyi},
  journal={arXiv preprint arXiv:1710.09412},
  year={2017}
}

@inproceedings{heusel2017gans,
  title={Gans trained by a two time-scale update rule converge to a local nash equilibrium},
  author={Heusel, Martin and Ramsauer, Hubert and Unterthiner, Thomas and Nessler, Bernhard and Hochreiter, Sepp},
  booktitle={NeurIPS},
  year={2017}
}

@article{dong2017learning,
  title={Learning text to image synthesis with textual data augmentation},
  author={Dong, Hao and Zhang, Jingqing and McIlwraith, Douglas and Guo, Yike},
  journal={arXiv preprint arXiv:1703.06676},
  volume={2},
  year={2017}
}

@inproceedings{goyal2017making,
  title={Making the v in vqa matter: Elevating the role of image understanding in visual question answering},
  author={Goyal, Yash and Khot, Tejas and Summers-Stay, Douglas and Batra, Dhruv and Parikh, Devi},
  booktitle={CVPR},
  year={2017}
}

@inproceedings{vaswani2017attention,
  title={Attention is all you need},
  author={Vaswani, A},
  booktitle={NeurIPS},
  year={2017}
}

@inproceedings{van2017neural,
  title={Neural discrete representation learning},
  author={Van Den Oord, Aaron and Vinyals, Oriol and others},
  booktitle={NeurIPS},
  year={2017}
}

@article{krishna2017visual,
  title={Visual genome: Connecting language and vision using crowdsourced dense image annotations},
  author={Krishna, Ranjay and Zhu, Yuke and Groth, Oliver and Johnson, Justin and Hata, Kenji and Kravitz, Joshua and Chen, Stephanie and Kalantidis, Yannis and Li, Li-Jia and Shamma, David A and others},
  journal={International journal of computer vision},
  volume={123},
  pages={32--73},
  year={2017},
  publisher={Springer}
}

@article{dash2017tac,
  title={Tac-gan-text conditioned auxiliary classifier generative adversarial network},
  author={Dash, Ayushman and Gamboa, John Cristian Borges and Ahmed, Sheraz and Liwicki, Marcus and Afzal, Muhammad Zeshan},
  journal={arXiv preprint arXiv:1703.06412},
  year={2017}
}

@inproceedings{cui2018learning,
  title={Learning to evaluate image captioning},
  author={Cui, Yin and Yang, Guandao and Veit, Andreas and Huang, Xun and Belongie, Serge},
  booktitle={CVPR},
  year={2018}
}

@inproceedings{kolouri2018sliced,
  title={Sliced wasserstein distance for learning gaussian mixture models},
  author={Kolouri, Soheil and Rohde, Gustavo K and Hoffmann, Heiko},
  booktitle={CVPR},
  year={2018}
}

@inproceedings{zhang2018photographic,
  title={Photographic text-to-image synthesis with a hierarchically-nested adversarial network},
  author={Zhang, Zizhao and Xie, Yuanpu and Yang, Lin},
  booktitle={CVPR},
  year={2018}
}

@article{binkowski2018demystifying,
  title={Demystifying mmd gans},
  author={Bi{\'n}kowski, Miko{\l}aj and Sutherland, Danica J and Arbel, Michael and Gretton, Arthur},
  journal={arXiv preprint arXiv:1801.01401},
  year={2018}
}

@article{radford2018improving,
  title={Improving language understanding by generative pre-training},
  author={Radford, Alec},
  year={2018}
}

@article{brock2018large,
  title={Large Scale GAN Training for High Fidelity Natural Image Synthesis},
  author={Brock, Andrew},
  journal={arXiv preprint arXiv:1809.11096},
  year={2018}
}

@inproceedings{xu2018attngan,
  title={Attngan: Fine-grained text to image generation with attentional generative adversarial networks},
  author={Xu, Tao and Zhang, Pengchuan and Huang, Qiuyuan and Zhang, Han and Gan, Zhe and Huang, Xiaolei and He, Xiaodong},
  booktitle={CVPR},
  year={2018}
}

@inproceedings{sajjadi2018assessing,
  title={Assessing generative models via precision and recall},
  author={Sajjadi, Mehdi SM and Bachem, Olivier and Lucic, Mario and Bousquet, Olivier and Gelly, Sylvain},
  booktitle={NeurIPS},
  volume={31},
  year={2018}
}

@article{devlin2018bert,
  title={Bert: Pre-training of deep bidirectional transformers for language understanding},
  author={Devlin, Jacob},
  journal={arXiv preprint arXiv:1810.04805},
  year={2018}
}

@article{zhang2018stackgan++,
  title={Stackgan++: Realistic image synthesis with stacked generative adversarial networks},
  author={Zhang, Han and Xu, Tao and Li, Hongsheng and Zhang, Shaoting and Wang, Xiaogang and Huang, Xiaolei and Metaxas, Dimitris N},
  journal={IEEE TPAMI},
  volume={41},
  number={8},
  pages={1947--1962},
  year={2018}
}

@article{kudo2018sentencepiece,
  title={Sentencepiece: A simple and language independent subword tokenizer and detokenizer for neural text processing},
  author={Kudo, T},
  journal={arXiv preprint arXiv:1808.06226},
  year={2018}
}

@article{liu2018large,
  title={Large-scale celebfaces attributes (celeba) dataset},
  author={Liu, Ziwei and Luo, Ping and Wang, Xiaogang and Tang, Xiaoou},
  journal={Retrieved August},
  volume={15},
  number={2018},
  pages={11},
  year={2018}
}

@inproceedings{sharma2018conceptual,
  title={Conceptual captions: A cleaned, hypernymed, image alt-text dataset for automatic image captioning},
  author={Sharma, Piyush and Ding, Nan and Goodman, Sebastian and Soricut, Radu},
  booktitle={ACL},
  pages={2556--2565},
  year={2018}
}

@inproceedings{bodla2018semi,
  title={Semi-supervised FusedGAN for conditional image generation},
  author={Bodla, Navaneeth and Hua, Gang and Chellappa, Rama},
  booktitle={ECCV},
  year={2018}
}

@inproceedings{kynkaanniemi2019improved,
  title={Improved precision and recall metric for assessing generative models},
  author={Kynk{\"a}{\"a}nniemi, Tuomas and Karras, Tero and Laine, Samuli and Lehtinen, Jaakko and Aila, Timo},
  booktitle={NeurIPS},
  year={2019}
}

@article{madhyastha2019vifidel,
  title={VIFIDEL: Evaluating the visual fidelity of image descriptions},
  author={Madhyastha, Pranava and Wang, Josiah and Specia, Lucia},
  journal={arXiv preprint arXiv:1907.09340},
  year={2019}
}

@inproceedings{zhu2019dm,
  title={Dm-gan: Dynamic memory generative adversarial networks for text-to-image synthesis},
  author={Zhu, Minfeng and Pan, Pingbo and Chen, Wei and Yang, Yi},
  booktitle={CVPR},
  year={2019}
}

@inproceedings{li2019object,
  title={Object-driven text-to-image synthesis via adversarial training},
  author={Li, Wenbo and Zhang, Pengchuan and Zhang, Lei and Huang, Qiuyuan and He, Xiaodong and Lyu, Siwei and Gao, Jianfeng},
  booktitle={CVPR},
  year={2019}
}

@article{jiang2019tiger,
  title={Tiger: Text-to-image grounding for image caption evaluation},
  author={Jiang, Ming and Huang, Qiuyuan and Zhang, Lei and Wang, Xin and Zhang, Pengchuan and Gan, Zhe and Diesner, Jana and Gao, Jianfeng},
  journal={arXiv preprint arXiv:1909.02050},
  year={2019}
}

@inproceedings{bachman2019learning,
  title={Learning representations by maximizing mutual information across views},
  author={Bachman, Philip and Hjelm, R Devon and Buchwalter, William},
  booktitle={NeurIPS},
  year={2019}
}

@inproceedings{agrawal2019nocaps,
  title={Nocaps: Novel object captioning at scale},
  author={Agrawal, Harsh and Desai, Karan and Wang, Yufei and Chen, Xinlei and Jain, Rishabh and Johnson, Mark and Batra, Dhruv and Parikh, Devi and Lee, Stefan and Anderson, Peter},
  booktitle={ICCV},
  year={2019}
}

@article{pei2019towards,
  title={Towards artificial general intelligence with hybrid Tianjic chip architecture},
  author={Pei, Jing and Deng, Lei and Song, Sen and Zhao, Mingguo and Zhang, Youhui and Wu, Shuang and Wang, Guanrui and Zou, Zhe and Wu, Zhenzhi and He, Wei and others},
  journal={Nature},
  volume={572},
  number={7767},
  pages={106--111},
  year={2019},
  publisher={Nature Publishing Group UK London}
}

@inproceedings{cha2019adversarial,
  title={Adversarial learning of semantic relevance in text to image synthesis},
  author={Cha, Miriam and Gwon, Youngjune L and Kung, HT},
  booktitle={AAAI},
  year={2019}
}

@inproceedings{qiao2019mirrorgan,
  title={Mirrorgan: Learning text-to-image generation by redescription},
  author={Qiao, Tingting and Zhang, Jing and Xu, Duanqing and Tao, Dacheng},
  booktitle={CVPR},
  year={2019}
}

@inproceedings{zellers2019recognition,
  title={From recognition to cognition: Visual commonsense reasoning},
  author={Zellers, Rowan and Bisk, Yonatan and Farhadi, Ali and Choi, Yejin},
  booktitle={CVPR},
  year={2019}
}

@incollection{park2019gaugan,
  title={Gaugan: semantic image synthesis with spatially adaptive normalization},
  author={Park, Taesung and Liu, Ming-Yu and Wang, Ting-Chun and Zhu, Jun-Yan},
  booktitle={ACM SIGGRAPH 2019 Real-Time Live!},
  year={2019}
}

@article{lee2019controllable,
  title={Controllable generative adversarial network},
  author={Lee, Minhyeok and Seok, Junhee},
  journal={Ieee Access},
  volume={7},
  pages={28158--28169},
  year={2019},
  publisher={IEEE}
}

@article{brown2020language,
  title={Language models are few-shot learners},
  author={Brown, Tom B},
  journal={arXiv preprint arXiv:2005.14165},
  year={2020}
}

@article{hinz2020semantic,
  title={Semantic object accuracy for generative text-to-image synthesis},
  author={Hinz, Tobias and Heinrich, Stefan and Wermter, Stefan},
  journal={IEEE TPAMI},
  volume={44},
  number={3},
  pages={1552--1565},
  year={2020}
}

@inproceedings{lee2020vilbertscore,
  title={Vilbertscore: Evaluating image caption using vision-and-language bert},
  author={Lee, Hwanhee and Yoon, Seunghyun and Dernoncourt, Franck and Kim, Doo Soon and Bui, Trung and Jung, Kyomin},
  booktitle={Proceedings of the First Workshop on Evaluation and Comparison of NLP Systems},
  pages={34--39},
  year={2020}
}

@inproceedings{zhu2020cookgan,
  title={CookGAN: Causality based text-to-image synthesis},
  author={Zhu, Bin and Ngo, Chong-Wah},
  booktitle={CVPR},
  year={2020}
}

@inproceedings{chen2020generative,
  title={Generative pretraining from pixels},
  author={Chen, Mark and Radford, Alec and Child, Rewon and Wu, Jeffrey and Jun, Heewoo and Luan, David and Sutskever, Ilya},
  booktitle={ICML},
  year={2020}
}

@inproceedings{liang2020cpgan,
  title={Cpgan: Content-parsing generative adversarial networks for text-to-image synthesis},
  author={Liang, Jiadong and Pei, Wenjie and Lu, Feng},
  booktitle={ECCV},
  year={2020}
}

@inproceedings{ho2020denoising,
  title={Denoising diffusion probabilistic models},
  author={Ho, Jonathan and Jain, Ajay and Abbeel, Pieter},
  booktitle={NeurIPS},
  year={2020}
}

@inproceedings{cheng2020rifegan,
  title={RiFeGAN: Rich feature generation for text-to-image synthesis from prior knowledge},
  author={Cheng, Jun and Wu, Fuxiang and Tian, Yanling and Wang, Lei and Tao, Dapeng},
  booktitle={CVPR},
  year={2020}
}

@article{gou2020segattngan,
  title={SegAttnGAN: Text to image generation with segmentation attention},
  author={Gou, Yuchuan and Wu, Qiancheng and Li, Minghao and Gong, Bo and Han, Mei},
  journal={arXiv preprint arXiv:2005.12444},
  year={2020}
}

@article{dodge2020fine,
  title={Fine-tuning pretrained language models: Weight initializations, data orders, and early stopping},
  author={Dodge, Jesse and Ilharco, Gabriel and Schwartz, Roy and Farhadi, Ali and Hajishirzi, Hannaneh and Smith, Noah},
  journal={arXiv preprint arXiv:2002.06305},
  year={2020}
}

@article{dosovitskiy2020image,
  title={An image is worth 16x16 words: Transformers for image recognition at scale},
  author={Dosovitskiy, Alexey},
  journal={arXiv preprint arXiv:2010.11929},
  year={2020}
}

@inproceedings{karras2020analyzing,
  title={Analyzing and improving the image quality of stylegan},
  author={Karras, Tero and Laine, Samuli and Aittala, Miika and Hellsten, Janne and Lehtinen, Jaakko and Aila, Timo},
  booktitle={CVPR},
  year={2020}
}

@inproceedings{sitzmann2020implicit,
  title={Implicit neural representations with periodic activation functions},
  author={Sitzmann, Vincent and Martel, Julien and Bergman, Alexander and Lindell, David and Wetzstein, Gordon},
  booktitle={NeurIPS},
  year={2020}
}

@inproceedings{park2021benchmark,
  title={Benchmark for compositional text-to-image synthesis},
  author={Park, Dong Huk and Azadi, Samaneh and Liu, Xihui and Darrell, Trevor and Rohrbach, Anna},
  booktitle={Thirty-fifth Conference on Neural Information Processing Systems Datasets and Benchmarks Track (Round 1)},
  year={2021}
}

@article{zhang2021ufc,
  title={UFC-BERT: Unifying multi-modal controls for conditional image synthesis},
  author={Zhang, Zhu and Ma, Jianxin and Zhou, Chang and Men, Rui and Li, Zhikang and Ding, Ming and Tang, Jie and Zhou, Jingren and Yang, Hongxia},
  journal={Advances in Neural Information Processing Systems},
  volume={34},
  pages={27196--27208},
  year={2021}
}

@article{hessel2021clipscore,
  title={Clipscore: A reference-free evaluation metric for image captioning},
  author={Hessel, Jack and Holtzman, Ari and Forbes, Maxwell and Bras, Ronan Le and Choi, Yejin},
  journal={arXiv preprint arXiv:2104.08718},
  year={2021}
}

@inproceedings{bai2021training,
  title={On training sample memorization: Lessons from benchmarking generative modeling with a large-scale competition},
  author={Bai, Ching-Yuan and Lin, Hsuan-Tien and Raffel, Colin and Kan, Wendy Chi-wen},
  booktitle={Proceedings of the 27th ACM SIGKDD conference on knowledge discovery \& data mining},
  pages={2534--2542},
  year={2021}
}

@inproceedings{palash2021fine,
  title={Fine-grained image generation from bangla text description using attentional generative adversarial network},
  author={Palash, Md Aminul Haque and Al Nasim, Md Abdullah and Dhali, Aditi and Afrin, Faria},
  booktitle={2021 IEEE International Conference on Robotics, Automation, Artificial-Intelligence and Internet-of-Things (RAAICON)},
  pages={79--84},
  year={2021},
  organization={IEEE}
}

@inproceedings{ramesh2021zero,
  title={Zero-shot text-to-image generation},
  author={Ramesh, Aditya and Pavlov, Mikhail and Goh, Gabriel and Gray, Scott and Voss, Chelsea and Radford, Alec and Chen, Mark and Sutskever, Ilya},
  booktitle={ICML},
  year={2021}
}

@article{zhang2021delving,
  title={Delving deep into label smoothing},
  author={Zhang, Chang-Bin and Jiang, Peng-Tao and Hou, Qibin and Wei, Yunchao and Han, Qi and Li, Zhen and Cheng, Ming-Ming},
  journal={IEEE TIP},
  volume={30},
  pages={5984--5996},
  year={2021}
}

@article{nichol2021glide,
  title={Glide: Towards photorealistic image generation and editing with text-guided diffusion models},
  author={Nichol, Alex and Dhariwal, Prafulla and Ramesh, Aditya and Shyam, Pranav and Mishkin, Pamela and McGrew, Bob and Sutskever, Ilya and Chen, Mark},
  journal={arXiv preprint arXiv:2112.10741},
  year={2021}
}

@inproceedings{wu2021styleformer,
  title={Styleformer: Real-time arbitrary style transfer via parametric style composition},
  author={Wu, Xiaolei and Hu, Zhihao and Sheng, Lu and Xu, Dong},
  booktitle={ICCV},
  year={2021}
}

@inproceedings{radford2021learning,
  title={Learning transferable visual models from natural language supervision},
  author={Radford, Alec and Kim, Jong Wook and Hallacy, Chris and Ramesh, Aditya and Goh, Gabriel and Agarwal, Sandhini and Sastry, Girish and Askell, Amanda and Mishkin, Pamela and Clark, Jack and others},
  booktitle={ICML},
  year={2021}
}

@article{zhou2021survey,
  title={A survey on generative adversarial network-based text-to-image synthesis},
  author={Zhou, Rui and Jiang, Cong and Xu, Qingyang},
  journal={Elsevier Neurocomputing},
  volume={451},
  pages={316--336},
  year={2021}
}

@article{frolov2021adversarial,
  title={Adversarial text-to-image synthesis: A review},
  author={Frolov, Stanislav and Hinz, Tobias and Raue, Federico and Hees, J{\"o}rn and Dengel, Andreas},
  journal={Elsevier Neural Networks},
  volume={144},
  pages={187--209},
  year={2021}
}

@inproceedings{dhariwal2021diffusion,
  title={Diffusion models beat gans on image synthesis},
  author={Dhariwal, Prafulla and Nichol, Alexander},
  booktitle={NeurIPS},
  year={2021}
}

@inproceedings{ding2021cogview,
  title={Cogview: Mastering text-to-image generation via transformers},
  author={Ding, Ming and Yang, Zhuoyi and Hong, Wenyi and Zheng, Wendi and Zhou, Chang and Yin, Da and Lin, Junyang and Zou, Xu and Shao, Zhou and Yang, Hongxia and others},
  booktitle={NeurIPS},
  year={2021}
}

@article{schuhmann2021laion,
  title={Laion-400m: Open dataset of clip-filtered 400 million image-text pairs},
  author={Schuhmann, Christoph and Vencu, Richard and Beaumont, Romain and Kaczmarczyk, Robert and Mullis, Clayton and Katta, Aarush and Coombes, Theo and Jitsev, Jenia and Komatsuzaki, Aran},
  journal={arXiv preprint arXiv:2111.02114},
  year={2021}
}

@inproceedings{esser2021taming,
  title={Taming transformers for high-resolution image synthesis},
  author={Esser, Patrick and Rombach, Robin and Ommer, Bjorn},
  booktitle={CVPR},
  year={2021}
}

@inproceedings{zhang2021cross,
  title={Cross-modal contrastive learning for text-to-image generation},
  author={Zhang, Han and Koh, Jing Yu and Baldridge, Jason and Lee, Honglak and Yang, Yinfei},
  booktitle={CVPR},
  year={2021}
}

@article{hu2021lora,
  title={Lora: Low-rank adaptation of large language models},
  author={Hu, Edward J and Shen, Yelong and Wallis, Phillip and Allen-Zhu, Zeyuan and Li, Yuanzhi and Wang, Shean and Wang, Lu and Chen, Weizhu},
  journal={arXiv preprint arXiv:2106.09685},
  year={2021}
}

@inproceedings{ruan2021dae,
  title={Dae-gan: Dynamic aspect-aware gan for text-to-image synthesis},
  author={Ruan, Shulan and Zhang, Yong and Zhang, Kun and Fan, Yanbo and Tang, Fan and Liu, Qi and Chen, Enhong},
  booktitle={ICCV},
  year={2021}
}

@inproceedings{wang2021cycle,
  title={Cycle-consistent inverse GAN for text-to-image synthesis},
  author={Wang, Hao and Lin, Guosheng and Hoi, Steven CH and Miao, Chunyan},
  booktitle={ACM MM},
  year={2021}
}

@article{meng2021sdedit,
  title={Sdedit: Guided image synthesis and editing with stochastic differential equations},
  author={Meng, Chenlin and He, Yutong and Song, Yang and Song, Jiaming and Wu, Jiajun and Zhu, Jun-Yan and Ermon, Stefano},
  journal={arXiv preprint arXiv:2108.01073},
  year={2021}
}

@article{niu2021review,
  title={A review on the attention mechanism of deep learning},
  author={Niu, Zhaoyang and Zhong, Guoqiang and Yu, Hui},
  journal={Neurocomputing},
  volume={452},
  pages={48--62},
  year={2021},
  publisher={Elsevier}
}

@article{lin2021m6,
  title={M6: A chinese multimodal pretrainer},
  author={Lin, Junyang and Men, Rui and Yang, An and Zhou, Chang and Ding, Ming and Zhang, Yichang and Wang, Peng and Wang, Ang and Jiang, Le and Jia, Xianyan and others},
  journal={arXiv preprint arXiv:2103.00823},
  year={2021}
}

@inproceedings{qiao2021r,
  title={R-GAN: Exploring human-like way for reasonable text-to-image synthesis via generative adversarial networks},
  author={Qiao, Yanyuan and Chen, Qi and Deng, Chaorui and Ding, Ning and Qi, Yuankai and Tan, Mingkui and Ren, Xincheng and Wu, Qi},
  booktitle={ACM MM},
  year={2021}
}

@inproceedings{changpinyo2021conceptual,
  title={Conceptual 12m: Pushing web-scale image-text pre-training to recognize long-tail visual concepts},
  author={Changpinyo, Soravit and Sharma, Piyush and Ding, Nan and Soricut, Radu},
  booktitle={CVPR},
  year={2021}
}

@article{gokhale2022benchmarking,
  title={Benchmarking spatial relationships in text-to-image generation},
  author={Gokhale, Tejas and Palangi, Hamid and Nushi, Besmira and Vineet, Vibhav and Horvitz, Eric and Kamar, Ece and Baral, Chitta and Yang, Yezhou},
  journal={arXiv preprint arXiv:2212.10015},
  year={2022}
}

@article{ramesh2022hierarchical,
  title={Hierarchical text-conditional image generation with clip latents},
  author={Ramesh, Aditya and Dhariwal, Prafulla and Nichol, Alex and Chu, Casey and Chen, Mark},
  journal={arXiv preprint arXiv:2204.06125},
  volume={1},
  number={2},
  pages={3},
  year={2022}
}

@inproceedings{zhang2022perceptual,
  title={Perceptual artifacts localization for inpainting},
  author={Zhang, Lingzhi and Zhou, Yuqian and Barnes, Connelly and Amirghodsi, Sohrab and Lin, Zhe and Shechtman, Eli and Shi, Jianbo},
  booktitle={ECCV},
  year={2022}
}

@article{yuksekgonul2022and,
  title={When and why vision-language models behave like bags-of-words, and what to do about it?},
  author={Yuksekgonul, Mert and Bianchi, Federico and Kalluri, Pratyusha and Jurafsky, Dan and Zou, James},
  journal={arXiv preprint arXiv:2210.01936},
  year={2022}
}

@inproceedings{lezama2022improved,
  title={Improved masked image generation with token-critic},
  author={Lezama, Jos{\'e} and Chang, Huiwen and Jiang, Lu and Essa, Irfan},
  booktitle={ECCV},
  year={2022}
}

@article{wang2022diffusiondb,
  title={Diffusiondb: A large-scale prompt gallery dataset for text-to-image generative models},
  author={Wang, Zijie J and Montoya, Evan and Munechika, David and Yang, Haoyang and Hoover, Benjamin and Chau, Duen Horng},
  journal={arXiv preprint arXiv:2210.14896},
  year={2022}
}

@article{bai2022training,
  title={Training a helpful and harmless assistant with reinforcement learning from human feedback},
  author={Bai, Yuntao and Jones, Andy and Ndousse, Kamal and Askell, Amanda and Chen, Anna and DasSarma, Nova and Drain, Dawn and Fort, Stanislav and Ganguli, Deep and Henighan, Tom and others},
  journal={arXiv preprint arXiv:2204.05862},
  year={2022}
}

@inproceedings{zhou2022towards,
  title={Towards language-free training for text-to-image generation},
  author={Zhou, Yufan and Zhang, Ruiyi and Chen, Changyou and Li, Chunyuan and Tensmeyer, Chris and Yu, Tong and Gu, Jiuxiang and Xu, Jinhui and Sun, Tong},
  year={2022}
}

@inproceedings{wu2022adma,
  title={Adma-GAN: Attribute-Driven Memory Augmented GANs for Text-to-Image Generation.},
  author={Wu, Xintian and Zhao, Hanbin and Zheng, Liangli and Ding, Shouhong and Li, Xi},
  booktitle={ACM MM},
  year={2022}
}

@inproceedings{schuhmann2022laion,
  title={Laion-5b: An open large-scale dataset for training next generation image-text models},
  author={Schuhmann, Christoph and Beaumont, Romain and Vencu, Richard and Gordon, Cade and Wightman, Ross and Cherti, Mehdi and Coombes, Theo and Katta, Aarush and Mullis, Clayton and Wortsman, Mitchell and others},
  booktitle={NeurIPS},
  year={2022}
}

@inproceedings{maharana2022storydall,
  title={Storydall-e: Adapting pretrained text-to-image transformers for story continuation},
  author={Maharana, Adyasha and Hannan, Darryl and Bansal, Mohit},
  booktitle={ECCV},
  year={2022}
}

@inproceedings{liao2022text,
  title={Text to image generation with semantic-spatial aware gan},
  author={Liao, Wentong and Hu, Kai and Yang, Michael Ying and Rosenhahn, Bodo},
  year={2022}
}

@inproceedings{li2022stylet2i,
  title={Stylet2i: Toward compositional and high-fidelity text-to-image synthesis},
  author={Li, Zhiheng and Min, Martin Renqiang and Li, Kai and Xu, Chenliang},
  year={2022}
}

@inproceedings{crowson2022vqgan,
  title={Vqgan-clip: Open domain image generation and editing with natural language guidance},
  author={Crowson, Katherine and Biderman, Stella and Kornis, Daniel and Stander, Dashiell and Hallahan, Eric and Castricato, Louis and Raff, Edward},
  booktitle={ECCV},
  year={2022}
}

@article{rando2022red,
  title={Red-teaming the stable diffusion safety filter},
  author={Rando, Javier and Paleka, Daniel and Lindner, David and Heim, Lennart and Tram{\`e}r, Florian},
  journal={arXiv preprint arXiv:2210.04610},
  year={2022}
}

@article{cheng2022vision,
  title={Vision-language matching for text-to-image synthesis via generative adversarial networks},
  author={Cheng, Qingrong and Wen, Keyu and Gu, Xiaodong},
  journal={IEEE Transactions on Multimedia},
  volume={25},
  pages={7062--7075},
  year={2022},
  publisher={IEEE}
}

@article{jiang2022text2human,
  title={Text2human: Text-driven controllable human image generation},
  author={Jiang, Yuming and Yang, Shuai and Qiu, Haonan and Wu, Wayne and Loy, Chen Change and Liu, Ziwei},
  journal={ACM Transactions on Graphics (TOG)},
  volume={41},
  number={4},
  pages={1--11},
  year={2022},
  publisher={ACM New York, NY, USA}
}

@article{salimans2022progressive,
  title={Progressive distillation for fast sampling of diffusion models},
  author={Salimans, Tim and Ho, Jonathan},
  journal={arXiv preprint arXiv:2202.00512},
  year={2022}
}

@inproceedings{saharia2022photorealistic,
  title={Photorealistic text-to-image diffusion models with deep language understanding},
  author={Saharia, Chitwan and Chan, William and Saxena, Saurabh and Li, Lala and Whang, Jay and Denton, Emily L and Ghasemipour, Kamyar and Gontijo Lopes, Raphael and Karagol Ayan, Burcu and Salimans, Tim and others},
  booktitle={NeurIPS},
  year={2022}
}

@article{ho2022classifier,
  title={Classifier-free diffusion guidance},
  author={Ho, Jonathan and Salimans, Tim},
  journal={arXiv preprint arXiv:2207.12598},
  year={2022}
}

@inproceedings{gu2022vector,
  title={Vector quantized diffusion model for text-to-image synthesis},
  author={Gu, Shuyang and Chen, Dong and Bao, Jianmin and Wen, Fang and Zhang, Bo and Chen, Dongdong and Yuan, Lu and Guo, Baining},
  booktitle={CVPR},
  year={2022}
}

@inproceedings{tao2022df,
  title={Df-gan: A simple and effective baseline for text-to-image synthesis},
  author={Tao, Ming and Tang, Hao and Wu, Fei and Jing, Xiao-Yuan and Bao, Bing-Kun and Xu, Changsheng},
  booktitle={CVPR},
  year={2022}
}

@inproceedings{rombach2022high,
  title={High-resolution image synthesis with latent diffusion models},
  author={Rombach, Robin and Blattmann, Andreas and Lorenz, Dominik and Esser, Patrick and Ommer, Bj{\"o}rn},
  booktitle={CVPR},
  year={2022}
}

@inproceedings{tang2020local,
  title={Local class-specific and global image-level generative adversarial networks for semantic-guided scene generation},
  author={Tang, Hao and Xu, Dan and Yan, Yan and Torr, Philip HS and Sebe, Nicu},
  booktitle={CVPR},
  year={2020}
}

@inproceedings{yan2022trace,
  title={Trace controlled text to image generation},
  author={Yan, Kun and Ji, Lei and Wu, Chenfei and Bao, Jianmin and Zhou, Ming and Duan, Nan and Ma, Shuai},
  booktitle={ECCV},
  year={2022}
}

@inproceedings{tang2019multi,
  title={Multi-channel attention selection gan with cascaded semantic guidance for cross-view image translation},
  author={Tang, Hao and Xu, Dan and Sebe, Nicu and Wang, Yanzhi and Corso, Jason J and Yan, Yan},
  booktitle={CVPR},
  year={2019}
}

@inproceedings{avrahami2022blended,
  title={Blended diffusion for text-driven editing of natural images},
  author={Avrahami, Omri and Lischinski, Dani and Fried, Ohad},
  booktitle={CVPR},
  year={2022}
}

@inproceedings{ding2022cogview2,
  title={Cogview2: Faster and better text-to-image generation via hierarchical transformers},
  author={Ding, Ming and Zheng, Wendi and Hong, Wenyi and Tang, Jie},
  booktitle={NeurIPS},
  year={2022}
}

@article{yu2022scaling,
  title={Scaling Autoregressive Models for Content-Rich Text-to-Image Generation},
  author={Yu, Jiahui and Xu, Yuanzhong and Koh, Jing Yu and Luong, Thang and Baid, Gunjan and Wang, Zirui and Vasudevan, Vijay and Ku, Alexander and Yang, Yinfei and Ayan, Burcu Karagol and others},
  journal={TMLR}
}

@inproceedings{shi2022athom,
  title={AtHom: Two divergent attentions stimulated by homomorphic training in text-to-image synthesis},
  author={Shi, Zhenbo and Chen, Zhi and Xu, Zhenbo and Yang, Wei and Huang, Liusheng},
  booktitle={ACM MM},
  year={2022}
}

@inproceedings{huang2022dse,
  title={Dse-gan: Dynamic semantic evolution generative adversarial network for text-to-image generation},
  author={Huang, Mengqi and Mao, Zhendong and Wang, Penghui and Wang, Quan and Zhang, Yongdong},
  booktitle={ACM MM},
  year={2022}
}

@inproceedings{chen2022background,
  title={Background layout generation and object knowledge transfer for text-to-image generation},
  author={Chen, Zhuowei and Mao, Zhendong and Fang, Shancheng and Hu, Bo},
  booktitle={ACM MM},
  year={2022}
}

@inproceedings{gafni2022make,
  title={Make-a-scene: Scene-based text-to-image generation with human priors},
  author={Gafni, Oran and Polyak, Adam and Ashual, Oron and Sheynin, Shelly and Parikh, Devi and Taigman, Yaniv},
  booktitle={ECCV},
  year={2022}
}

@inproceedings{wu2022nuwa,
  title={N{\"u}wa: Visual synthesis pre-training for neural visual world creation},
  author={Wu, Chenfei and Liang, Jian and Ji, Lei and Yang, Fan and Fang, Yuejian and Jiang, Daxin and Duan, Nan},
  booktitle={ECCV},
  year={2022}
}

@inproceedings{wei2022chain,
  title={Chain-of-thought prompting elicits reasoning in large language models},
  author={Wei, Jason and Wang, Xuezhi and Schuurmans, Dale and Bosma, Maarten and Xia, Fei and Chi, Ed and Le, Quoc V and Zhou, Denny and others},
  booktitle={NeurIPS},
  year={2022}
}

@article{feng2022training,
  title={Training-free structured diffusion guidance for compositional text-to-image synthesis},
  author={Feng, Weixi and He, Xuehai and Fu, Tsu-Jui and Jampani, Varun and Akula, Arjun and Narayana, Pradyumna and Basu, Sugato and Wang, Xin Eric and Wang, William Yang},
  journal={arXiv preprint arXiv:2212.05032},
  year={2022}
}

@inproceedings{kim2022diffusionclip,
  title={Diffusionclip: Text-guided diffusion models for robust image manipulation},
  author={Kim, Gwanghyun and Kwon, Taesung and Ye, Jong Chul},
  booktitle={CVPR},
  year={2022}
}

@article{gal2022image,
  title={An image is worth one word: Personalizing text-to-image generation using textual inversion},
  author={Gal, Rinon and Alaluf, Yuval and Atzmon, Yuval and Patashnik, Or and Bermano, Amit H and Chechik, Gal and Cohen-Or, Daniel},
  journal={arXiv preprint arXiv:2208.01618},
  year={2022}
}

@inproceedings{thrush2022winoground,
  title={Winoground: Probing vision and language models for visio-linguistic compositionality},
  author={Thrush, Tristan and Jiang, Ryan and Bartolo, Max and Singh, Amanpreet and Williams, Adina and Kiela, Douwe and Ross, Candace},
  booktitle={CVPR},
  year={2022}
}

@inproceedings{dinh2022tise,
  title={Tise: Bag of metrics for text-to-image synthesis evaluation},
  author={Dinh, Tan M and Nguyen, Rang and Hua, Binh-Son},
  booktitle={ECCV},
  year={2022}
}

@inproceedings{wu2022text,
  title={Text-to-image synthesis based on object-guided joint-decoding transformer},
  author={Wu, Fuxiang and Liu, Liu and Hao, Fusheng and He, Fengxiang and Cheng, Jun},
  booktitle={CVPR},
  year={2022}
}

@inproceedings{li2022grounded,
  title={Grounded language-image pre-training},
  author={Li, Liunian Harold and Zhang, Pengchuan and Zhang, Haotian and Yang, Jianwei and Li, Chunyuan and Zhong, Yiwu and Wang, Lijuan and Yuan, Lu and Zhang, Lei and Hwang, Jenq-Neng and others},
  booktitle={CVPR},
  year={2022}
}

@inproceedings{li2022blip,
  title={Blip: Bootstrapping language-image pre-training for unified vision-language understanding and generation},
  author={Li, Junnan and Li, Dongxu and Xiong, Caiming and Hoi, Steven},
  booktitle={ICML},
  year={2022}
}

@inproceedings{kim2022mutual,
  title={Mutual information divergence: A unified metric for multimodal generative models},
  author={Kim, Jin-Hwa and Kim, Yunji and Lee, Jiyoung and Yoo, Kang Min and Lee, Sang-Woo},
  year={2022}
}

@article{ku2023viescore,
  title={Viescore: Towards explainable metrics for conditional image synthesis evaluation},
  author={Ku, Max and Jiang, Dongfu and Wei, Cong and Yue, Xiang and Chen, Wenhu},
  journal={arXiv preprint arXiv:2312.14867},
  year={2023}
}

@inproceedings{singh2023divide,
  title={Divide, evaluate, and refine: Evaluating and improving text-to-image alignment with iterative vqa feedback},
  author={Singh, Jaskirat and Zheng, Liang},
  booktitle={NeurIPS},
  year={2023}
}

@article{chang2023muse,
  title={Muse: Text-to-image generation via masked generative transformers},
  author={Chang, Huiwen and Zhang, Han and Barber, Jarred and Maschinot, AJ and Lezama, Jose and Jiang, Lu and Yang, Ming-Hsuan and Murphy, Kevin and Freeman, William T and Rubinstein, Michael and others},
  journal={arXiv preprint arXiv:2301.00704},
  year={2023}
}

@inproceedings{betti2023let,
  title={Let's ViCE! Mimicking Human Cognitive Behavior in Image Generation Evaluation},
  author={Betti, Federico and Staiano, Jacopo and Baraldi, Lorenzo and Baraldi, Lorenzo and Cucchiara, Rita and Sebe, Nicu},
  booktitle={ACM MM},
  year={2023}
}

@inproceedings{hu2023tifa,
  title={Tifa: Accurate and interpretable text-to-image faithfulness evaluation with question answering},
  author={Hu, Yushi and Liu, Benlin and Kasai, Jungo and Wang, Yizhong and Ostendorf, Mari and Krishna, Ranjay and Smith, Noah A},
  booktitle={ICCV},
  year={2023}
}

@inproceedings{wang2023exploring,
  title={Exploring clip for assessing the look and feel of images},
  author={Wang, Jianyi and Chan, Kelvin CK and Loy, Chen Change},
  booktitle={AAAI},
  year={2023}
}

@inproceedings{zhang2023perceptual,
  title={Perceptual artifacts localization for image synthesis tasks},
  author={Zhang, Lingzhi and Xu, Zhengjie and Barnes, Connelly and Zhou, Yuqian and Liu, Qing and Zhang, He and Amirghodsi, Sohrab and Lin, Zhe and Shechtman, Eli and Shi, Jianbo},
  booktitle={ICCV},
  year={2023}
}

@inproceedings{zhou2023vision+,
  title={Vision+ language applications: A survey},
  author={Zhou, Yutong and Shimada, Nobutaka},
  booktitle={CVPRW},
  year={2023}
}

@article{singh2023coarse,
  title={Coarse-to-fine contrastive learning in image-text-graph space for improved vision-language compositionality},
  author={Singh, Harman and Zhang, Pengchuan and Wang, Qifan and Wang, Mengjiao and Xiong, Wenhan and Du, Jingfei and Chen, Yu},
  journal={arXiv preprint arXiv:2305.13812},
  year={2023}
}

@inproceedings{kirstain2023pick,
  title={Pick-a-pic: An open dataset of user preferences for text-to-image generation},
  author={Kirstain, Yuval and Polyak, Adam and Singer, Uriel and Matiana, Shahbuland and Penna, Joe and Levy, Omer},
  booktitle={NeurIPS},
  year={2023}
}

@article{wu2023paragraph,
  title={Paragraph-to-image generation with information-enriched diffusion model},
  author={Wu, Weijia and Li, Zhuang and He, Yefei and Shou, Mike Zheng and Shen, Chunhua and Cheng, Lele and Li, Yan and Gao, Tingting and Zhang, Di and Wang, Zhongyuan},
  journal={arXiv preprint arXiv:2311.14284},
  year={2023}
}

@inproceedings{xie2023difffit,
  title={Difffit: Unlocking transferability of large diffusion models via simple parameter-efficient fine-tuning},
  author={Xie, Enze and Yao, Lewei and Shi, Han and Liu, Zhili and Zhou, Daquan and Liu, Zhaoqiang and Li, Jiawei and Li, Zhenguo},
  booktitle={ICCV},
  year={2023}
}

@article{kodaira2023streamdiffusion,
  title={Streamdiffusion: A pipeline-level solution for real-time interactive generation},
  author={Kodaira, Akio and Xu, Chenfeng and Hazama, Toshiki and Yoshimoto, Takanori and Ohno, Kohei and Mitsuhori, Shogo and Sugano, Soichi and Cho, Hanying and Liu, Zhijian and Keutzer, Kurt},
  journal={arXiv preprint arXiv:2312.12491},
  year={2023}
}

@inproceedings{kang2023scaling,
  title={Scaling up gans for text-to-image synthesis},
  author={Kang, Minguk and Zhu, Jun-Yan and Zhang, Richard and Park, Jaesik and Shechtman, Eli and Paris, Sylvain and Park, Taesung},
  booktitle={CVPR},
  year={2023}
}

@inproceedings{liu2023riatig,
  title={Riatig: Reliable and imperceptible adversarial text-to-image generation with natural prompts},
  author={Liu, Han and Wu, Yuhao and Zhai, Shixuan and Yuan, Bo and Zhang, Ning},
  booktitle={CVPR},
  year={2023}
}

@inproceedings{pan2023drag,
  title={Drag your gan: Interactive point-based manipulation on the generative image manifold},
  author={Pan, Xingang and Tewari, Ayush and Leimk{\"u}hler, Thomas and Liu, Lingjie and Meka, Abhimitra and Theobalt, Christian},
  booktitle={ACM SIGGRAPH 2023 Conference Proceedings},
  pages={1--11},
  year={2023}
}

@inproceedings{mokady2023null,
  title={Null-text inversion for editing real images using guided diffusion models},
  author={Mokady, Ron and Hertz, Amir and Aberman, Kfir and Pritch, Yael and Cohen-Or, Daniel},
  booktitle={CVPR},
  year={2023}
}

@inproceedings{kawar2023imagic,
  title={Imagic: Text-based real image editing with diffusion models},
  author={Kawar, Bahjat and Zada, Shiran and Lang, Oran and Tov, Omer and Chang, Huiwen and Dekel, Tali and Mosseri, Inbar and Irani, Michal},
  booktitle={CVPR},
  year={2023}
}

@inproceedings{zhang2023inversion,
  title={Inversion-based style transfer with diffusion models},
  author={Zhang, Yuxin and Huang, Nisha and Tang, Fan and Huang, Haibin and Ma, Chongyang and Dong, Weiming and Xu, Changsheng},
  booktitle={CVPR},
  year={2023}
}

@article{pang2023cross,
  title={Cross initialization for personalized text-to-image generation},
  author={Pang, Lianyu and Yin, Jian and Xie, Haoran and Wang, Qiping and Li, Qing and Mao, Xudong},
  journal={arXiv preprint arXiv:2312.15905},
  year={2023}
}

@inproceedings{wu2023uncovering,
  title={Uncovering the disentanglement capability in text-to-image diffusion models},
  author={Wu, Qiucheng and Liu, Yujian and Zhao, Handong and Kale, Ajinkya and Bui, Trung and Yu, Tong and Lin, Zhe and Zhang, Yang and Chang, Shiyu},
  booktitle={CVPR},
  year={2023}
}

@inproceedings{lin2023zero,
  title={Zero-shot everything sketch-based image retrieval, and in explainable style},
  author={Lin, Fengyin and Li, Mingkang and Li, Da and Hospedales, Timothy and Song, Yi-Zhe and Qi, Yonggang},
  booktitle={CVPR},
  year={2023}
}

@article{yu2023scaling,
  title={Scaling autoregressive multi-modal models: Pretraining and instruction tuning},
  author={Yu, Lili and Shi, Bowen and Pasunuru, Ramakanth and Muller, Benjamin and Golovneva, Olga and Wang, Tianlu and Babu, Arun and Tang, Binh and Karrer, Brian and Sheynin, Shelly and others},
  journal={arXiv preprint arXiv:2309.02591},
  volume={2},
  number={3},
  year={2023}
}

@inproceedings{zhang2023sine,
  title={Sine: Single image editing with text-to-image diffusion models},
  author={Zhang, Zhixing and Han, Ligong and Ghosh, Arnab and Metaxas, Dimitris N and Ren, Jian},
  booktitle={CVPR},
  year={2023}
}

@article{wu2023ai,
  title={Ai-generated content (aigc): A survey},
  author={Wu, Jiayang and Gan, Wensheng and Chen, Zefeng and Wan, Shicheng and Lin, Hong},
  journal={arXiv preprint arXiv:2304.06632},
  year={2023}
}

@article{chen2023pixart,
  title={Pixart-\textbackslash$\alpha$: Fast training of diffusion transformer for photorealistic text-to-image synthesis},
  author={Chen, Junsong and Yu, Jincheng and Ge, Chongjian and Yao, Lewei and Xie, Enze and Wu, Yue and Wang, Zhongdao and Kwok, James and Luo, Ping and Lu, Huchuan and others},
  journal={arXiv preprint arXiv:2310.00426},
  year={2023}
}

@article{betker2023improving,
  title={Improving image generation with better captions},
  author={Betker, James and Goh, Gabriel and Jing, Li and Brooks, Tim and Wang, Jianfeng and Li, Linjie and Ouyang, Long and Zhuang, Juntang and Lee, Joyce and Guo, Yufei and others},
  journal={Computer Science. https://cdn. openai. com/papers/dall-e-3. pdf},
  volume={2},
  number={3},
  pages={8},
  year={2023}
}

@inproceedings{lu2023specialist,
  title={Specialist diffusion: Plug-and-play sample-efficient fine-tuning of text-to-image diffusion models to learn any unseen style},
  author={Lu, Haoming and Tunanyan, Hazarapet and Wang, Kai and Navasardyan, Shant and Wang, Zhangyang and Shi, Humphrey},
  booktitle={CVPR},
  year={2023}
}

@article{tuo2023anytext,
  title={Anytext: Multilingual visual text generation and editing},
  author={Tuo, Yuxiang and Xiang, Wangmeng and He, Jun-Yan and Geng, Yifeng and Xie, Xuansong},
  journal={arXiv preprint arXiv:2311.03054},
  year={2023}
}

@inproceedings{farshad2023scenegenie,
  title={Scenegenie: Scene graph guided diffusion models for image synthesis},
  author={Farshad, Azade and Yeganeh, Yousef and Chi, Yu and Shen, Chengzhi and Ommer, B{\"o}jrn and Navab, Nassir},
  booktitle={ICCV},
  year={2023}
}

@inproceedings{couairon2023zero,
  title={Zero-shot spatial layout conditioning for text-to-image diffusion models},
  author={Couairon, Guillaume and Careil, Marlene and Cord, Matthieu and Lathuiliere, St{\'e}phane and Verbeek, Jakob},
  booktitle={ICCV},
  year={2023}
}

@inproceedings{bao2023one,
  title={One transformer fits all distributions in multi-modal diffusion at scale},
  author={Bao, Fan and Nie, Shen and Xue, Kaiwen and Li, Chongxuan and Pu, Shi and Wang, Yaole and Yue, Gang and Cao, Yue and Su, Hang and Zhu, Jun},
  booktitle={ICML},
  year={2023}
}

@article{podell2023sdxl,
  title={Sdxl: Improving latent diffusion models for high-resolution image synthesis},
  author={Podell, Dustin and English, Zion and Lacey, Kyle and Blattmann, Andreas and Dockhorn, Tim and M{\"u}ller, Jonas and Penna, Joe and Rombach, Robin},
  journal={arXiv preprint arXiv:2307.01952},
  year={2023}
}

@article{dai2023emu,
  title={Emu: Enhancing image generation models using photogenic needles in a haystack},
  author={Dai, Xiaoliang and Hou, Ji and Ma, Chih-Yao and Tsai, Sam and Wang, Jialiang and Wang, Rui and Zhang, Peizhao and Vandenhende, Simon and Wang, Xiaofang and Dubey, Abhimanyu and others},
  journal={arXiv preprint arXiv:2309.15807},
  year={2023}
}

@article{teng2023relay,
  title={Relay diffusion: Unifying diffusion process across resolutions for image synthesis},
  author={Teng, Jiayan and Zheng, Wendi and Ding, Ming and Hong, Wenyi and Wangni, Jianqiao and Yang, Zhuoyi and Tang, Jie},
  journal={arXiv preprint arXiv:2309.03350},
  year={2023}
}

@inproceedings{tao2023galip,
  title={Galip: Generative adversarial clips for text-to-image synthesis},
  author={Tao, Ming and Bao, Bing-Kun and Tang, Hao and Xu, Changsheng},
  booktitle={CVPR},
  year={2023}
}

@inproceedings{liu2023instaflow,
  title={Instaflow: One step is enough for high-quality diffusion-based text-to-image generation},
  author={Liu, Xingchao and Zhang, Xiwen and Ma, Jianzhu and Peng, Jian and others},
  booktitle={The Twelfth International Conference on Learning Representations},
  year={2023}
}

@article{achiam2023gpt,
  title={Gpt-4 technical report},
  author={Achiam, Josh and Adler, Steven and Agarwal, Sandhini and Ahmad, Lama and Akkaya, Ilge and Aleman, Florencia Leoni and Almeida, Diogo and Altenschmidt, Janko and Altman, Sam and Anadkat, Shyamal and others},
  journal={arXiv preprint arXiv:2303.08774},
  year={2023}
}

@inproceedings{huang2023t2i,
  title={T2i-compbench: A comprehensive benchmark for open-world compositional text-to-image generation},
  author={Huang, Kaiyi and Sun, Kaiyue and Xie, Enze and Li, Zhenguo and Liu, Xihui},
  booktitle={NeurIPS},
  year={2023}
}

@inproceedings{li2023blip,
  title={Blip-2: Bootstrapping language-image pre-training with frozen image encoders and large language models},
  author={Li, Junnan and Li, Dongxu and Savarese, Silvio and Hoi, Steven},
  booktitle={ICML},
  year={2023}
}

@article{ye2023recurrent,
  title={Recurrent affine transformation for text-to-image synthesis},
  author={Ye, Senmao and Wang, Huan and Tan, Mingkui and Liu, Fei},
  journal={IEEE TMM},
  volume={26},
  pages={462--473},
  year={2023}
}

@inproceedings{hertz2023delta,
  title={Delta denoising score},
  author={Hertz, Amir and Aberman, Kfir and Cohen-Or, Daniel},
  booktitle={ICCV},
  year={2023}
}

@inproceedings{schramowski2023safe,
  title={Safe latent diffusion: Mitigating inappropriate degeneration in diffusion models},
  author={Schramowski, Patrick and Brack, Manuel and Deiseroth, Bj{\"o}rn and Kersting, Kristian},
  booktitle={CVPR},
  year={2023}
}

@inproceedings{gandikota2023erasing,
  title={Erasing concepts from diffusion models},
  author={Gandikota, Rohit and Materzynska, Joanna and Fiotto-Kaufman, Jaden and Bau, David},
  booktitle={ICCV},
  year={2023}
}

@article{zhang2023multimodal,
  title={Multimodal chain-of-thought reasoning in language models},
  author={Zhang, Zhuosheng and Zhang, Aston and Li, Mu and Zhao, Hai and Karypis, George and Smola, Alex},
  journal={arXiv preprint arXiv:2302.00923},
  year={2023}
}

@article{wang2023tokencompose,
  title={Tokencompose: Grounding diffusion with token-level supervision},
  author={Wang, Zirui and Sha, Zhizhou and Ding, Zheng and Wang, Yilin and Tu, Zhuowen},
  journal={arXiv preprint arXiv:2312.03626},
  year={2023}
}

@inproceedings{patashnik2023localizing,
  title={Localizing object-level shape variations with text-to-image diffusion models},
  author={Patashnik, Or and Garibi, Daniel and Azuri, Idan and Averbuch-Elor, Hadar and Cohen-Or, Daniel},
  booktitle={ICCV},
  year={2023}
}

@inproceedings{xu2023versatile,
  title={Versatile diffusion: Text, images and variations all in one diffusion model},
  author={Xu, Xingqian and Wang, Zhangyang and Zhang, Gong and Wang, Kai and Shi, Humphrey},
  booktitle={ICCV},
  year={2023}
}

@inproceedings{kumari2023multi,
  title={Multi-concept customization of text-to-image diffusion},
  author={Kumari, Nupur and Zhang, Bingliang and Zhang, Richard and Shechtman, Eli and Zhu, Jun-Yan},
  booktitle={CVPR},
  year={2023}
}

@inproceedings{rahman2023make,
  title={Make-a-story: Visual memory conditioned consistent story generation},
  author={Rahman, Tanzila and Lee, Hsin-Ying and Ren, Jian and Tulyakov, Sergey and Mahajan, Shweta and Sigal, Leonid},
  booktitle={CVPR},
  year={2023}
}

@inproceedings{yang2023paint,
  title={Paint by example: Exemplar-based image editing with diffusion models},
  author={Yang, Binxin and Gu, Shuyang and Zhang, Bo and Zhang, Ting and Chen, Xuejin and Sun, Xiaoyan and Chen, Dong and Wen, Fang},
  booktitle={CVPR},
  year={2023}
}

@article{ye2023ip,
  title={Ip-adapter: Text compatible image prompt adapter for text-to-image diffusion models},
  author={Ye, Hu and Zhang, Jun and Liu, Sibo and Han, Xiao and Yang, Wei},
  journal={arXiv preprint arXiv:2308.06721},
  year={2023}
}

@inproceedings{han2023svdiff,
  title={Svdiff: Compact parameter space for diffusion fine-tuning},
  author={Han, Ligong and Li, Yinxiao and Zhang, Han and Milanfar, Peyman and Metaxas, Dimitris and Yang, Feng},
  booktitle={ICCV},
  year={2023}
}

@inproceedings{sauer2023stylegan,
  title={Stylegan-t: Unlocking the power of gans for fast large-scale text-to-image synthesis},
  author={Sauer, Axel and Karras, Tero and Laine, Samuli and Geiger, Andreas and Aila, Timo},
  booktitle={ICML},
  year={2023}
}

@inproceedings{feng2023ernie,
  title={Ernie-vilg 2.0: Improving text-to-image diffusion model with knowledge-enhanced mixture-of-denoising-experts},
  author={Feng, Zhida and Zhang, Zhenyu and Yu, Xintong and Fang, Yewei and Li, Lanxin and Chen, Xuyi and Lu, Yuxiang and Liu, Jiaxiang and Yin, Weichong and Feng, Shikun and others},
  booktitle={CVPR},
  year={2023}
}

@inproceedings{zhou2023shifted,
  title={Shifted diffusion for text-to-image generation},
  author={Zhou, Yufan and Liu, Bingchen and Zhu, Yizhe and Yang, Xiao and Chen, Changyou and Xu, Jinhui},
  booktitle={CVPR},
  year={2023}
}

@inproceedings{wang2023context,
  title={In-context learning unlocked for diffusion models},
  author={Wang, Zhendong and Jiang, Yifan and Lu, Yadong and He, Pengcheng and Chen, Weizhu and Wang, Zhangyang and Zhou, Mingyuan and others},
  booktitle={NeurIPS},
  year={2023}
}

@inproceedings{liu2023more,
  title={More control for free! image synthesis with semantic diffusion guidance},
  author={Liu, Xihui and Park, Dong Huk and Azadi, Samaneh and Zhang, Gong and Chopikyan, Arman and Hu, Yuxiao and Shi, Humphrey and Rohrbach, Anna and Darrell, Trevor},
  booktitle={WACV},
  year={2023}
}

@article{xu2023inversion,
  title={Inversion-free image editing with natural language},
  author={Xu, Sihan and Huang, Yidong and Pan, Jiayi and Ma, Ziqiao and Chai, Joyce},
  journal={arXiv preprint arXiv:2312.04965},
  year={2023}
}

@inproceedings{cho2023dall,
  title={Dall-eval: Probing the reasoning skills and social biases of text-to-image generation models},
  author={Cho, Jaemin and Zala, Abhay and Bansal, Mohit},
  booktitle={ICCV},
  year={2023}
}

@inproceedings{zhong2023adapter,
  title={Sur-adapter: Enhancing text-to-image pre-trained diffusion models with large language models},
  author={Zhong, Shanshan and Huang, Zhongzhan and Wen, Weushao and Qin, Jinghui and Lin, Liang},
  booktitle={ACM MM},
  year={2023}
}

@article{fu2023dreamsim,
  title={Dreamsim: Learning new dimensions of human visual similarity using synthetic data},
  author={Fu, Stephanie and Tamir, Netanel and Sundaram, Shobhita and Chai, Lucy and Zhang, Richard and Dekel, Tali and Isola, Phillip},
  journal={arXiv preprint arXiv:2306.09344},
  year={2023}
}

@article{liang2023mist,
  title={Mist: Towards improved adversarial examples for diffusion models},
  author={Liang, Chumeng and Wu, Xiaoyu},
  journal={arXiv preprint arXiv:2305.12683},
  year={2023}
}

@article{hua2023dreamtuner,
  title={Dreamtuner: Single image is enough for subject-driven generation},
  author={Hua, Miao and Liu, Jiawei and Ding, Fei and Liu, Wei and Wu, Jie and He, Qian},
  journal={arXiv preprint arXiv:2312.13691},
  year={2023}
}

@article{fu2023guiding,
  title={Guiding instruction-based image editing via multimodal large language models},
  author={Fu, Tsu-Jui and Hu, Wenze and Du, Xianzhi and Wang, William Yang and Yang, Yinfei and Gan, Zhe},
  journal={arXiv preprint arXiv:2309.17102},
  year={2023}
}

@inproceedings{zheng2023layoutdiffusion,
  title={Layoutdiffusion: Controllable diffusion model for layout-to-image generation},
  author={Zheng, Guangcong and Zhou, Xianpan and Li, Xuewei and Qi, Zhongang and Shan, Ying and Li, Xi},
  booktitle={CVPR},
  year={2023}
}

@inproceedings{wei2023elite,
  title={Elite: Encoding visual concepts into textual embeddings for customized text-to-image generation},
  author={Wei, Yuxiang and Zhang, Yabo and Ji, Zhilong and Bai, Jinfeng and Zhang, Lei and Zuo, Wangmeng},
  booktitle={ICCV},
  year={2023}
}

@incollection{li2023layerdiffusion,
  title={Layerdiffusion: Layered controlled image editing with diffusion models},
  author={Li, Pengzhi and Huang, Qinxuan and Ding, Yikang and Li, Zhiheng},
  booktitle={SIGGRAPH Asia 2023 Technical Communications},
  pages={1--4},
  year={2023}
}

@article{gong2023talecrafter,
  title={Talecrafter: Interactive story visualization with multiple characters},
  author={Gong, Yuan and Pang, Youxin and Cun, Xiaodong and Xia, Menghan and He, Yingqing and Chen, Haoxin and Wang, Longyue and Zhang, Yong and Wang, Xintao and Shan, Ying and others},
  journal={arXiv preprint arXiv:2305.18247},
  year={2023}
}

@article{bar2023multidiffusion,
  title={Multidiffusion: Fusing diffusion paths for controlled image generation},
  author={Bar-Tal, Omer and Yariv, Lior and Lipman, Yaron and Dekel, Tali},
  year={2023}
}

@article{skandarani2023gans,
  title={Gans for medical image synthesis: An empirical study},
  author={Skandarani, Youssef and Jodoin, Pierre-Marc and Lalande, Alain},
  journal={Journal of Imaging},
  volume={9},
  number={3},
  pages={69},
  year={2023},
  publisher={MDPI}
}

@article{pernias2023wurstchen,
  title={W{\"u}rstchen: An efficient architecture for large-scale text-to-image diffusion models},
  author={Pernias, Pablo and Rampas, Dominic and Richter, Mats L and Pal, Christopher J and Aubreville, Marc},
  journal={arXiv preprint arXiv:2306.00637},
  year={2023}
}

@article{voynov2023p+,
  title={p+: Extended textual conditioning in text-to-image generation},
  author={Voynov, Andrey and Chu, Qinghao and Cohen-Or, Daniel and Aberman, Kfir},
  journal={arXiv preprint arXiv:2303.09522},
  year={2023}
}

@article{zhang2023text,
  title={Text-to-image diffusion models in generative ai: A survey},
  author={Zhang, Chenshuang and Zhang, Chaoning and Zhang, Mengchun and Kweon, In So},
  journal={arXiv preprint arXiv:2303.07909},
  year={2023}
}

@inproceedings{yang2023reco,
  title={Reco: Region-controlled text-to-image generation},
  author={Yang, Zhengyuan and Wang, Jianfeng and Gan, Zhe and Li, Linjie and Lin, Kevin and Wu, Chenfei and Duan, Nan and Liu, Zicheng and Liu, Ce and Zeng, Michael and others},
  booktitle={CVPR},
  year={2023}
}

@article{yin2023survey,
  title={A survey on multimodal large language models},
  author={Yin, Shukang and Fu, Chaoyou and Zhao, Sirui and Li, Ke and Sun, Xing and Xu, Tong and Chen, Enhong},
  journal={arXiv preprint arXiv:2306.13549},
  year={2023}
}

@inproceedings{avrahami2023spatext,
  title={Spatext: Spatio-textual representation for controllable image generation},
  author={Avrahami, Omri and Hayes, Thomas and Gafni, Oran and Gupta, Sonal and Taigman, Yaniv and Parikh, Devi and Lischinski, Dani and Fried, Ohad and Yin, Xi},
  booktitle={CVPR},
  year={2023}
}

@article{vice2023quantifying,
  title={Quantifying bias in text-to-image generative models},
  author={Vice, Jordan and Akhtar, Naveed and Hartley, Richard and Mian, Ajmal},
  journal={arXiv preprint arXiv:2312.13053},
  year={2023}
}

@inproceedings{li2023gligen,
  title={Gligen: Open-set grounded text-to-image generation},
  author={Li, Yuheng and Liu, Haotian and Wu, Qingyang and Mu, Fangzhou and Yang, Jianwei and Gao, Jianfeng and Li, Chunyuan and Lee, Yong Jae},
  booktitle={CVPR},
  year={2023}
}

@inproceedings{chen2023controlstyle,
  title={Controlstyle: Text-driven stylized image generation using diffusion priors},
  author={Chen, Jingwen and Pan, Yingwei and Yao, Ting and Mei, Tao},
  booktitle={ACM MM},
  year={2023}
}

@article{chefer2023attend,
  title={Attend-and-excite: Attention-based semantic guidance for text-to-image diffusion models},
  author={Chefer, Hila and Alaluf, Yuval and Vinker, Yael and Wolf, Lior and Cohen-Or, Daniel},
  journal={ACM TOG},
  volume={42},
  number={4},
  pages={1--10},
  year={2023}
}

@inproceedings{qu2023layoutllm,
  title={Layoutllm-t2i: Eliciting layout guidance from llm for text-to-image generation},
  author={Qu, Leigang and Wu, Shengqiong and Fei, Hao and Nie, Liqiang and Chua, Tat-Seng},
  booktitle={ACM MM},
  year={2023}
}

@article{zala2023diagrammergpt,
  title={DiagrammerGPT: Generating Open-Domain, Open-Platform Diagrams via LLM Planning},
  author={Zala, Abhay and Lin, Han and Cho, Jaemin and Bansal, Mohit},
  journal={arXiv preprint arXiv:2310.12128},
  year={2023}
}

@inproceedings{ruiz2023dreambooth,
  title={Dreambooth: Fine tuning text-to-image diffusion models for subject-driven generation},
  author={Ruiz, Nataniel and Li, Yuanzhen and Jampani, Varun and Pritch, Yael and Rubinstein, Michael and Aberman, Kfir},
  booktitle={CVPR},
  year={2023}
}

@article{hao2023vico,
  title={ViCo: Plug-and-play Visual Condition for Personalized Text-to-image Generation},
  author={Hao, Shaozhe and Han, Kai and Zhao, Shihao and Wong, Kwan-Yee K},
  journal={arXiv preprint arXiv:2306.00971},
  year={2023}
}

@inproceedings{cao2023masactrl,
  title={Masactrl: Tuning-free mutual self-attention control for consistent image synthesis and editing},
  author={Cao, Mingdeng and Wang, Xintao and Qi, Zhongang and Shan, Ying and Qie, Xiaohu and Zheng, Yinqiang},
  booktitle={ICCV},
  year={2023}
}

@inproceedings{zhang2023adding,
  title={Adding conditional control to text-to-image diffusion models},
  author={Zhang, Lvmin and Rao, Anyi and Agrawala, Maneesh},
  booktitle={ICCV},
  year={2023}
}

@article{shen2023advancing,
  title={Advancing pose-guided image synthesis with progressive conditional diffusion models},
  author={Shen, Fei and Ye, Hu and Zhang, Jun and Wang, Cong and Han, Xiao and Yang, Wei},
  journal={arXiv preprint arXiv:2310.06313},
  year={2023}
}

@article{wang2023autostory,
  title={Autostory: Generating diverse storytelling images with minimal human effort},
  author={Wang, Wen and Zhao, Canyu and Chen, Hao and Chen, Zhekai and Zheng, Kecheng and Shen, Chunhua},
  journal={arXiv preprint arXiv:2311.11243},
  year={2023}
}

@article{lian2023llm,
  title={Llm-grounded diffusion: Enhancing prompt understanding of text-to-image diffusion models with large language models},
  author={Lian, Long and Li, Boyi and Yala, Adam and Darrell, Trevor},
  journal={arXiv preprint arXiv:2305.13655},
  year={2023}
}

@inproceedings{van2023anti,
  title={Anti-dreambooth: Protecting users from personalized text-to-image synthesis},
  author={Van Le, Thanh and Phung, Hao and Nguyen, Thuan Hoang and Dao, Quan and Tran, Ngoc N and Tran, Anh},
  booktitle={ICCV},
  year={2023}
}

@article{liang2023adversarial,
  title={Adversarial example does good: Preventing painting imitation from diffusion models via adversarial examples},
  author={Liang, Chumeng and Wu, Xiaoyu and Hua, Yang and Zhang, Jiaru and Xue, Yiming and Song, Tao and Xue, Zhengui and Ma, Ruhui and Guan, Haibing},
  journal={arXiv preprint arXiv:2302.04578},
  year={2023}
}

@inproceedings{wu2023human,
  title={Human preference score: Better aligning text-to-image models with human preference},
  author={Wu, Xiaoshi and Sun, Keqiang and Zhu, Feng and Zhao, Rui and Li, Hongsheng},
  booktitle={ICCV},
  year={2023}
}

@article{wu2023human2,
  title={Human preference score v2: A solid benchmark for evaluating human preferences of text-to-image synthesis},
  author={Wu, Xiaoshi and Hao, Yiming and Sun, Keqiang and Chen, Yixiong and Zhu, Feng and Zhao, Rui and Li, Hongsheng},
  journal={arXiv preprint arXiv:2306.09341},
  year={2023}
}

@article{ma2024cobra,
  title={Cobra Effect in Reference-Free Image Captioning Metrics},
  author={Ma, Zheng and Wang, Changxin and Ouyang, Yawen and Zhao, Fei and Zhang, Jianbing and Huang, Shujian and Chen, Jiajun},
  journal={arXiv preprint arXiv:2402.11572},
  year={2024}
}

@inproceedings{chen2024tailored,
  title={Tailored visions: Enhancing text-to-image generation with personalized prompt rewriting},
  author={Chen, Zijie and Zhang, Lichao and Weng, Fangsheng and Pan, Lili and Lan, Zhenzhong},
  booktitle={CVPR},
  year={2024}
}

@inproceedings{hoe2024interactdiffusion,
  title={InteractDiffusion: Interaction Control in Text-to-Image Diffusion Models},
  author={Hoe, Jiun Tian and Jiang, Xudong and Chan, Chee Seng and Tan, Yap-Peng and Hu, Weipeng},
  booktitle={CVPR},
  year={2024}
}

@inproceedings{shirakawa2024noisecollage,
  title={NoiseCollage: A Layout-Aware Text-to-Image Diffusion Model Based on Noise Cropping and Merging},
  author={Shirakawa, Takahiro and Uchida, Seiichi},
  booktitle={CVPR},
  year={2024}
}

@inproceedings{kwon2024concept,
  title={Concept Weaver: Enabling Multi-Concept Fusion in Text-to-Image Models},
  author={Kwon, Gihyun and Jenni, Simon and Li, Dingzeyu and Lee, Joon-Young and Ye, Jong Chul and Heilbron, Fabian Caba},
  booktitle={CVPR},
  year={2024}
}

@inproceedings{meral2024conform,
  title={Conform: Contrast is all you need for high-fidelity text-to-image diffusion models},
  author={Meral, Tuna Han Salih and Simsar, Enis and Tombari, Federico and Yanardag, Pinar},
  booktitle={CVPR},
  year={2024}
}

@inproceedings{jayasumana2024markovgen,
  title={MarkovGen: Structured Prediction for Efficient Text-to-Image Generation},
  author={Jayasumana, Sadeep and Glasner, Daniel and Ramalingam, Srikumar and Veit, Andreas and Chakrabarti, Ayan and Kumar, Sanjiv},
  booktitle={Proceedings of the IEEE/CVF Conference on Computer Vision and Pattern Recognition},
  pages={9316--9325},
  year={2024}
}

@article{li2024playground,
  title={Playground v2. 5: Three insights towards enhancing aesthetic quality in text-to-image generation},
  author={Li, Daiqing and Kamko, Aleks and Akhgari, Ehsan and Sabet, Ali and Xu, Linmiao and Doshi, Suhail},
  journal={arXiv preprint arXiv:2402.17245},
  year={2024}
}

@inproceedings{grimal2024tiam,
  title={TIAM-A metric for evaluating alignment in Text-to-Image generation},
  author={Grimal, Paul and Le Borgne, Herv{\'e} and Ferret, Olivier and Tourille, Julien},
  booktitle={Proceedings of the IEEE/CVF Winter Conference on Applications of Computer Vision},
  pages={2890--2899},
  year={2024}
}

@article{zhang2024transparent,
  title={Transparent image layer diffusion using latent transparency},
  author={Zhang, Lvmin and Agrawala, Maneesh},
  journal={arXiv preprint arXiv:2402.17113},
  year={2024}
}

@inproceedings{zhang2024tackling,
  title={Tackling the Singularities at the Endpoints of Time Intervals in Diffusion Models},
  author={Zhang, Pengze and Yin, Hubery and Li, Chen and Xie, Xiaohua},
  booktitle={CVPR},
  year={2024}
}

@inproceedings{lu2024coarse,
  title={Coarse-to-fine latent diffusion for pose-guided person image synthesis},
  author={Lu, Yanzuo and Zhang, Manlin and Ma, Andy J and Xie, Xiaohua and Lai, Jianhuang},
  booktitle={CVPR},
  year={2024}
}

@inproceedings{wang2024detdiffusion,
  title={Detdiffusion: Synergizing generative and perceptive models for enhanced data generation and perception},
  author={Wang, Yibo and Gao, Ruiyuan and Chen, Kai and Zhou, Kaiqiang and Cai, Yingjie and Hong, Lanqing and Li, Zhenguo and Jiang, Lihui and Yeung, Dit-Yan and Xu, Qiang and others},
  booktitle={CVPR},
  year={2024}
}

@inproceedings{cai2024condition,
  title={Condition-Aware Neural Network for Controlled Image Generation},
  author={Cai, Han and Li, Muyang and Zhang, Qinsheng and Liu, Ming-Yu and Han, Song},
  booktitle={CVPR},
  year={2024}
}

@inproceedings{ren2024move,
  title={Move Anything with Layered Scene Diffusion},
  author={Ren, Jiawei and Xu, Mengmeng and Wu, Jui-Chieh and Liu, Ziwei and Xiang, Tao and Toisoul, Antoine},
  booktitle={CVPR},
  year={2024}
}

@inproceedings{wang2024towards,
  title={Towards Effective Usage of Human-Centric Priors in Diffusion Models for Text-based Human Image Generation},
  author={Wang, Junyan and Sun, Zhenhong and Tan, Zhiyu and Chen, Xuanbai and Chen, Weihua and Li, Hao and Zhang, Cheng and Song, Yang},
  booktitle={CVPR},
  year={2024}
}

@article{geng2024motion,
  title={Motion guidance: Diffusion-based image editing with differentiable motion estimators},
  author={Geng, Daniel and Owens, Andrew},
  journal={arXiv preprint arXiv:2401.18085},
  year={2024}
}

@inproceedings{ohanyan2024zero,
  title={Zero-Painter: Training-Free Layout Control for Text-to-Image Synthesis},
  author={Ohanyan, Marianna and Manukyan, Hayk and Wang, Zhangyang and Navasardyan, Shant and Shi, Humphrey},
  booktitle={CVPR},
  year={2024}
}

@inproceedings{mo2024freecontrol,
  title={Freecontrol: Training-free spatial control of any text-to-image diffusion model with any condition},
  author={Mo, Sicheng and Mu, Fangzhou and Lin, Kuan Heng and Liu, Yanli and Guan, Bochen and Li, Yin and Zhou, Bolei},
  booktitle={CVPR},
  year={2024}
}

@inproceedings{ma2024adapedit,
  title={AdapEdit: Spatio-Temporal Guided Adaptive Editing Algorithm for Text-Based Continuity-Sensitive Image Editing},
  author={Ma, Zhiyuan and Jia, Guoli and Zhou, Bowen},
  booktitle={AAAI},
  year={2024}
}

@inproceedings{yu2024accelerating,
  title={Accelerating Text-to-Image Editing via Cache-Enabled Sparse Diffusion Inference},
  author={Yu, Zihao and Li, Haoyang and Fu, Fangcheng and Miao, Xupeng and Cui, Bin},
  booktitle={AAAI},
  year={2024}
}

@inproceedings{brack2024ledits++,
  title={Ledits++: Limitless image editing using text-to-image models},
  author={Brack, Manuel and Friedrich, Felix and Kornmeier, Katharia and Tsaban, Linoy and Schramowski, Patrick and Kersting, Kristian and Passos, Apolin{\'a}rio},
  booktitle={CVPR},
  year={2024}
}

@inproceedings{mahajan2024prompting,
  title={Prompting Hard or Hardly Prompting: Prompt Inversion for Text-to-Image Diffusion Models},
  author={Mahajan, Shweta and Rahman, Tanzila and Yi, Kwang Moo and Sigal, Leonid},
  booktitle={CVPR},
  year={2024}
}

@inproceedings{wang2024promptcharm,
  title={PromptCharm: Text-to-Image Generation through Multi-modal Prompting and Refinement},
  author={Wang, Zhijie and Huang, Yuheng and Song, Da and Ma, Lei and Zhang, Tianyi},
  booktitle={Proceedings of the CHI Conference on Human Factors in Computing Systems},
  year={2024}
}

@inproceedings{xue2024raphael,
  title={Raphael: Text-to-image generation via large mixture of diffusion paths},
  author={Xue, Zeyue and Song, Guanglu and Guo, Qiushan and Liu, Boxiao and Zong, Zhuofan and Liu, Yu and Luo, Ping},
  booktitle={NeurIPS},
  year={2024}
}

@inproceedings{sueyoshi2024predicated,
  title={Predicated Diffusion: Predicate Logic-Based Attention Guidance for Text-to-Image Diffusion Models},
  author={Sueyoshi, Kota and Matsubara, Takashi},
  booktitle={CVPR},
  year={2024}
}

@inproceedings{qiao2024baret,
  title={BARET: Balanced Attention Based Real Image Editing Driven by Target-Text Inversion},
  author={Qiao, Yuming and Wang, Fanyi and Su, Jingwen and Zhang, Yanhao and Yu, Yunjie and Wu, Siyu and Qi, Guo-Jun},
  booktitle={AAAI},
  year={2024}
}

@article{feng2024item,
  title={An item is worth a prompt: Versatile image editing with disentangled control},
  author={Feng, Aosong and Qiu, Weikang and Bai, Jinbin and Zhang, Xiao and Dong, Zhen and Zhou, Kaicheng and Ying, Rex and Tassiulas, Leandros},
  journal={arXiv preprint arXiv:2403.04880},
  year={2024}
}

@inproceedings{zhang2024learning,
  title={Learning Multi-dimensional Human Preference for Text-to-Image Generation},
  author={Zhang, Sixian and Wang, Bohan and Wu, Junqiang and Li, Yan and Gao, Tingting and Zhang, Di and Wang, Zhongyuan},
  booktitle={CVPR},
  year={2024}
}

@inproceedings{huang2024realcustom,
  title={RealCustom: Narrowing Real Text Word for Real-Time Open-Domain Text-to-Image Customization},
  author={Huang, Mengqi and Mao, Zhendong and Liu, Mingcong and He, Qian and Zhang, Yongdong},
  booktitle={CVPR},
  year={2024}
}

@inproceedings{chan2024improving,
  title={Improving Subject-Driven Image Synthesis with Subject-Agnostic Guidance},
  author={Chan, Kelvin CK and Zhao, Yang and Jia, Xuhui and Yang, Ming-Hsuan and Wang, Huisheng},
  booktitle={CVPR},
  year={2024}
}

@article{cheng2024resadapter,
  title={ResAdapter: Domain Consistent Resolution Adapter for Diffusion Models},
  author={Cheng, Jiaxiang and Xie, Pan and Xia, Xin and Li, Jiashi and Wu, Jie and Ren, Yuxi and Li, Huixia and Xiao, Xuefeng and Zheng, Min and Fu, Lean},
  journal={arXiv preprint arXiv:2403.02084},
  year={2024}
}

@article{wang2024instantid,
  title={Instantid: Zero-shot identity-preserving generation in seconds},
  author={Wang, Qixun and Bai, Xu and Wang, Haofan and Qin, Zekui and Chen, Anthony and Li, Huaxia and Tang, Xu and Hu, Yao},
  journal={arXiv preprint arXiv:2401.07519},
  year={2024}
}

@inproceedings{ham2024personalized,
  title={Personalized Residuals for Concept-Driven Text-to-Image Generation},
  author={Ham, Cusuh and Fisher, Matthew and Hays, James and Kolkin, Nicholas and Liu, Yuchen and Zhang, Richard and Hinz, Tobias},
  booktitle={CVPR},
  year={2024}
}

@inproceedings{li2024stylegan,
  title={When stylegan meets stable diffusion: a w+ adapter for personalized image generation},
  author={Li, Xiaoming and Hou, Xinyu and Loy, Chen Change},
  booktitle={CVPR},
  year={2024}
}

@inproceedings{mei2024codi,
  title={CoDi: Conditional Diffusion Distillation for Higher-Fidelity and Faster Image Generation},
  author={Mei, Kangfu and Delbracio, Mauricio and Talebi, Hossein and Tu, Zhengzhong and Patel, Vishal M and Milanfar, Peyman},
  booktitle={CVPR},
  year={2024}
}

@article{li2024styletokenizer,
  title={StyleTokenizer: Defining Image Style by a Single Instance for Controlling Diffusion Models},
  author={Li, Wen and Fang, Muyuan and Zou, Cheng and Gong, Biao and Zheng, Ruobing and Wang, Meng and Chen, Jingdong and Yang, Ming},
  journal={arXiv preprint arXiv:2409.02543},
  year={2024}
}

@article{xiao2024fastcomposer,
  title={Fastcomposer: Tuning-free multi-subject image generation with localized attention},
  author={Xiao, Guangxuan and Yin, Tianwei and Freeman, William T and Durand, Fr{\'e}do and Han, Song},
  journal={Springer IJCV},
  pages={1--20},
  year={2024}
}

@inproceedings{ruiz2024hyperdreambooth,
  title={Hyperdreambooth: Hypernetworks for fast personalization of text-to-image models},
  author={Ruiz, Nataniel and Li, Yuanzhen and Jampani, Varun and Wei, Wei and Hou, Tingbo and Pritch, Yael and Wadhwa, Neal and Rubinstein, Michael and Aberman, Kfir},
  booktitle={CVPR},
  year={2024}
}

@inproceedings{li2024photomaker,
  title={Photomaker: Customizing realistic human photos via stacked id embedding},
  author={Li, Zhen and Cao, Mingdeng and Wang, Xintao and Qi, Zhongang and Cheng, Ming-Ming and Shan, Ying},
  booktitle={CVPR},
  year={2024}
}

@inproceedings{huang2024learning,
  title={Learning disentangled identifiers for action-customized text-to-image generation},
  author={Huang, Siteng and Gong, Biao and Feng, Yutong and Chen, Xi and Fu, Yuqian and Liu, Yu and Wang, Donglin},
  booktitle={CVPR},
  year={2024}
}

@inproceedings{chen2024artadapter,
  title={ArtAdapter: Text-to-Image Style Transfer using Multi-Level Style Encoder and Explicit Adaptation},
  author={Chen, Dar-Yen and Tennent, Hamish and Hsu, Ching-Wen},
  booktitle={CVPR},
  year={2024}
}

@inproceedings{shi2024instantbooth,
  title={Instantbooth: Personalized text-to-image generation without test-time finetuning},
  author={Shi, Jing and Xiong, Wei and Lin, Zhe and Jung, Hyun Joon},
  booktitle={CVPR},
  year={2024}
}

@article{kim2024automatic,
  title={Automatic Jailbreaking of the Text-to-Image Generative AI Systems},
  author={Kim, Minseon and Lee, Hyomin and Gong, Boqing and Zhang, Huishuai and Hwang, Sung Ju},
  journal={arXiv preprint arXiv:2405.16567},
  year={2024}
}

\clearpage
\normalsize
\appendix
\section{Development of GAN, AR, and DM in T2I}
\label{section3back}

\subsection{Developments of GAN Models in T2I}

The work \cite{reed2016generative} in 2016 marked the first application of GANs in T2I, utilizing a conditional GAN (cGAN) structure \cite{mirza2014conditional}, which laid the foundation for subsequent research.

The following year, StackGAN \cite{zhang2017stackgan} was introduced, which generates images in a two-stage manner. In the first stage, basic shapes and colors were drawn on the basis of the given text description to produce low-resolution images. In the second stage, the output from the first stage was refined by combining it with the text description to generate high-resolution images, significantly improving the quality of the generated images. Building on this, StackGAN++ \cite{zhang2018stackgan++} adopted a tree structure with multiple generators and discriminators to improve image generation. The images were progressively refined from low to high resolution through different branches of the tree, further improving image details and consistency, while enhancing the model's ability to understand text descriptions.
Meanwhile, TAC-GAN \cite{dash2017tac} introduced an adaptive control mechanism that incorporated auxiliary classification tasks into the discriminator, allowing it to more accurately identify text categories and thereby produce images that better align with text descriptions.

In 2018, AttnGAN \cite{xu2018attngan} built on StackGAN++ by integrating attention mechanisms into a multistage optimization pipeline. This integration allowed the model to focus more on the key elements of the text, thereby enhancing semantic consistency in the generated images. Beyond using global sentence vectors, the attention mechanism also enabled the synthesis of fine-grained details based on relevant words. Fused GAN \cite{bodla2018semi} combined multimodal information from both images and text, using two generators for conditional and unconditional image synthesis. These generators partially shared a common latent space, allowing one generator to perform both types of synthesis simultaneously. HDGAN \cite{zhang2018photographic} focused on generating high-resolution images, with a generator designed to capture complex image features. Using a deep generative model architecture, HDGAN improved the detail representation of images, resulting in more coherent output.

In 2019, the development of GANs in the T2I field flourished. ControlGAN \cite{lee2019controllable} enabled users to control specific attributes of generated images, allowing manipulation of visual properties such as categories, textures, and colors. DM-GAN \cite{zhu2019dm} integrated diffusion model concepts and introduced a dynamic memory module to refine blurred image content, helping to prevent poor results during initial image generation. GauGAN \cite{park2019gaugan} was noted for generating realistic images from user-drawn sketches, using semantic layouts to produce photorealistic effects, thereby providing an interactive image generation experience. MirrorGAN \cite{qiao2019mirrorgan} employed a bidirectional generation mechanism to map text to images and back, improving semantic coherence in the outputs. This model introduced a new framework that combined global and local attention with semantic preservation, redefining text-to-image generation. Obj-GAN \cite{li2019object} focused on generating images with specific objects and attributes, enhancing the targeted nature of generation and enabling the synthesis of complex scenes from textual descriptions through attention-driven multistage optimization. Text-SeGAN \cite{cha2019adversarial} utilized sequential generation to improve coherence in generated images, allowing the synthesis of diverse images from text descriptions.

In 2020, CookGAN \cite{zhu2020cookgan} focused on image generation in specific domains, optimizing the generation process to simulate visual effects within causal chains, thus demonstrating the flexibility of GANs. CPGAN \cite{liang2020cpgan} enhanced semantic consistency between text and images by delving deeply into their content. RiFeGAN \cite{cheng2020rifegan} adopted a fine-grained control strategy, becoming a T2I model with novel and rich features. SegAttnGAN \cite{gou2020segattngan} integrated additional segmentation information with attention mechanisms, further improving the realism of generated images.

In 2021, VQ-GAN \cite{esser2021taming} achieved efficient image generation by compressing continuous data and using vector quantization. XMC-GAN \cite{zhang2021cross} enhanced the complementarity between text and images through cross-modal learning, employing intermodal and intramodal contrastive learning to maximize mutual information, thus addressing the cross-modal contrastive loss issue in T2I generation. DAE-GAN \cite{ruan2021dae} incorporated the concepts of denoising autoencoders to improve the robustness of generated images. CI-GAN \cite{wang2021cycle} focused on diversifying image content and style, enhancing the artistic quality of the generated output. R-GAN \cite{qiao2021r} generated coherent images in a human-like manner, demonstrating naturalness and consistency in its generation process.

In 2022, DF-GAN \cite{tao2022df} directly generated HR images in the first stage and introduced a novel target-aware discriminator in the second stage to improve the consistency between text and images. LAFITE \cite{zhou2022towards} used CLIP to guide the generation process, enhancing the quality and interpretability of the generated images. SSA-GAN \cite{liao2022text} introduced an adaptive selection mechanism to refine generation details based on different text descriptions. VLMGAN \cite{cheng2022vision} combined visual and language models to improve semantic understanding capabilities in image generation. VQGAN-CLIP \cite{crowson2022vqgan} integrated VQGAN and CLIP models to allow efficient T2I generation. AttnGAN++ \cite{dinh2022tise} expanded on AttnGAN by employing spectral normalization to stabilize training.

\begin{figure*}   
	\centering
    \includegraphics[width=1\linewidth]{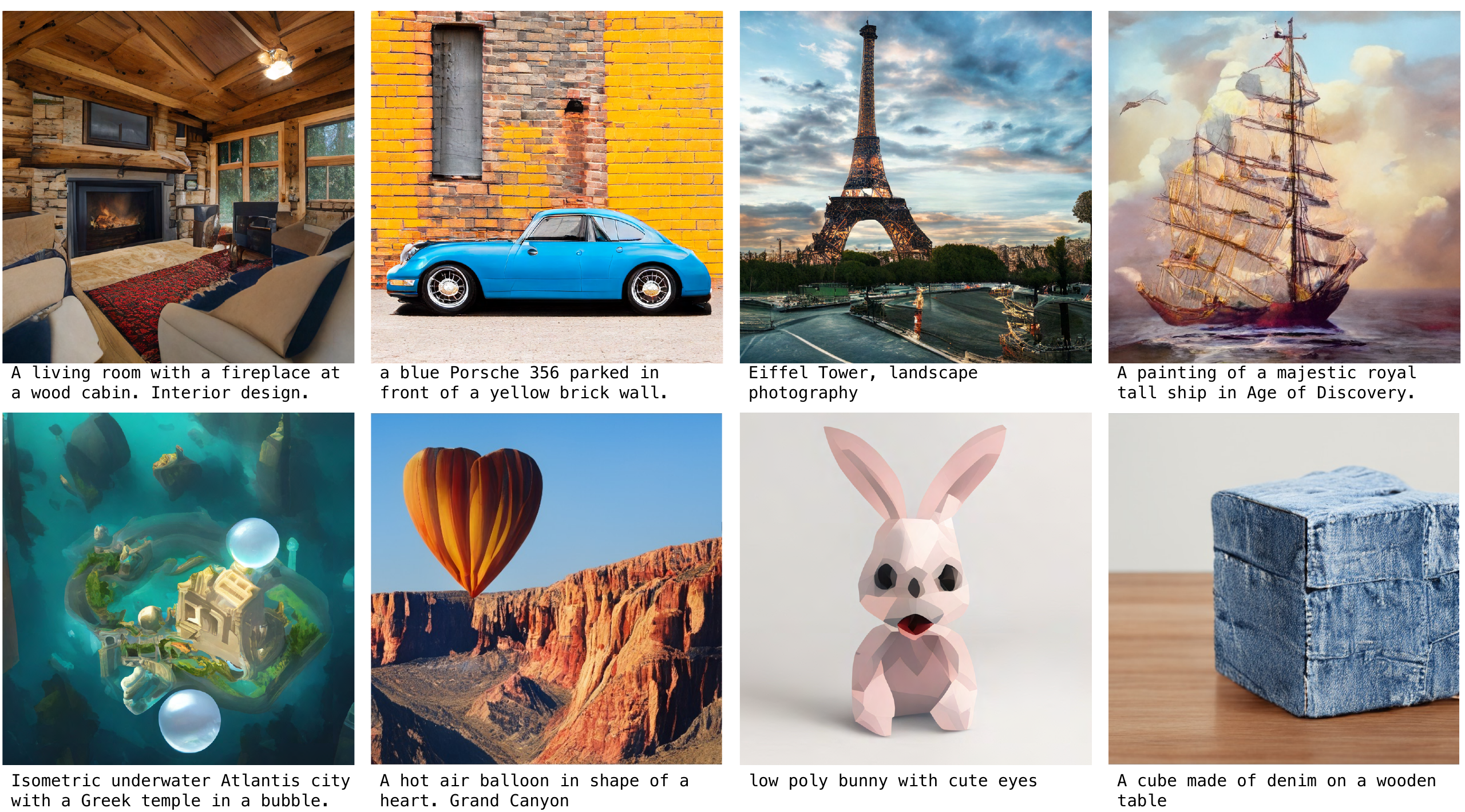} 
	\caption{The model, GigaGAN \cite{kang2023scaling}, shows GAN frameworks can also be scaled up for general T2I synthesis and superresolution tasks, generating a 512px output at an interactive speed of 0.13s, and 4096px within 3.7s. Selected examples at 2K
resolution. }  
	\label{gigaGan}
        \vspace{-0.4cm}
\end{figure*}

In 2023, RAT-GAN \cite{ye2023recurrent} adopted recursive affine transformations to incorporate conditional information during generation. GALIP \cite{tao2023galip} improved the quality of the generated images by combining image generation with language understanding, integrating CLIP into both the discriminator and generator. GigaGAN \cite{kang2023scaling} scales GAN training to large datasets and reports image quality competitive with contemporary diffusion models while offering faster, near-single-step sampling, as illustrated in Figure \ref{gigaGan}. StyleGAN-T \cite{sauer2023stylegan} focused on style transfer, increasing the diversity of generated images. Layout-VQGAN \cite{wu2022text} used layout information to generate complex scenes, enhancing the structural representation of the generated images. Meanwhile, DragGAN \cite{pan2023drag} introduced a point-based interactive image editing method, allowing pixel-level precision in the editing results.

In recent research, UFOGen \cite{xu2024ufogen} aims at ultra-fast one-step T2I using a hybrid approach that combines diffusion models with GAN objectives. HyperCGAN \cite{haydarov2024adversarial} addresses the T2I task from a different perspective, representing two-dimensional images as implicit neural representations (INRs) \cite{sitzmann2020implicit} and controlling the INR-GAN generation process using weight modulation operators based on word-level attention, through the use of a supernetwork.

Collectively, these GAN variants improved generation quality and controllability and broadened the application scope of T2I.

\subsection{Developments of AR Models in T2I}
DALL-E \cite{ramesh2021zero}, based on the Transformer architecture \cite{vaswani2017attention}, became the first large-scale autoregressive T2I model, demonstrating the tremendous potential of autoregressive models in T2I tasks. Building on this foundation, CogView \cite{ding2021cogview} further improved generation quality by combining VQ-VAE with autoregressive strategies, producing diverse images, and improving the user-interaction experience. Meanwhile, M6 \cite{lin2021m6} provided a large dataset for multimodal pre-training in Chinese, and Parti \cite{yu2022scaling} enhanced the detail representation of generated images through image patch processing, similar to machine translation, where image token sequences are used as target outputs.

CogView2 \cite{ding2022cogview2}, an upgraded version of CogView, not only improved generation quality but also supported text-based interactive image editing. Make-A-Scene \cite{gafni2022make} introduced scene understanding capabilities, allowing users to specify elements in the generated images more precisely, thus enhancing control and accuracy. NÜWA \cite{wu2022nuwa} combined autoregressive generation with multimodal information, allowing the creation of new visual data or the manipulation of existing visual data (such as images and videos) to meet various visual synthesis tasks. Emu \cite{dai2023emu} proposed a quality tuning method that effectively guides the pretrained model to generate visually appealing images while maintaining generality with respect to visual concepts. CM3Leon \cite{yu2023scaling} used an improved autoregressive mechanism and became the first multimodal model trained using an approach adapted from text-specific language models, including a large-scale retrieval-augmented pretraining phase followed by a multitask supervised fine-tuning (SFT) phase.

\begin{figure*}    
	\centering
    \includegraphics[width=1\linewidth]{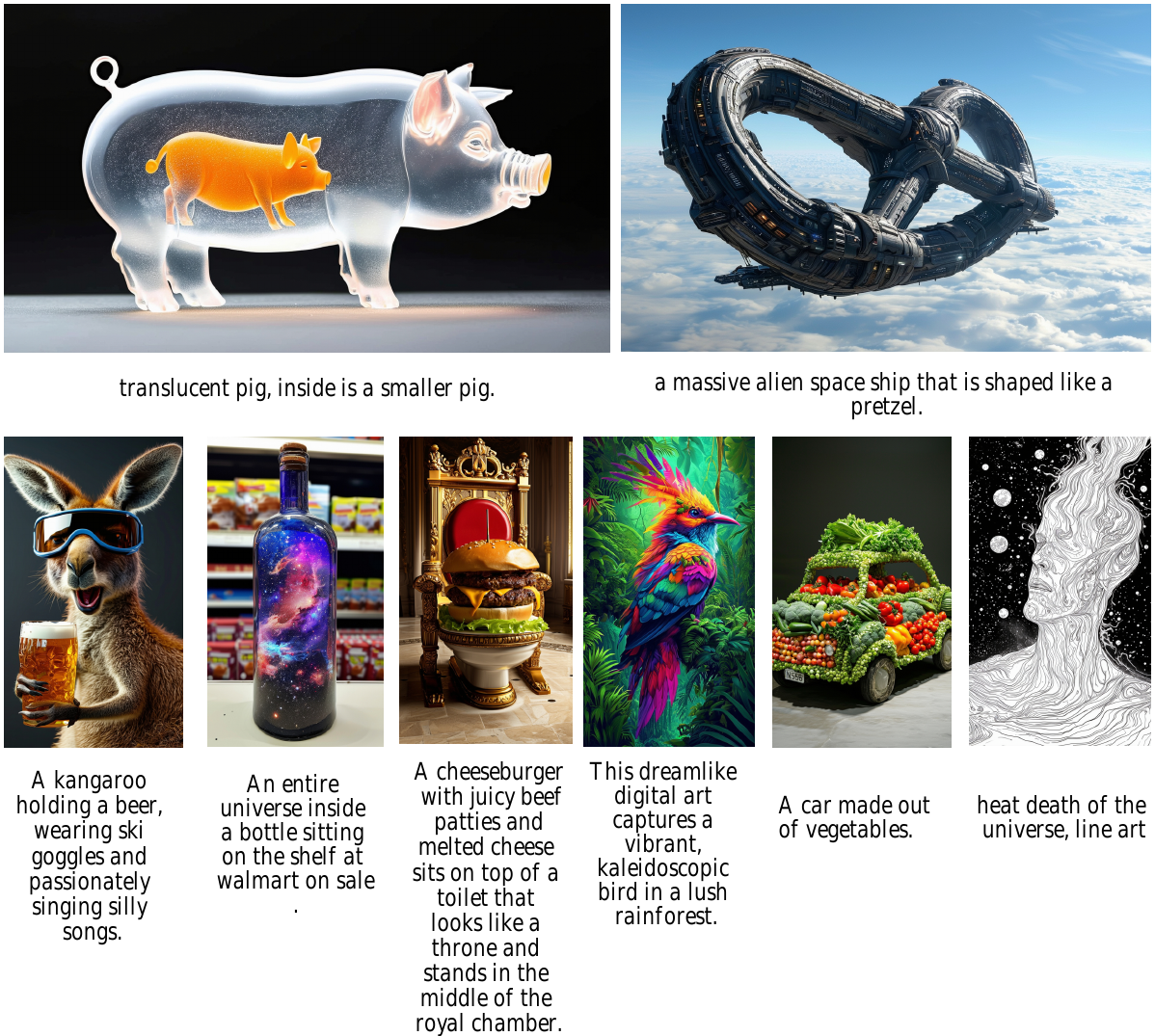}  
	\caption{Sampling is derived from Stable Diffusion 3 \cite{esser2024scaling}, generating fantastical pictures.} 
	\label{sd3}
        \vspace{-0.4cm}
\end{figure*}

GILL \cite{koh2024generating} integrated frozen text-only large language models (LLMs) \cite{radford2018improving,devlin2018bert} with pretrained image encoder and decoder models through autoregressive modeling, mapping between their embedding spaces. MARS \cite{he2024mars} retained the NLP capabilities of LLMs while also providing them with enhanced visual understanding. Unified-IO 2 \cite{lu2024unified} consolidated autoregressive generative capabilities across multiple tasks, including T2I generation, image understanding, and image-to-image generation, demonstrating broad applications in T2I. STAR \cite{ma2024star} adopted a scale-wise autoregressive paradigm for T2I modeling, enhancing the coherence of the generated images and improving alignment with the text descriptions. HART \cite{tang2024hart} introduced a hybrid tokenizer to achieve efficient high-resolution visual synthesis, combining autoregressive models with adversarial generation to enhance the diversity and realism of the generated images.

Recently, the SJD method used in AR \cite{teng2024accelerating} enhanced the richness of generated content by integrating autoregressive strategies with structured data and introducing Jacobi techniques to accelerate the autoregressive model. On the other hand, DART \cite{gu2024dart} utilized a denoising autoregressive encoder, incorporating a dynamic adjustment mechanism into the generation process, thus improving the adaptability to complex text descriptions and increasing the diversity of generated outputs.

These autoregressive models progressively improved generation quality and multimodal flexibility, and their token interface later became the basis for the unified understanding-and-generation systems discussed in Section~\ref{sec:unified}.

\subsection{Rapid Rise of Diffusion Models in T2I}
GLIDE \cite{nichol2021glide} was the pioneering application of diffusion models in T2I, marking an important milestone and highlighting the potential of diffusion models for T2I tasks.

VQ-Diffusion \cite{gu2022vector} uses vector quantization technology for efficient generation, while Imagen \cite{saharia2022photorealistic} combines the strengths of Transformer language models and high-fidelity diffusion models by using a pretrained language model to encode text. Stable Diffusion (LDMs) \cite{rombach2022high} offers strong generative capabilities through latent space optimization. DALL-E 2 \cite{ramesh2022hierarchical} is a two-stage model: first, a prior model generates CLIP image embeddings based on text descriptions, and then a decoder generates images conditioned on these embeddings.

\begin{figure*}  
	\centering
    \includegraphics[width=1\linewidth]{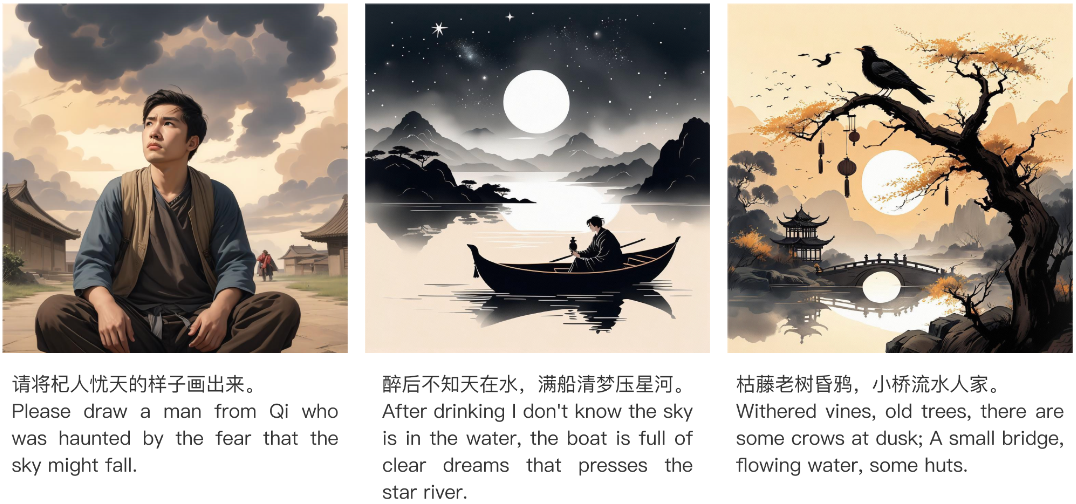} 
	\caption{Hunyuan-DiT \cite{li2024hunyuan} can generate images containing Chinese elements; the examples shown are generated directly from Chinese prompts.}
	\label{hunyuan}
        \vspace{-0.4cm}
\end{figure*}

Additionally, Blended Diffusion \cite{avrahami2022blended} uses natural language guidance to generate realistic and diverse images. PixArt-$ \alpha $ \cite{chen2023pixart}, based on the Transformer T2I diffusion model, excels in artistic style generation. DALLE-3 \cite{betker2023improving} combines an image encoder with the GPT-4 language model \cite{achiam2023gpt}, allowing it to generate detailed textual descriptions for each image, further enhancing text understanding and generation quality. Stable Diffusion XL \cite{podell2023sdxl} is an upgraded version of Stable Diffusion, featuring a backbone network \cite{ronneberger2015u} that is three times larger. DreamBooth \cite{ruiz2023dreambooth} allows users to customize image generation with a few examples, while ControlNet \cite{zhang2023adding} introduces additional control parameters into the generation process, enhancing flexibility.

Specialist Diffusion \cite{lu2023specialist} extends the existing single-stream diffusion process into a multi-task, multi-modal network called Versatile Diffusion (VD), which can handle multiple streams such as text-to-image, image-to-text, and their variants within a unified model. UniDiffuser \cite{bao2023one} integrates various generative capabilities based on a Transformer framework that accommodates multimodal data distributions, allowing it to process text-to-image, image-to-text, and joint image-text tasks simultaneously. Multimodal-CoT \cite{zhang2023multimodal} uses a chain-of-thought mechanism \cite{wei2022chain} to combine language (text) and visual (image) modalities in a two-stage framework that separates inference generation from answer reasoning. This approach allows answer reasoning to build on a stronger inferential foundation generated from multi-modal information.

Versatile Diffusion \cite{xu2023versatile} is the first unified multistream multimodal diffusion framework, while Custom Diffusion \cite{kumari2023multi} is an efficient enhancement method for existing T2I models, allowing users to customize the generation process and achieve rapid fine-tuning. SVDiff \cite{han2023svdiff} involves fine-tuning the singular values of weight matrices \cite{dodge2020fine}, creating a compact and efficient parameter space that reduces the risk of overfitting and language drift. ERNIE-ViLG 2 \cite{feng2023ernie} is the first large T2I model in the Chinese language domain, integrating fine-grained textual knowledge and key visual elements, such as adding part-of-speech descriptions (verbs, nouns, adjectives) to the text and incorporating object recognition preprocessing for images.

Corgi \cite{zhou2023shifted} proposes an offset diffusion model that improves the generation of image embeddings from input text. Würstchen \cite{pernias2023wurstchen} develops a latent diffusion technique that uses a detailed yet compact semantic image representation to guide the diffusion process. ConPreDiff \cite{yang2024improving} improves diffusion-based image synthesis through context prediction, while ContextDiff \cite{yang2024cross} incorporates the cross-modal context, including interactions and alignment between text conditions and visual samples, into both forward and backward diffusion processes.

CogView3 \cite{zheng2024cogview3} is the first model to implement relay diffusion \cite{teng2023relay} in T2I. Stable Diffusion 3 (SD3) \cite{esser2024scaling} introduces a new Transformer-based architecture for T2I that assigns different weights to text and image modalities, allowing bidirectional information flow between images and text annotations; examples of its generated results are shown in Figure \ref{sd3}.
Ranni \cite{feng2024ranni} provides a control panel for managing T2I, while RPG \cite{yang2024mastering} uses a Multimodal Large Language Model (MLLM) \cite{brown2020language} as a global planner. It breaks down the process of generating complex images into simple generation tasks across multiple subregions, using complementary regional diffusion to achieve region-level combinatorial generation. Additionally, RPG integrates text-guided image generation and editing in a closed-loop manner, enhancing the model's generalization capabilities. DistriFusion \cite{li2024distrifusion} leverages parallelism across multiple GPUs, supports asynchronous communication, and can be integrated with computational pipelines for fast T2I generation. InstanceDiffusion \cite{wang2024instancediffusion} enhances precise instance-level control, while Instruct-Imagen \cite{hu2024instruct} is designed to handle diverse image generation tasks and generalize to unseen tasks.

\begin{figure*}[!t]     
	\centering
    \includegraphics[width=0.8\linewidth]{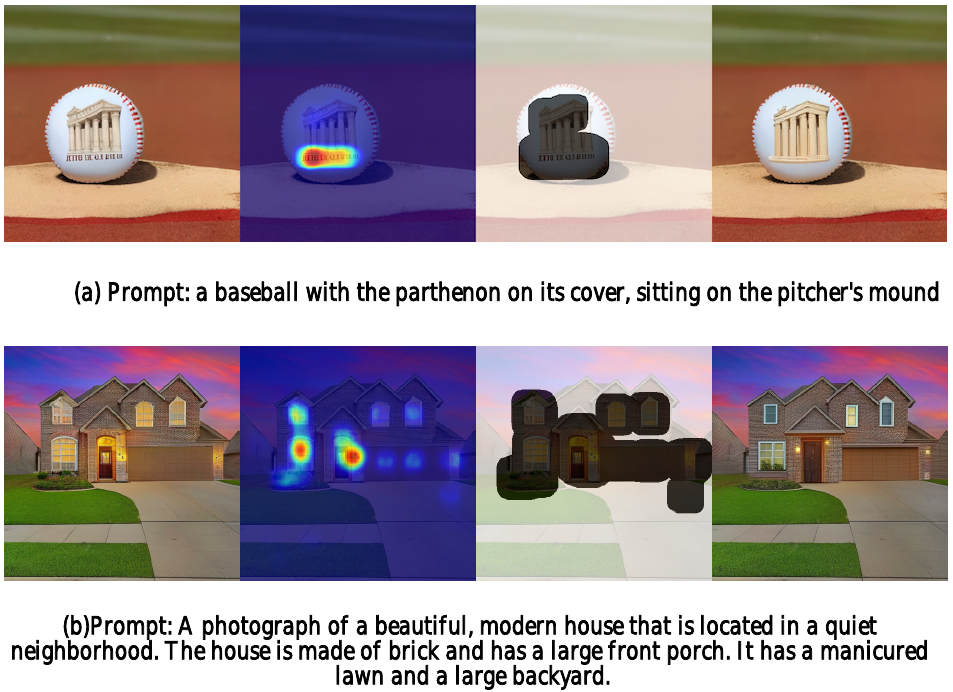} 
	\caption{RAHF\cite{liang2024rich} region inpainting with Muse\cite{chang2023muse} generative model. From left to right, original images with artifacts from Muse, implausibility heatmaps predicted by RAHF, masks by processing (thresholding, dilating) the heatmaps, and new images from Muse
region inpainting with the mask, respectively.} 
	\label{rahf}
        \vspace{-0.4cm}
\end{figure*}

ElasticDiffusion \cite{haji2024elasticdiffusion} introduces a new decoding strategy that uses a pretrained diffusion model to generate images of arbitrary sizes while maintaining a constant memory footprint during inference. PIXART-$\delta$ \cite{chen2024pixart1} significantly accelerates inference speed through the integration of LCM \cite{bachman2019learning}, enabling high-quality image generation in just 2-4 steps. It also introduces an innovative ControlNet-Transformer \cite{zhang2023adding} architecture specifically designed for transformers, providing explicit controllability and high-quality image generation. PixArt-$\Sigma$ \cite{chen2024pixart2} offers significantly higher image fidelity and better alignment with text prompts.

Recently, SD3-Turbo \cite{sauer2024fast} introduced Latent Adversarial Diffusion Distillation (LADD), a novel distillation method that overcomes the limitations of Adversarial Diffusion Distillation (ADD) \cite{sauer2025adversarial}. Unlike the pixel-based ADD, LADD leverages the generative features of a pretrained latent diffusion model. StreamMultiDiffusion \cite{lee2024streammultidiffusion} is the first real-time region-based T2I generation framework. OmniGen \cite{xiao2024omnigen} presents a new type of diffusion model for unified image generation, while Hunyuan-DiT \cite{li2024hunyuan} is a T2I diffusion transformer with fine-grained understanding capabilities. The results of the Hunyuan-DiT generation are shown in Figure \ref{hunyuan}. In terms of image generation in Chinese, Hunyuan-DiT outperforms existing open source models, including SD3, in text-image consistency, elimination of AI artifacts, subject clarity and aesthetics. Its performance in subject clarity and aesthetics is comparable to that of leading closed-source models like DALLE-3. DragDiffusion \cite{shi2024dragdiffusion} optimizes only the latent variables for a single time step, achieving efficient and precise spatial control, which significantly enhances the applicability of interactive point-based editing for both real and diffusion-generated images.

The rapid development of these models has not only advanced T2I generation technology but also created new opportunities for multimodal generation.

\end{document}